\documentclass[10pt,twocolumn,letterpaper]{article}

\usepackage{cvpr}

\usepackage[normalem]{ulem}
\usepackage{circledsteps}
\usepackage{pifont}
\usepackage{makecell}
\usepackage{xcolor}
\usepackage{colortbl}
\usepackage{tablefootnote}
\usepackage{multirow}
\usepackage{subcaption}
\usepackage{booktabs}
\usepackage{graphicx}
\usepackage{adjustbox}
\usepackage{soul}
\usepackage{afterpage}
\usepackage[table]{xcolor}
\usepackage{threeparttable}
\usepackage{algorithm}
\usepackage{algpseudocode}
\usepackage{soul}
\usepackage{cite}

\algrenewcommand\algorithmicrequire{\textbf{Input:}}
\algrenewcommand\algorithmicensure{\textbf{Output:}}

\newcommand{\softmidrule}{\arrayrulecolor{gray!40}\midrule\arrayrulecolor{black}}

\usepackage{pifont}
\newcommand{\xmark}{\ding{55}}

\newcommand{\namehighlight}[1]{\underline{\textbf{#1}}} 

\definecolor{darkblue}{RGB}{0, 50, 150}
\definecolor{mediumblue}{RGB}{0, 100, 200}
\newcommand{\darkblueref}[2]{\hyperref[#1]{\textcolor{mediumblue}{\textbf{#2}}}}
\newcommand{\blueref}[2]{\hyperref[#1]{\textcolor{mediumblue}{#2}}}

\definecolor{darkgreen}{rgb}{0.0, 0.5, 0.0}
\definecolor{darkred}{rgb}{0.55, 0.0, 0.0}

\definecolor{cvprblue}{rgb}{0.21,0.49,0.74}
\usepackage[pagebackref,breaklinks,colorlinks,citecolor=cvprblue]{hyperref}

\title{SNAP: Towards Segmenting Anything in Any Point Cloud}

\author{Aniket Gupta$^{1, *}$ \quad Hanhui Wang$^{1, *}$ \quad Charles Saunders$^2$ \quad Aruni RoyChowdhury$^2$ \\
Hanumant Singh$^1$ \quad Huaizu Jiang$^1$ \\
\\
$^1$Northeastern University \quad $^2$MathWorks \\
\small{ $^*$ Equal contribution}
}
\newcommand{\shortname}{SNAP}
\begin{document}

\maketitle

\begin{abstract}

Interactive 3D point cloud segmentation enables efficient annotation of complex 3D scenes through user-guided prompts. However, current approaches are typically restricted in scope to a single domain (indoor or outdoor), and to a single form of user interaction (either spatial clicks or textual prompts). Moreover, training on multiple datasets often leads to negative transfer, resulting in domain-specific tools that lack generalizability.
To address these limitations, we present \textbf{SNAP} (\textbf{S}egment a\textbf{N}ything in \textbf{A}ny \textbf{P}oint cloud), a unified model for interactive 3D segmentation that supports both point-based and text-based prompts across diverse domains. Our approach achieves cross-domain generalizability by training on 7 datasets spanning indoor, outdoor, and aerial environments, while employing domain-adaptive normalization to prevent negative transfer. For text-prompted segmentation, we automatically generate mask proposals without human intervention and match them against CLIP embeddings of textual queries, enabling both panoptic and open-vocabulary segmentation.
We achieve state-of-the-art performance on 8 out of 9 zero-shot benchmarks for spatial-prompted segmentation and demonstrate competitive results on all 5 text-prompted benchmarks. These results show that a unified model can match or exceed specialized domain-specific approaches, providing a practical tool for scalable 3D annotation. Project page is at \href{https://neu-vi.github.io/SNAP/}{https://neu-vi.github.io/SNAP/}
\end{abstract}

\vspace{-1em}
\section{Introduction}

\begin{figure}[t]
    \centering
    \includegraphics[width=1\linewidth]{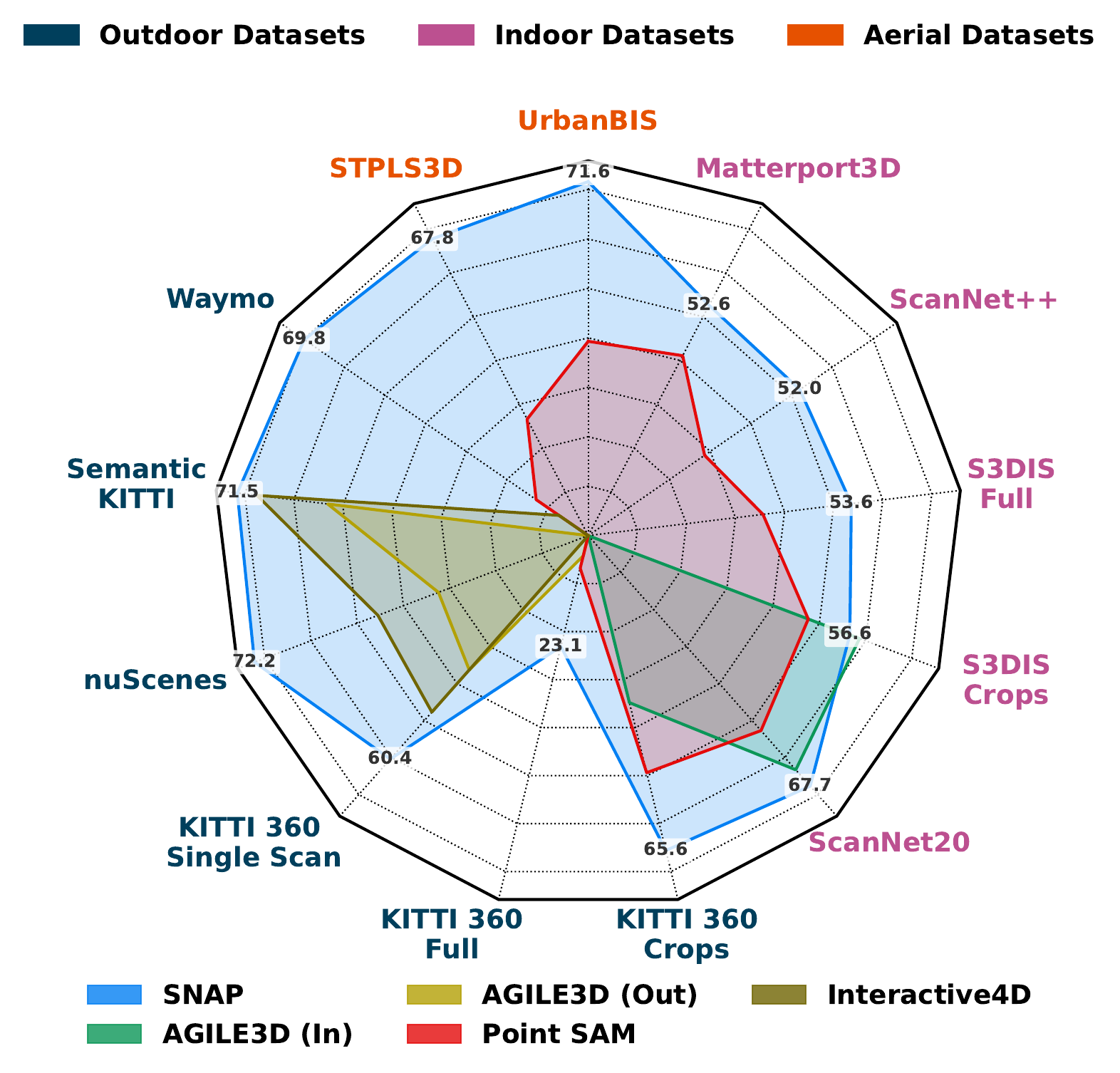} \\
    \vspace{-0.8em}
    \caption{\textbf{Comparison of models on IoU@1 Click across multiple domains.} \shortname\ is a unified interactive point cloud segmentation model trained on multiple datasets spanning multiple domains. It generalizes robustly across a wide array of benchmarks. 
    }
    \vspace{-15pt}
    \label{fig:intro_fig}
\end{figure}

Inspired by the success of SAM~\cite{SAM} for 2D images, we study interactive segmentation for 3D point clouds in this paper, allowing users to create high-quality annotations at scale with minimal effort. 
While supervised deep learning has fueled significant progress in visual learning, its reliance on vast labeled datasets presents a major bottleneck in the 3D domain, where manual annotation is notoriously difficult and time consuming.

Current interactive point cloud segmentation methods suffer from critical limitations that hinder their adoption as general-purpose annotation tools. Most existing approaches lack \emph{generalizability}, being designed for specific domains such as indoor scenes~\cite{InterObject3D, AGILE3D, POINT-SAM} or outdoor environments~\cite{interactive4D}. This domain-specific design stems from the significant statistical differences between point cloud types: indoor scenes feature dense, structured environments with clear object boundaries, while outdoor scenes contain sparse, large scale structures with varying point densities. Training models across these diverse domains introduces negative transfer effects that degrades performance without careful architectural considerations~\cite{pdnorm_short}. 
Additionally, current methods lack \emph{flexibility} in prompt modalities, typically supporting either spatial inputs (\eg, clicks)~\cite{POINT-SAM, AGILE3D, InterObject3D, interactive4D} or text descriptions\cite{SAL}, but rarely both. 
This limitation restricts users to specific annotation workflows, preventing adaptation to different labeling needs and use cases. 

To address these limitations, we present \textbf{\shortname} (\namehighlight{S}egment \namehighlight{A}nything in a\namehighlight{N}y \namehighlight{P}oint cloud), a unified model 
that supports both spatial and text-based prompts
To achieve cross-domain generalizability, we train on seven diverse datasets~\cite{semantickitti, nuscenes, pandaset, scannet, ramakrishnan2021hm3d, stpls3d, dales_instance}
spanning indoor, outdoor, and aerial domains, employing domain-wise normalization to mitigate negative transfer effects caused by statistical shifts between datasets~\cite{pdnorm}. 
For text-prompted segmentation, we first introduce a a simple iterative algorithm to automatically generate prompt points without human intervention. These prompt points are used via the spatial-prompted segmentation pipeline to generate mask proposals, which are then matched against 
CLIP~\cite{clip} embeddings of textual prompts to run text-prompted segmentation. 
During inference, this approach 
supports both panoptic segmentation with predefined categories and open-vocabulary segmentation with novel classes.

Extensive experiments demonstrate that \shortname\ consistently predicts high-quality spatial masks and correct class labels across a diverse range of indoor~\cite{scannet, scannetpp, Matterport3D, s3dis}, outdoor~\cite{kitti, kitti360, semantickitti, nuscenes, Sun_2020_CVPR, pandaset}, and aerial~\cite{stpls3d, dales_instance, varney2020dales, UrbanBIS} point clouds. 
For point-prompted segmentation, \shortname\ sets a new state-of-the-art on 8 out of 9 zero-shot benchmarks. 
For text-prompted segmentation, \shortname\ shows competitive results on all 5 evaluated benchmarks. 
These results demonstrate that a single, unified model can match or exceed the performance of specialized, domain-specific approaches, shown in \textbf{Fig.~\ref{fig:intro_fig}}. In summary, our contributions are as follows:
\begin{itemize}
    \item We introduce \shortname, a model for interactive point cloud segmentation that works across indoor, outdoor, and aerial domains.
    \item SNAP offers flexibility to use multi-modal prompts like points and text to segment objects of interest, and predicts classes as well as spatial masks.
    \item We introduce an automatic prompt points generation algorithm that bootstraps the panoptic labeling process without human intervention. 
    \item SNAP achieves state-of-the-art performance across multiple datasets across various domains and is usable as an out-of-the-box semi-automated labeling tool.
\end{itemize}

\section{Related Work}
\label{sec:related}

\noindent\textbf{Interactive 3D segmentation with spatial prompts.} Interactive segmentation has become well established in 2D since the introduction of SAM~\cite{SAM}, but remains comparatively underexplored in 3D. Early works such as InterObject3D\cite{InterObject3D} and AGILE3D\cite{AGILE3D} focused on indoor point clouds, using positive/negative clicks for single or multi-object segmentation. Point-SAM\cite{POINT-SAM} leveraged SAM-generated pseudo-labels to support indoor and part-level segmentation. 
Interactive4D~\cite{interactive4D} adopted a 4D setup on outdoor LiDAR sequences. 
However, these works are generally restricted in their usability to either indoor or outdoor scenes and show limited generalizability to out of domain point clouds. \shortname, in contrast, is designed to perform robustly across different domains.

\vspace{0.3em}
\noindent\textbf{Text-based 3D segmentation.} Another line of research introduces natural language as a flexible interface for 3D segmentation. Within this area, \textit{open-vocabulary segmentation} methods~\cite{Peng2023OpenScene,huang2024openins3dsnaplookup3d,openmask3d,yin2024sai3dsegmentinstance3d} aim to recognize novel or user-specified categories beyond a fixed label set. OpenScene~\cite{Peng2023OpenScene} achieves this directly in the 3D domain by predicting per-point CLIP embeddings without relying on images. In contrast, \cite{huang2024openins3dsnaplookup3d,openmask3d,yin2024sai3dsegmentinstance3d} utilize either original RGB images or rendered multiview images to extract CLIP image features, which are aligned with text embeddings to enable text-based segmentation. 
A complementary direction is \textit{panoptic segmentation}, where the goal is to provide instance-level predictions across all categories. ~\cite{SAL, Roh_2024_CVPR_EASE} exemplify this approach by predicting per-instance CLIP tokens and aligning them with a predefined class vocabulary to support class-specific instance segmentation. \shortname~is able to accommodate both \textit{open-vocabulary} and \textit{panoptic} settings within a single framework.

\vspace{0.3em}
\noindent\textbf{Lifting 2D foundation models for 3D segmentation.} 
Due to the limited availability of annotated 3D data, recent works like SAM3D~\cite{SAM3D} and SAMPro3D~\cite{sampro3d} focus on lifting robust 2D foundation models into the 3D domain. However, they require paired image and point cloud data, which limits their usability in real-world use cases with LiDAR only datasets or legacy datasets. \cite{SAL,Segment3D,POINT-SAM} employ SAM~\cite{SAM} to generate pseudo-labels provided RGB images and use them to train 3D models on indoor scenes, with Point-SAM~\cite{POINT-SAM} further addressing part-level segmentation. While this process helps generate significant amounts of training data, this pseudo-labeling process invariably introduces label noise. \shortname\ avoids this problem altogether by pooling publicly available datasets for training and establishes state-of-the-art performance on several unseen datasets.

\section{Method}
\label{sec:method}

\begin{figure}[tb]
    \centering
    \includegraphics[width=1\linewidth]{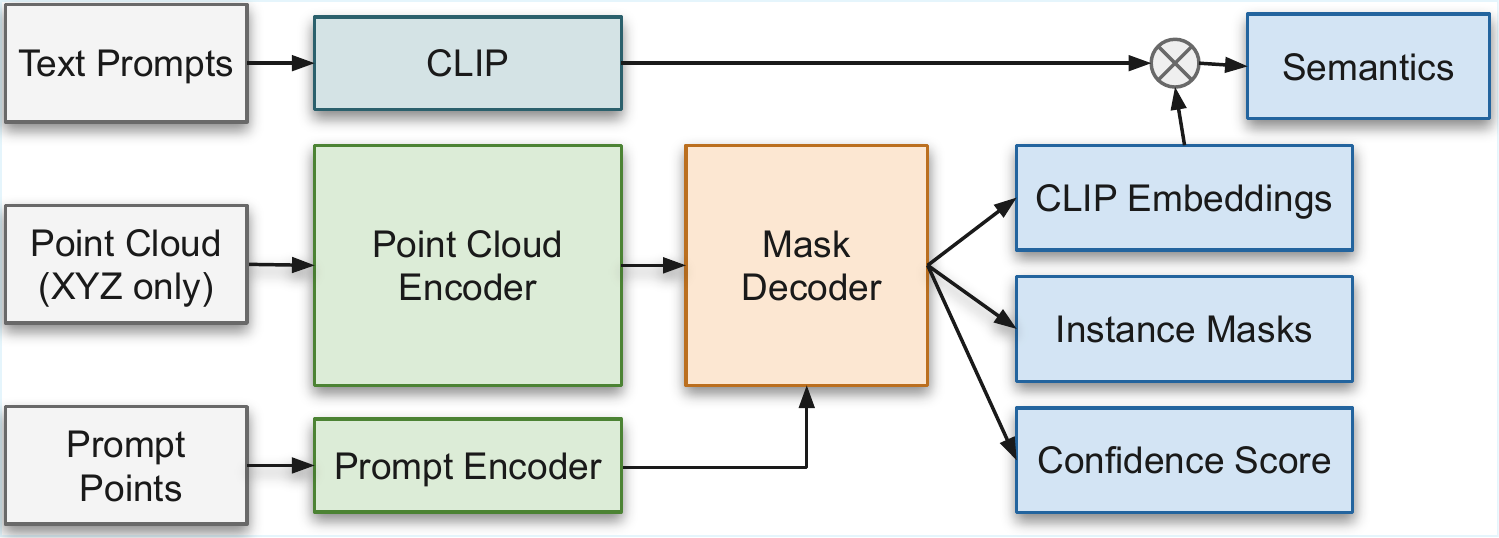} 
    \vspace{-1.2em}
    \caption{\textbf{Overview of SNAP.} SNAP encodes point clouds and prompts separately, then uses a Mask Decoder to generate segmentation masks. Text prompts are handled by matching CLIP embeddings with predicted mask embeddings for semantic classification.
    }
    \label{fig:model_arch}
    \vspace{-4pt}
\end{figure}

In this section, we present the overall framework of \shortname, which is organized into 4 parts: (1) Point Cloud Encoding, (2) Spatial-prompted Segmentation, (3) Text-prompted Segmentation, and (4) Training. \textbf{Fig.~\ref{fig:model_arch}} provides an overview of our approach.

\subsection{Point Cloud Encoding}
The input point cloud is represented by its XYZ coordinates, denoted as $\mathbf{P} = \{ \mathbf{p}_i \in \mathbb{R}^3 \}_{i=1}^N$, where $N$ is the number of points. We use the Point Transformer V3 (PTv3 \cite{ptv3}) to extract point-wise embeddings $\mathbf{F}_{\text{pc}} \in \mathbb{R}^{N\times D}$.
To support cross-domain generalization, we replace the regular batch normalization in PTv3 with domain normalization.

\begin{figure}[t]

    \centering
    \setlength{\tabcolsep}{1pt}
    \begin{small}
    \begin{tabular}{@{}ccc@{}}
         \includegraphics[width=0.3\linewidth]{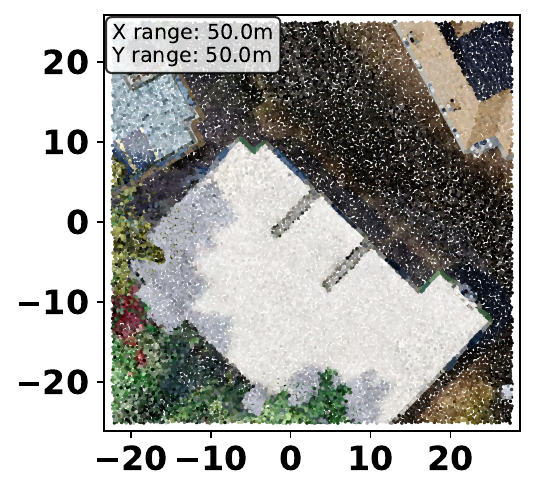} & 
         \includegraphics[width=0.38\linewidth]{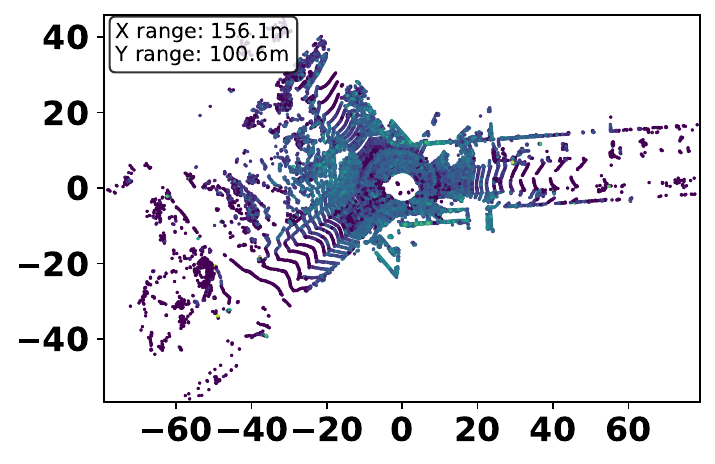} & 
         \includegraphics[width=0.3\linewidth]{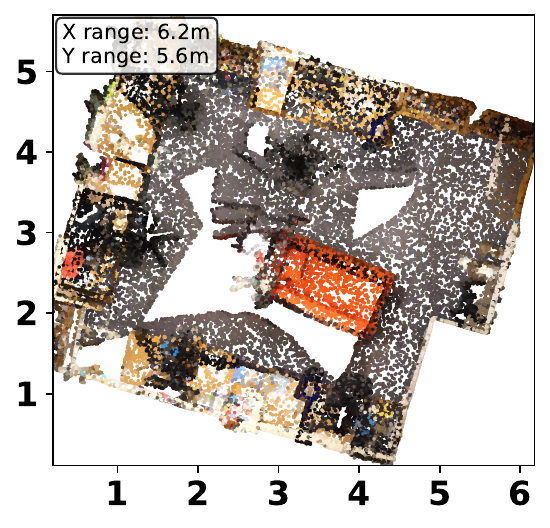} \\
         (a) STPLS3D & (b) KITTI & (c) ScanNet
    \end{tabular}
    \end{small}
    \vspace{-0.6em}
    \caption{\textbf{Point clouds from different domains vary significantly in their properties.} \textbf{(a)} STPLS3D provides dense point clouds from an aerial view using RGB photogrammetry in the 50m range, \textbf{(b)} KITTI provides lidar data in the 150m range, \textbf{(c)} ScanNet provides point clouds in the 10m range. 
    }
    \label{fig:scale_fig}
    \vspace{-15pt}
\end{figure}

\noindent\textbf{Domain Normalization for Multi-Dataset Training.}
\noindent Training a \textit{single} model on multiple point cloud datasets often leads to lower performance than \textit{multiple per-dataset} models due to negative transfer caused by significant distributional differences between various datasets~\cite{pdnorm_short}, as shown in \textbf{Fig.~\ref{fig:scale_fig}}. A straightforward approach to mitigate this is \textit{dataset-specific} normalization~\cite{pdnorm_short}, which learns unique normalization parameters for each dataset in the training set. While effective at addressing distributional differences among known datasets, this method presents practical challenges: users need to select appropriate normalization parameters at test time, which can be ambiguous when the source of a point cloud at test-time differs from the training datasets. Moreover, dataset-specific normalization may limit knowledge transfer between related datasets that could benefit from shared representations.

\begin{figure}[tb]
    \centering
    \setlength{\tabcolsep}{1pt}
    \begin{small}
    \begin{tabular}{@{}c|c@{}}
         \includegraphics[width=0.49\linewidth]{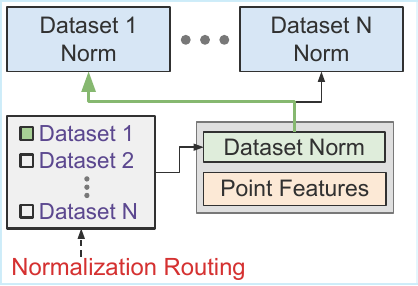} & 
         \includegraphics[width=0.49\linewidth]{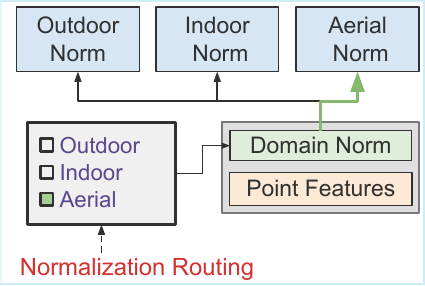} \\
         \small(a) Dataset Norm & \small(b) Domain Norm
    \end{tabular}
    \end{small}
    \vspace{-0.8em}
    \caption{\textbf{Dataset Norm vs Domain Norm}. Domain-norm simplifies the overall architecture and improves zero-shot generalization.
    }
    \label{fig:domain_norm}
    \vspace{-12pt}
\end{figure}

To address this limitation, we propose \textit{domain-specific} normalization, which groups datasets into broader \textit{domains} with similar statistical properties (\eg, indoor, outdoor, or aerial) and learns a separate set of normalization parameters for each domain. This strategy allows our model to effectively adapt to different data distributions while maintaining the flexibility to be applied to new datasets by identifying their general domain; a more intuitive decision than selecting specific dataset parameters, as shown in \textbf{Fig.~\ref{fig:domain_norm}}.

Formally, we denote the set of domains as $\mathbb{D} = \{ \mathcal{D}_j \}$. Domain  normalization for the $k$-th layer's activations from the $j$-th domain, $\bar{\mathbf{x}}^k_j$, is expressed as:

\vspace{-0.5em}{
\begin{equation}
\bar{\mathbf{x}}^k_j = \frac{\mathbf{x}^k_j - \mathbb{E}[\mathbf{x}^k_j]}{\sqrt{\text{Var}[\mathbf{x}^k_j]+\epsilon}} \cdot \gamma^k_j + \beta^k_j,
\end{equation}
}
where $\gamma^k_j$ and $\beta^k_j$ are learned domain-specific scale and shift parameters. Compared with standard batch and dataset-specific normalizations, this domain-specific design allows the model to effectively adapt the normalization to the distinct statistics of each domain while sharing the core model weights across all domains.

\subsection{Spatial-prompted Segmentation}
\label{sec:subsec_spatial_segmentation}
\noindent\textbf{Spatial Prompt Encoder.}
Given a spatial prompt point \( \mathbf{p}_{\text{sp}} \in \mathbb{R}^3 \), we identify its nearest neighbor in the point cloud:
$
i^* = \arg\min_i \left\| \mathbf{p}_{\text{sp}} - \mathbf{p}_i \right\|_2
$, and retrieve the corresponding point embedding,
$
\mathbf{f}_{\text{sem}} = \mathbf{f}_{i^*}
$, from the point-wise PTv3 embeddings $\mathbf{F}_{\text{pc}}$. 
We also compute positional encodings of the prompt point:
$
\mathbf{f}_{\text{pos}} = \phi(\mathbf{p}_{\text{sp}})
$, 
where \( \phi(\cdot) \) is the Fourier encoding function as in \cite{tancik2020fourier}.
The final spatial prompt embedding combines semantic features and the positional encodings, which help preserve spatial information:
$
\mathbf{F}_{\text{sp}} = \Phi_{\text{sp}}\left( \left[ \mathbf{f}_{\text{sem}} \, \| \, \mathbf{f}_{\text{pos}} \right] \right) \in \mathbb{R}^D
$
where \( \Phi_{\text{sp}}(\cdot) \) is a learned projection function, and \( [\cdot \, \| \, \cdot] \) is feature concatenation.
Given $P$ prompt points (clicks) on $M$ objects, we get $\mathbf{F}_{\text{sp}} \in \mathbb{R}^{M \times P \times D}$. 

\vspace{0.3em}
\noindent\textbf{Mask Decoder.}
The mask decoder module takes in two inputs: $\mathbf{F}_{\text{pc}} \in \mathbb{R}^{N \times D}$ for the point cloud embeddings, and $\mathbf{F}_{\text{sp}} \in \mathbb{R}^{M \times P \times D}$ for the prompt embeddings. 
Inspired by the design in SAM~\cite{SAM}, we additionally introduce \textit{three} task-specific learnable tokens per object. These tokens are responsible for predicting the final mask, its confidence score, and CLIP embeddings, respectively, resulting in $\mathbf{F}_{\text{sp}} \in \mathbb{R}^{M \times( P + 3) \times D}$.
Following the approach of SAM~\cite{SAM}, we first tile the point embeddings to match the prompt dimension:
$\tilde{\mathbf{F}}_{\text{pc}} \in \mathbb{R}^{M \times N \times D}$. The decoder then uses a series of transformer blocks to iteratively refine the embeddings. The process is designed to first incorporate contextual information from the point cloud into the prompt embeddings, and then use the refined prompts to condition the point cloud embeddings. Within each block, the updates occur in the following sequence: 
\begin{itemize}
    \item Prompt Self-Attention:
$
\mathbf{Z}_1 = \Psi_1(\mathbf{F}_{\text{sp}}, \mathbf{F}_{\text{sp}}).
$
    \item Prompt-to-point Cross-Attention:
$
\mathbf{Z}_2 = \Psi_2(\mathbf{Z}_1, \tilde{\mathbf{F}}_{\text{pc}}).
$
    \item Feedforward Network (FFN):
$
\mathbf{Z}_{\text{sp}} = \Phi(\mathbf{Z}_2).
$
    \item Point-to-Prompt Cross-Attention:
$
\mathbf{Z}_{\text{pc}} = \Psi_3(\tilde{\mathbf{F}}_{\text{pc}}, \mathbf{Z}_{\text{sp}}).
$
\end{itemize}
Here, each $\Psi_{( \cdot )}$ denotes a multi-head attention block.
Each of them takes two inputs, where the first denotes the query and the second indicates the key and value. 
$\Phi$ is a position-wise feedforward network. 
The final output of the entire decoder module is a set of refined prompt embeddings $\mathbf{Z}_{\text{sp}} \in \mathbb{R}^{M \times (P + 3) \times D}$ and conditioned point cloud embeddings $\mathbf{Z}_{\text{pc}} \in \mathbb{R}^{M \times N \times D}$.

To generate predictions, we first separate the learnable token embeddings from the refined prompt embeddings $\mathbf{Z}_{\text{sp}}$. Recall that during prompt encoding, we appended three learnable tokens 
to each object's prompt sequence. We extract the following embeddings accordingly: mask token embeddings $\mathbf{Z}_{\text{mask}} \in \mathbb{R}^{M \times 1 \times D}$, CLIP token embeddings $\mathbf{Z}_{\text{CLIP}} \in \mathbb{R}^{M \times 1 \times D}$, mask confidence score token embeddings $\mathbf{Z}_{\text{S}} \in \mathbb{R}^{M \times 1 \times D}$, and retain the original prompt point embeddings as auxiliary embeddings $\mathbf{Z}_{\text{aux}} \in \mathbb{R}^{M \times P \times D}$. 
The first three token embeddings are fed into dedicated prediction heads to generate mask, mask confidence scores, and CLIP embeddings, while the auxiliary embeddings are used for an additional supervision described in Sec.~\ref{subsec:training}.

\begin{itemize}
\item \noindent\textbf{Mask Head.} We pass the mask token embeddings $\mathbf{Z}_{\text{mask}} \in \mathbb{R}^{M \times 1 \times D}$ through a small MLP and then compute the dot product between the mask token embeddings and point cloud embeddings to get $M$ segmentation logits $\mathbf{L}_\text{mask}=\mathbf{Z}_\text{pc} \cdot (\mathbf{F}_\text{mask})^\top \in \mathbb{R}^{M \times N}$. The final segmentation masks can then be obtained by applying a sigmoid function $\sigma$ to the logits: $\mathcal{M}_{\text{mask}}=\sigma(\mathbf{L}_{\text{mask}})$.

\vspace{0.3em}
\item \noindent\textbf{Confidence Score Head.} Mask confidence score predictions can be similarly obtained by passing the mask score token embeddings through a MLP: $\mathbf{L}_\text{S}=\Phi_\text{S}(\mathbf{Z}_\text{S})$ and applying a sigmoid function $\sigma$: $\mathcal{S}=\sigma(\mathbf{L}_{\text{S}})$.

\vspace{0.3em}
\item \noindent\textbf{CLIP Embedding Head.} Motivated by SAL~\cite{SAL}, we learn to predict CLIP embeddings for each mask using a MLP $\Phi_\text{CLIP}$:
$\mathbf{L}_\text{CLIP}=\Phi_\text{CLIP}(\mathbf{Z}_\text{CLIP})$.
We show in \textbf{Tab.~\ref{table:ablation_loss_functions}} that this also improves segmentation accuracy.
\end{itemize}

\subsection{Text-prompted Segmentation}
\label{sec:subsec_text_prompted_segmentation}
Our text-prompted segmentation pipeline consists of two stages: (1) automatic generation of prompt points to produce mask proposals via our spatial-prompted segmentation module, and (2) matching of predicted CLIP embeddings from each mask to the input textual prompt.

\vspace{0.3em}
\noindent\textbf{Automatic Prompt Points Generation.}
To comprehensively segment the input point cloud without manual prompts, we employ an iterative coarse-to-fine strategy. Starting with coarse voxelization, we use voxel centers as prompt points to generate initial mask proposals. We then identify unsegmented regions and generate new prompts using progressively smaller voxel sizes in the unsegmented regions, continuing for a fixed number of iterations. Finally, non-maximum suppression removes overlapping masks. This approach ensures comprehensive coverage while avoiding redundant computation in well-segmented regions, striking a better balance between accuracy and efficiency compared to uniform grid sampling as in the 2D counterpart of SAM~\cite{SAM}. 
See \S~\ref{sec:auto_prompt_generation} for more details.

\vspace{0.3em}
\noindent\textbf{Text Prompt Encoder and Matching.}
To classify the generated mask proposals, we encode input text prompts using the CLIP text encoder~\cite{clip}. During training, we use category names as text prompts to train \shortname. Specifically, we wrap category names in full sentences (\eg, \textbf{\texttt{a photo of a {class\_name}}}).
At inference, the generated mask proposals are matched to text queries by comparing their predicted CLIP embeddings with encoded text prompts. This enables both panoptic segmentation using predefined vocabularies and open-vocabulary segmentation for novel object classes.

\subsection{Training}
\label{subsec:training}
\noindent\textbf{Click Sampling Strategy.} Following AGILE3D \cite{AGILE3D} and Interactive4D \cite{interactive4D}, we use an iterative click sampling strategy to simulate user clicks. Unlike their computationally intensive process of calculating and then ranking error regions for additional clicks, we adopt a simpler approach by randomly sampling clicks from the unsegmented regions of each object.

\vspace{0.3em}
\noindent\textbf{Training Losses.} 
In line with prior works \cite{SAM, AGILE3D, interactive4D}, we supervise our mask predictions using the ground truth annotated instance segmentation labels using Focal~\cite{lin2018focallossdenseobject} and Dice~\cite{diceloss} loss. Following ~\cite{interactive4D}, we also incorporate weights on these losses to modulate them based on proximity to the user clicks. Additionally, to encourage each click to independently yield a plausible mask prediction, we use an auxiliary mask loss ($\mathcal{L}_\text{aux}$), which adopts the auxiliary mask token embeddings (as defined in Sec.~\ref{sec:subsec_spatial_segmentation}) to generate additional mask predictions.
To improve the reliability of mask confidence score estimation, we also introduce a score prediction loss ($\mathcal{L}_\text{score}$). Specifically, the target score for each mask is defined as the Intersection-over-Union (IoU) between the predicted mask and its corresponding ground-truth mask. To supervise the predicted class labels, we use $\mathcal{L}_\text{text}$ to penalize incorrect alignment between the CLIP embeddings of predicted text tokens for each mask and the class vocabulary, using a cosine distance loss. 
Our overall loss function can be written as:
\vspace{-0.5em}{
\begin{align}
\mathcal{L}_\text{\shortname} &=\mathcal{L}_\text{focal} + \mathcal{L}_\text{dice} + \mathcal{L}_\text{aux} + \mathcal{L}_\text{score} + \mathcal{L}_\text{text}.
\end{align}
}
See \S~\ref{sec:loss_details} for additional details of each loss term.

\section{Experiments}

\subsection{Experimental Setup}
\noindent\textbf{Datasets.} We train \shortname\ on 7 diverse datasets with ground-truth instance segmentation labels, which span three domains: (i)~\textit{indoor} scenes from ScanNet~\cite{scannet} and HM3D~\cite{ramakrishnan2021hm3d,yadav2022habitat}; (ii)~\textit{outdoor} driving sequences from SemanticKITTI~\cite{semantickitti,kitti}, nuScenes~\cite{nuscenes}, and Pandaset~\cite{pandaset}; and (iii)~\textit{aerial} point clouds from STPLS3D~\cite{stpls3d} and DALES~\cite{varney2020dales,dales_instance}. We evaluate zero-shot performance on held-out datasets from each domain: S3DIS~\cite{s3dis} (full and crops), ScanNet++~\cite{scannetpp}, and Matterport3D~\cite{Matterport3D} for indoor; Waymo~\cite{Sun_2020_CVPR} and KITTI-360~\cite{kitti360} (full, crops, and single scan) for outdoor; and UrbanBIS~\cite{UrbanBIS} for aerial. 
See \S~\ref{sec:dataset_details}.

\vspace{0.3em}
\noindent\textbf{Evaluation Metrics.}
Following conventions from \cite{interactive4D, AGILE3D, InterObject3D, POINT-SAM}, we evaluate spatial-prompted segmentation using IoU@$k$, the average intersection over union (IoU) achieved with $k$ clicks per object across all objects. 
To assess object category prediction capabilities, we use the mean Average Precision (mAP) metric following ~\cite{openmask3d, huang2024openins3dsnaplookup3d, Peng2023OpenScene} and for panoptic segmentation, we use panoptic quality (PQ), segmentation quality (SQ), and recognition quality (RQ) as done in ~\cite{SAL}.

\begin{table}[t]
\centering
\small
\caption{\textbf{In-distribution interactive point cloud segmentation.} 
\shortname-\textit{dataset} refers to the model trained exclusively on the specified dataset. $\dagger$ denotes that the method is evaluated by us.
}
\label{table:spatial_in_distribution}
\vspace{-0.8em}
\begin{tabular}{lccccc}
\toprule
& \multicolumn{5}{c}{${\text{IoU}}@k \uparrow$} \\
\cmidrule(r){2-6}
\textbf{Method} & @1 & @2 & @3 & @5 & @10 \\
\midrule
\multicolumn{6}{c}{Trained and evaluated on SemanticKITTI}  \\ 
\midrule

AGILE3D~\cite{AGILE3D} & 53.1 & 63.7 & 70 & 76.7 & 83.3 \\
Interactive4D~\cite{interactive4D} & 67.5 & 73.9 & 78.3 & 83.4 & 88.2 \\
SNAP - KITTI & 68.1	& 75.9 & 80.1 &	84.5 & 88.7  \\
SNAP - C & \textbf{71.5} & \textbf{78.1} & \textbf{81.0} & \textbf{86.0} & \textbf{90.1} \\
\midrule
\multicolumn{6}{c}{Trained and evaluated on ScanNet20}  \\ 
\midrule
InterObject3D~\cite{InterObject3D} & 40.8 & 55.9 & 63.9 & 72.4 & 79.9 \\
AGILE3D~\cite{AGILE3D} & 63.3 & 70.9 & 75.4 & 79.9 & 83.7 \\
Point-SAM$^\dagger$~\cite{POINT-SAM} & 52.7 & 69.6 & 75.9 & 80.6 & 83.3 \\
SNAP - SN & \textbf{68.6} & 74.2 & 78.4 & 82.1 & 84.6 \\
SNAP - C & 67.7 & \textbf{74.7} & \textbf{78.5} & \textbf{82.3} & \textbf{85.5} \\
\bottomrule
\end{tabular}
\vspace{-10pt}
\end{table}

\vspace{0.3em}
\noindent\textbf{Model Variants}. 
\shortname\ operates on XYZ coordinates only, without relying on any additional per-point attributes (\eg, color, normals, intensity), as such features are not consistently available across datasets. 
As shown in Sec.~\ref{sec:ablation}, using only XYZ coordinates achieves performance comparable to models that leverage all available modalities, while avoiding the dependency on dataset-specific properties. 
We evaluate 6 \shortname\ variants to provide comprehensive comparisons. First, we train dataset-specific models: \shortname-KITTI on SemanticKITTI~\cite{semantickitti} and \shortname-SN on ScanNet~\cite{scannet}. Second, we train domain-specific models that leverage all available in-domain datasets: \shortname-Indoor, \shortname-Outdoor, and \shortname-Aerial. Finally, \shortname-C represents our complete model trained across all datasets, serving as our most generalizable variant. In Sec.~\ref{sec:expt_class_aware}, \shortname-C@auto refers to the automatic prompt points generation setting, in which the model is provided solely with the point cloud to enable a fair comparison with baseline methods.

\subsection{Spatial-prompted Interactive Segmentation}
\label{sec:expt_class_agnostic}
To fairly evaluate \shortname\ against other purely interactive segmentation baselines like \cite{AGILE3D,POINT-SAM,interactive4D}, we first evaluate the predicted instance mask quality without taking the class-prediction into account (class-agnostic), as typically done in prior methods on interactive point cloud segmentation. 

\vspace{0.3em}
\noindent\textbf{In-distribution Evaluation.}
In \textbf{Tab.~\ref{table:spatial_in_distribution}} we compare \shortname's performance on the SemanticKITTI~\cite{semantickitti} and ScanNet~\cite{scannet} datasets, where SNAP outperforms all prior baselines on both datasets. Notably, while AGILE3D~\cite{AGILE3D} needs to be trained separately for both datasets, \shortname\ is evaluated with the same set of parameters for both datasets. This versatility makes \shortname\ a more practical segmentation tool, which can thus be evaluated on a broader set of datasets.

\vspace{0.3em}
\noindent\textbf{Zero-Shot Evaluation.}
To test the generalization capabilities of the \shortname, we evaluate on 9 \textit{unseen} benchmarks covering indoor, outdoor, and aerial domains. As shown in \textbf{Tab.~\ref{table:spatial_zero_shot}}, \shortname-C outperforms the baseline AGILE3D~\cite{AGILE3D}, Point-SAM~\cite{POINT-SAM} and Interactive4D~\cite{interactive4D} on 8 out of 9 benchmarks. Particularly in the 1-Click experiments, \shortname-C provides a \textbf{20.6\%} average improvement across all benchmarks with a single set of parameters. This positions \shortname-C  as a practical, general-purpose tool for interactive segmentation, addressing the key limitation of existing domain-specific approaches.

\begin{table}[tb]
\centering
\small
\caption{\textbf{Zero-shot interactive point cloud segmentation}. We compare \shortname\ with the current methods for interactive segmentation across different domains on several unseen datasets. 
$^\dagger$ denotes that the methods are evaluated by us.}
\vspace{-0.8em}
\setlength{\tabcolsep}{4pt}
\begin{tabular}{l|l l c c c}
\toprule
& \multirow{2}{*}{\makecell{\textbf{Dataset}}} & \multirow{2}{*}{\makecell{\textbf{Method}}} & \multicolumn{3}{c}{${\text{IoU}}@k \uparrow$} \\
\cmidrule{4-6}
& & & \textbf{@1} & \textbf{@3} & \textbf{@5} \\
\midrule

\multirow{13}{*}{\rotatebox{90}{\textbf{Indoor}}}
& \multirow{3}{*}{ScanNet++} & Point-SAM$^\dagger$~\cite{POINT-SAM} & 28.6 & 56.3 & 62.9 \\
& & SNAP - C & \textbf{52.0} & \textbf{67.3} & \textbf{73.2} \\
& & \textit{\footnotesize $\Delta$} & \textit{\textcolor{teal}{\footnotesize +23.4}} & \textit{\textcolor{teal}{\footnotesize +11.0}} & \textit{\textcolor{teal}{\footnotesize +10.3}} \\
\cmidrule{2-6}

& \multirow{3}{*}{Matterport3D} & Point-SAM$^\dagger$~\cite{POINT-SAM} & 41.1 & 67.2 & 73.7 \\
& & SNAP - C & \textbf{52.6} & \textbf{69.6} & \textbf{75.2} \\
& & \textit{\footnotesize $\Delta$} & \textit{\textcolor{teal}{\footnotesize +11.5}} & \textit{\textcolor{teal}{\footnotesize +2.4}} & \textit{\textcolor{teal}{\footnotesize +1.5}} \\
\cmidrule{2-6}

& \multirow{4}{*}{\makecell{S3DIS\\(crops)}} & AGILE3D~\cite{AGILE3D} & \textbf{58.7} & \textbf{77.4} & 83.6 \\
& & Point-SAM$^\dagger$~\cite{POINT-SAM} & 45.9 & 77.6 & \textbf{84.6} \\
& & SNAP - C & 56.6 & 73.8 & 80.9 \\
& & \textit{\footnotesize $\Delta$} & \textit{\textcolor{darkred}{\footnotesize -2.1}} & \textit{\textcolor{darkred}{\footnotesize -3.6}} & \textit{\textcolor{darkred}{\footnotesize -3.7}} \\
\cmidrule{2-6}

& \multirow{3}{*}{\makecell{S3DIS\\(full)}} & Point-SAM$^\dagger$~\cite{POINT-SAM} & 35.6 & 68.0 & 76.3 \\
& & SNAP - C & \textbf{53.6} & \textbf{71.1} & \textbf{77.6} \\
& & \textit{\footnotesize $\Delta$} & \textit{\textcolor{teal}{\footnotesize +18.0}} & \textit{\textcolor{teal}{\footnotesize +3.1}} & \textit{\textcolor{teal}{\footnotesize +1.3}} \\
\midrule

\multirow{15}{*}{\rotatebox{90}{\textbf{Outdoor}}}
& \multirow{4}{*}{Waymo} & Point-SAM$^\dagger$~\cite{POINT-SAM} & 12.8 & 43.0 & 53.1 \\
& & Interactive 4D$^\dagger$~\cite{interactive4D} & 7.2 & 7.3 & 7.5 \\
& & SNAP - C & \textbf{69.8} & \textbf{82.3} & \textbf{86.6} \\
& & \textit{\footnotesize $\Delta$} & \textit{\textcolor{teal}{\footnotesize +57.0}} & \textit{\textcolor{teal}{\footnotesize +39.3}} & \textit{\textcolor{teal}{\footnotesize +33.5}} \\
\cmidrule{2-6}

& \multirow{3}{*}{\makecell{KITTI-360\\(full)}} & Point-SAM$^\dagger$~\cite{POINT-SAM} & 6.8 & 22.7 & 28.1 \\
& & SNAP - C & \textbf{23.1} & \textbf{40.1} & \textbf{48.1} \\
& & \textit{\footnotesize $\Delta$} & \textit{\textcolor{teal}{\footnotesize +16.3}} & \textit{\textcolor{teal}{\footnotesize +17.4}} & \textit{\textcolor{teal}{\footnotesize +20.0}} \\
\cmidrule{2-6}

& \multirow{4}{*}{\makecell{KITTI-360\\Single Scan}} & AGILE3D~\cite{AGILE3D} & 36.3 & 47.3 & 53.5 \\
& & Interactive 4D~\cite{interactive4D} & 47.7 & 59.4 & 64.1 \\
& & SNAP - C & \textbf{60.4} & \textbf{64.6} & \textbf{67.7} \\
& & \textit{\footnotesize $\Delta$} & \textit{\textcolor{teal}{\footnotesize +12.7}} & \textit{\textcolor{teal}{\footnotesize +5.2}} & \textit{\textcolor{teal}{\footnotesize +3.6}} \\
\cmidrule{2-6}

& \multirow{4}{*}{\makecell{KITTI-360\\(crops)}} & AGILE3D~\cite{AGILE3D} & 34.8 & 42.7 & 44.4 \\
& & Point-SAM~\cite{POINT-SAM} & 49.4 & 74.4 & \textbf{81.7} \\
& & SNAP - C & \textbf{65.6} & \textbf{76.1} & 80.0 \\
& & \textit{\footnotesize $\Delta$} & \textit{\textcolor{teal}{\footnotesize +16.2}} & \textit{\textcolor{teal}{\footnotesize +1.7}} & \textit{\textcolor{darkred}{\footnotesize -1.7}} \\
\midrule

\multirow{3}{*}{\rotatebox{90}{\textbf{Aerial}}}
& \multirow{3}{*}{UrbanBIS} & Point-SAM$^\dagger$~\cite{POINT-SAM} & 39.3 & 79.1 & 89.4 \\
& & SNAP - C & \textbf{71.6} & \textbf{86.2} & \textbf{90.2} \\
& & \textit{\footnotesize $\Delta$} & \textit{\textcolor{teal}{\footnotesize +32.3}} & \textit{\textcolor{teal}{\footnotesize +7.1}} & \textit{\textcolor{teal}{\footnotesize +0.8}} \\
\bottomrule
\end{tabular}
\label{table:spatial_zero_shot}
\vspace{-12pt}
\end{table}

\subsection{Text-prompted Segmentation}
\label{sec:expt_class_aware}
For text-prompted segmentation, we evaluate \shortname's effectiveness for panoptic segmentation 
and open-vocabulary segmentation.
Our primary evaluation targets \shortname-C@auto; we additionally report \shortname-C@1 Click as an upper-bound reference, with the click sampled from the \emph{ground-truth} masks to simulate ideal user input. \shortname-auto takes only the point cloud as input and outputs panoptic segmentation masks like the baselines we compare against.

\vspace{0.3em}
\noindent\textbf{Panoptic Segmentation}.
\label{sec:expt_class_aware_panoptic}
SAL~\cite{SAL} pioneered automated panoptic segmentation on outdoor datasets, making it our primary baseline for outdoor scenes.
As shown in \textbf{Tab.~\ref{table:panoptic_seg}}, 
\shortname-C@auto, which has the equivalent setting to SAL with only point clouds as input, performs robustly against it and improves PQ score by 3.4 points on SemanticKITTI while remaining comparable on nuScenes (37.9 PQ vs 38.4 from SAL).
With single-click guidance, \shortname-C@1 Click achieves even better results on both datasets.

\begin{table}[tb]
\centering
\small
\caption{\textbf{Panoptic segmentation on outdoor Lidar datasets.} We compare the \shortname-C @ auto variant with SAL~\cite{SAL} for panoptic segmentation on outdoor lidar datasets. We also show \shortname-C @ 1 Click results as a reference  that represents the upper-bound for \shortname-C @ auto. \uline{Note that \shortname-C @ auto represents the equivalent setting to SAL~\cite{SAL} since the only input is raw point clouds.} 
}
\vspace{-0.8em}
\label{table:panoptic_seg}
\begin{tabular}{l|c c c c c}
\toprule
Method & PQ & RQ & SQ & PQ$_{\mathrm{Th}}$ & PQ$_{\mathrm{St}}$ \\
\midrule
\multicolumn{6}{c}{Evaluation $\rightarrow$ SemanticKITTI }  \\ 
\midrule
SAL~\cite{SAL} & 24.8 & 32.3 & 66.8 & 17.4 & 30.2 \\
SNAP - C @ auto & \textbf{28.2} & \textbf{34.1} & \textbf{78.6} & \textbf{20.7} & \textbf{34.4} \\
\rowcolor{gray!30}SNAP - C @ 1 Click & 40.1 & 47.3 & 81.6 & 30.7 & 46.9 \\

\midrule
\multicolumn{6}{c}{Evaluation $\rightarrow$ nuScenes}  \\ 
\midrule
SAL~\cite{SAL} & \textbf{38.4} & \textbf{47.8} & 77.2 & \textbf{47.5} & 29.2 \\
SNAP - C @ auto & 37.9 & 47.1 & \textbf{82.2} & 33.1 & \textbf{52.4} \\
\rowcolor{gray!30}SNAP - C @ 1 Click & 47.2 & 56.1 & 83.9 & 40.3 & 56.2 \\

\bottomrule
\end{tabular}
\vspace{-12pt}
\end{table}

\vspace{0.3em}
\noindent\textbf{Open-Vocabulary Segmentation.}
There exist limited number of approaches for direct alignment between 3D point clouds and CLIP~\cite{clip} text embeddings without intermediate image representations, we therefore evaluate against three non-interactive instance segmentation baselines.
First, OpenScene (3D Distill)~\cite{Peng2023OpenScene} represents the most comparable approach to ours as it only uses the predicted per point CLIP embeddings during inference without requiring images.
Second, OpenIns3D~\cite{huang2024openins3dsnaplookup3d} generates class-agnostic instance masks and render multiview images of the input point cloud for CLIP~\cite{clip} based segmentation.
Third, OpenMask3D~\cite{openmask3d} and SAI3D~\cite{yin2024sai3dsegmentinstance3d} leverage the original 2D images from the dataset to perform open-vocabulary instance segmentation.

\begin{figure}
    \centering
    \begin{tabular}{c@{\hskip 5pt}c}
        \setlength{\fboxsep}{0pt}  %
        \setlength{\fboxrule}{1pt} %
        \fbox{\includegraphics[width=0.2\textwidth]{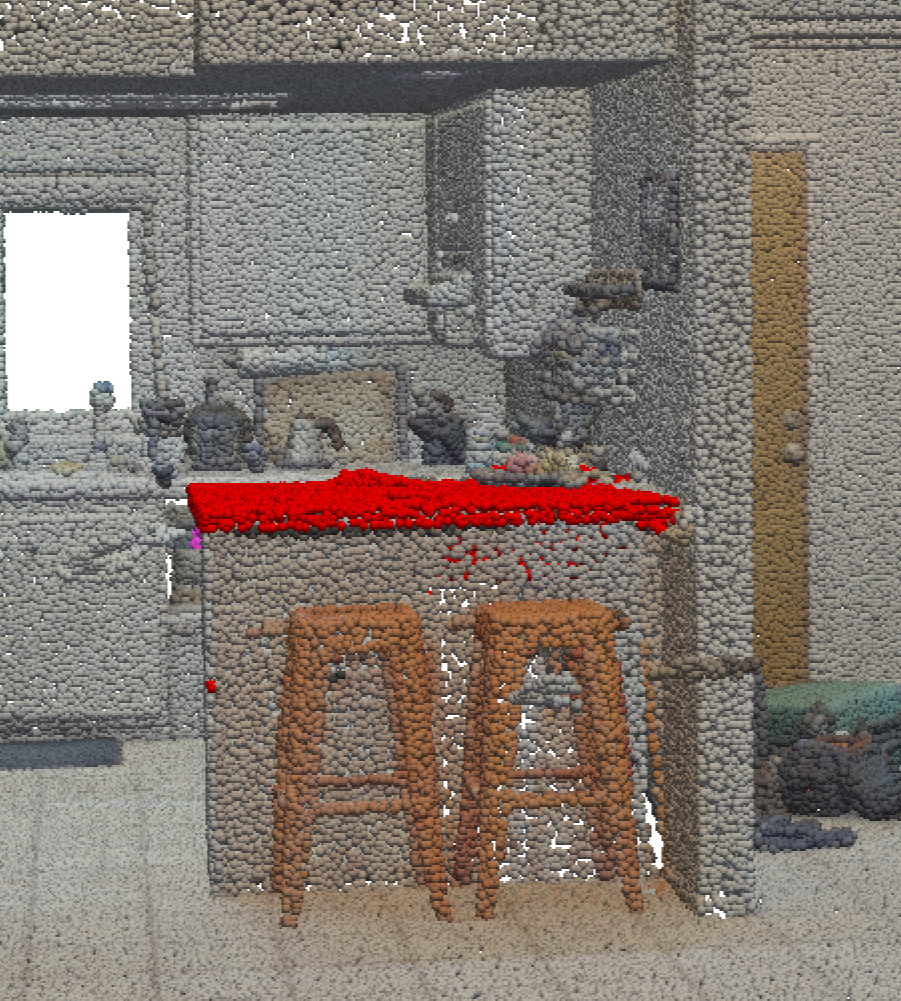}}
        & 
        \setlength{\fboxsep}{0pt}  %
        \setlength{\fboxrule}{1pt} %
        \fbox{\includegraphics[width=0.2\textwidth]{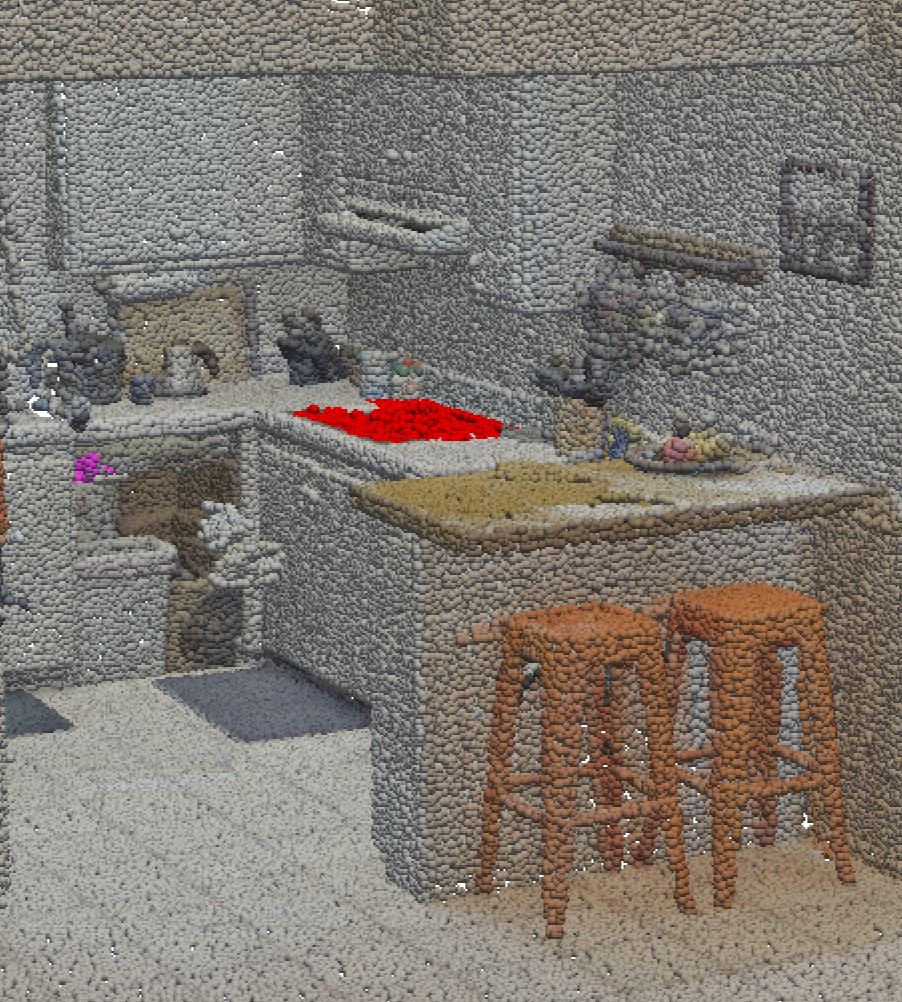}}\\
        \footnotesize{\textit{``Segment bar table.''}}
        & \footnotesize{\textit{``Segment cooking stove.''}}\\
    \end{tabular}
    \vspace{-0.8em}
    \caption{\textbf{Qualitative segmentation results of open-set scene understanding on the ScanNet++ Dataset.} Given a text prompt in the format of \textquotedblleft \textit{Segment \{open-set vocabulary\}}\textquotedblright, our \shortname~model finds the corresponding masks {\textcolor{red}{\raisebox{-0.1em}{\rule{0.8em}{0.8em}}}} in the scenes.
    }
    \label{fig:open-set}
    \vspace{-15pt}
\end{figure}

Results are summarized in \textbf{Tab.~\ref{table:open_vocab}}, where baseline methods are trained on ScanNet200~\cite{scannet}. 
\shortname-C demonstrates strong performance across 3 zero-shot datasets without using image embeddings.
On Matterport3D~\cite{Matterport3D}, \shortname-C@auto achieves 16.5 AP, substantially outperforming baselines which rely on 
images and CLIP image embeddings. This is largely attributed to the large overlap between semantic classes of Matterport3D~\cite{Matterport3D} and ScanNet200~\cite{scannet}. On Replica~\cite{straub2019replicadatasetdigitalreplica} and S3DIS~\cite{s3dis}, \shortname-C@auto shows competitive results against baselines that have access to visual information and surpasses the image-free OpenScene(3D Distill)~\cite{Peng2023OpenScene}. We also include qualitative results on the ScanNet++~\cite{scannetpp} dataset in \textbf{Fig.~\ref{fig:open-set}}.

In summary, unlike methods that need RGB images or pre-computed CLIP embeddings, SNAP-C@auto performs open-vocabulary segmentation directly on point clouds and achieves competitive results against image-based methods. The image-free approach also offers practical advantages for handling LiDAR-only datasets, synthetic data, or legacy datasets where corresponding images were never collected. 

\begin{table}[tb]
\centering
\small
\caption{\textbf{Open-Vocabulary Segmentation}. We evaluate the \shortname-C @ auto variant against image-based and image-free open vocabulary segmentation models. We also show \shortname-C @ 1 Click results as a reference that represents the upper-bound for \shortname-C @ auto. \uline{Note that \shortname-C @ auto represents the equivalent setting to baseline methods since the only input is raw point clouds.} $^{\dagger}$Uses RGB images, $^{\ddagger}$Uses CLIP image embeddings from rendered point cloud views.
}
\vspace{-0.8em}
\label{table:open_vocab}
\setlength{\tabcolsep}{2.8pt}
\begin{tabular}{l|c c c c c}
\toprule
Model & \makecell{Uses\\Img$^{\dagger}$} & \makecell{CLIP\\IE.$^{\ddagger}$} & AP & AP$_{50}$ & AP$_{25}$ \\

\midrule
\multicolumn{6}{c}{Zero Shot - Matterport3D}  \\ 
\midrule
OpenMask3D~\cite{openmask3d}          &	\checkmark & \checkmark  & 7.7	& 13.9 & 20.3 \\
SAI3D~\cite{yin2024sai3dsegmentinstance3d}               &	\checkmark & \checkmark  & 8.9	& 15.3 & 20.9 \\ 
\softmidrule
SNAP - C @ auto	    &	\xmark     & \xmark      & \textbf{16.5} & \textbf{25.2} & \textbf{30.7} \\
\rowcolor{gray!30}SNAP - C  @ 1 click &	\xmark     & \xmark	     & 18.3	& 28.2 & 34.7 \\

\midrule
\multicolumn{6}{c}{Zero Shot - Replica}  \\ 
\midrule
OpenMask3D~\cite{openmask3d} & \checkmark & \checkmark & 13.1 & 18.4 & 24.2 \\
OpenIns3D~\cite{huang2024openins3dsnaplookup3d} & \xmark & \checkmark & 13.6 & 18.0 & 19.7 \\
\softmidrule
OpenScene(3D Distill)~\cite{Peng2023OpenScene} & \xmark & \xmark & 8.2 & 10.5 & 12.6 \\
SNAP - C @ auto & \xmark & \xmark & \textbf{10.1} & \textbf{11.7} & \textbf{13.8} \\
\rowcolor{gray!30}SNAP - C @ 1 click & \xmark & \xmark & 11.7 & 14.5 & 15.4 \\

\midrule
\multicolumn{6}{c}{Zero Shot - S3DIS}  \\ 
\midrule
OpenIns3D~\cite{huang2024openins3dsnaplookup3d} & \xmark & \checkmark & 21.1 & 28.3 & 29.5 \\
\softmidrule
OpenScene(3D Distill)~\cite{Peng2023OpenScene} & \xmark & \xmark & 15.2 & 21.5  & 23.7 \\
SNAP - C @ auto & \xmark & \xmark & \textbf{16.1} & \textbf{23.8} & \textbf{30.9}  \\
\rowcolor{gray!30}SNAP - C @ 1 click & \xmark & \xmark & 17.6	& 25.5 & 33.6 \\

\bottomrule
\end{tabular}
\end{table}

\subsection{Ablation Studies}
\label{sec:ablation}

\begin{table}[t]
\centering
\small
\caption{\textbf{Input Properties Ablations.} X - XYZ coordinates, C - Color, N- Normals, S-Strength / Lidar intensity. - marks property data not available.}
\label{table:ablation_input_modalities}
\vspace{-0.8em}
\setlength{\tabcolsep}{4pt}
\begin{tabular}{l|c c c c c c c}
\toprule
Dataset & X & C & N & S & IoU@1 & IoU@5 & IoU@10 \\
\midrule
SemanticKITTI & \checkmark & - & - & \checkmark & 68.2 & 84.5 & 88.7 \\
SemanticKITTI & \checkmark & - & - & \xmark & 68.1 & 84.5 & 88.7 \\
\softmidrule
ScanNet20        & \checkmark & \checkmark & \checkmark & - & 69.6 & 83.0 & 85.5 \\
ScanNet20        & \checkmark & \checkmark & \xmark & - & 68.8 & 82.1 & 84.6 \\
ScanNet20        & \checkmark & \xmark & \checkmark & - & 69.3 & 83.0 & 85.5 \\
ScanNet20        & \checkmark & \xmark & \xmark & - & 68.6 & 82.1 & 84.6 \\
\softmidrule
STPLS3D        & \checkmark & \checkmark & - & - & 67.9 & 78.9 & 83.9 \\
STPLS3D        & \checkmark & \xmark & - & - & 67.2 & 78.9 & 83.9 \\
\bottomrule
\end{tabular}
\vspace{-5pt}
\end{table}

\begin{table}[t]
\centering
\small
\caption{\textbf{Loss Function Ablations.} We report ${\text{IoU}}@k$ with different combinations of Focal, Dice, Auxiliary, Weighted, and Text losses on the ScanNet20 dataset.
}
\vspace{-0.8em}
\label{table:ablation_loss_functions}
\setlength{\tabcolsep}{3pt}
\begin{tabular}{c c c c c|c c c}
\toprule
\multicolumn{5}{c}{Component} & \multicolumn{3}{c}{${\text{IoU}}@k \uparrow$} \\
\cmidrule{1-5} \cmidrule{6-8}
Dice & Focal & Aux & Weighted & Text & @1 & @5 & @10 \\
\midrule
\checkmark & \checkmark &  &  &  & 66.7 & 80.5 & 82.4 \\
\checkmark & \checkmark & \checkmark &  &  & 67.3 & 81.0 & 83.2 \\
\checkmark & \checkmark & \checkmark & \checkmark &  & 67.5 & 81.7 & 83.7 \\
\checkmark & \checkmark & \checkmark & \checkmark & \checkmark & \textbf{68.6} & \textbf{82.1} & \textbf{84.6} \\
\bottomrule
\end{tabular}
\vspace{-10pt}
\end{table}

\begin{table*}[tb]
\centering
\small
\caption{\textbf{Domain Normalization Ablation.} 
We compare 3 normalization strategies for training a multi-dataset model: Batch Norm, Dataset Norm, and Domain Norm. The applied normalization for Dataset Norm and Domain Norm is provided in \textcolor{teal}{\textit{teal}} beneath each result.
}
\vspace{-0.8em}
\label{table:ablation_domain_norm}
\begin{tabular}{l|cc cc cc | cc cc cc}
\toprule
\multirow{4}{*}{Model} & \multicolumn{12}{c}{${\text{IoU}}@k \uparrow$ 
} \\
\cmidrule{2-13}
\addlinespace[-2.5pt]
& \multicolumn{6}{c}{\cellcolor{mediumblue!20}In-Distribution} & \multicolumn{6}{c}{\cellcolor{green!20}Zero-Shot} \\
\addlinespace[-2pt]
\cmidrule{2-13}
& \multicolumn{2}{c}{SemanticKITTI} & \multicolumn{2}{c}{ScanNet20} & \multicolumn{2}{c}{STPLS3D} & \multicolumn{2}{c}{Waymo} & \multicolumn{2}{c}{ScanNet++} & \multicolumn{2}{c}{UrbanBIS} \\
\cmidrule(lr){2-3} \cmidrule(lr){4-5} \cmidrule(lr){6-7} \cmidrule(lr){8-9} \cmidrule(lr){10-11} \cmidrule(lr){12-13}
 & @1 & @5 & @1 & @5 & @1 & @5 & @1 & @5 & @1 & @5 & @1 & @5 \\
\midrule
Batch Norm   & 64.3 & 83.9 & 63.5 & 81.2 & 54.2 & 68.4 & 52.1 & 64.3 & 38.1 & 56.2 & 31.3 & 61.1 \\
\softmidrule
Dataset Norm & \textbf{72.0} & \textbf{86.1} & \textbf{69.5} & \textbf{ 83.8} & \textbf{67.9} & 79.0 & 56.2 & 76.3 & 48.5 & 67.3 & 61.8 & 77.8 \\

\textit{\textcolor{teal}{\footnotesize Norm used}} & \multicolumn{2}{c}{\textit{\textcolor{teal}{\footnotesize KITTI}}} & \multicolumn{2}{c}{\textit{\textcolor{teal}{\footnotesize ScanNet}}} & \multicolumn{2}{c}{\textit{\textcolor{teal}{\footnotesize STPLS3D}}} & \multicolumn{2}{c}{\textit{\textcolor{teal}{\footnotesize KITTI}}} & \multicolumn{2}{c}{\textit{\textcolor{teal}{\footnotesize ScanNet}}} & \multicolumn{2}{c}{\textit{\textcolor{teal}{\footnotesize STPLS3D}}}\\

\softmidrule
Domain Norm  & 71.5 & 86.0 & 67.7 & 82.3 & 67.8 & \textbf{80.4} & \textbf{69.8} & \textbf{86.6} & \textbf{52} & \textbf{73.2} & \textbf{71.6} & \textbf{90.2} \\

\textit{\textcolor{teal}{\footnotesize Norm used}} & \multicolumn{2}{c}{\textit{\textcolor{teal}{\footnotesize Outdoor}}} & \multicolumn{2}{c}{\textit{\textcolor{teal}{\footnotesize Indoor}}} & \multicolumn{2}{c}{\textit{\textcolor{teal}{\footnotesize Aerial}}} & \multicolumn{2}{c}{\textit{\textcolor{teal}{\footnotesize Outdoor}}} & \multicolumn{2}{c}{\textit{\textcolor{teal}{\footnotesize Indoor}}} & \multicolumn{2}{c}{\textit{\textcolor{teal}{\footnotesize Aerial}}}\\ 
\bottomrule
\end{tabular}
\end{table*}

\begin{table*}[tb]
\centering
\small
\caption{\textbf{Effect of Adding Datasets.} 
We progressively evaluate baseline models trained on single datasets (\textit{SNAP-SN} or \textit{SNAP-KITTI}) and baseline models trained on specific domains \textit{SNAP-Indoor}, \textit{SNAP-Outdoor}, and \textit{SNAP-Aerial} against \textit{SNAP-C}, the complete model trained on all data. The \textcolor{gray}{gray}-shaded cells denote evaluations performed on out-of-distribution datasets.
}
\vspace{-0.8em}
\label{table:effect_adding_datasets}
\setlength{\tabcolsep}{4pt}
\begin{tabular}{l|c c|c c|c c|c c|c c|c c|c c|c c}
\toprule
\multirow{5}{*}{Model} & \multicolumn{16}{c}{${\text{IoU}}@k \uparrow$ } \\
\cmidrule{2-17}
\addlinespace[-2.5pt]
& \multicolumn{6}{c}{\cellcolor{mediumblue!20}In-Distribution} & \multicolumn{10}{c}{\cellcolor{green!20}Zero-Shot} \\
\addlinespace[-2pt]
\cmidrule{2-17} 
& \multicolumn{2}{c}{SemanticKITTI} & \multicolumn{2}{c}{ScanNet20} & \multicolumn{2}{c}{STPLS3D} & \multicolumn{2}{c}{Matterport3D} & \multicolumn{2}{c}{S3DIS Full} & \multicolumn{2}{c}{KITTI-360 Full} & \multicolumn{2}{c}{Waymo} & \multicolumn{2}{c}{UrbanBIS} \\
\cmidrule(lr){2-3} \cmidrule(lr){4-5} \cmidrule(lr){6-7} \cmidrule(lr){8-9} \cmidrule(lr){10-11} \cmidrule(lr){12-13} \cmidrule(lr){14-15} \cmidrule(lr){16-17}
& @1 & @5 & @1 & @5 & @1 & @5 & @1 & @5 & @1 & @5 & @1 & @5 & @1 & @5 & @1 & @5 \\
\midrule

SNAP - SN &   \cellcolor{gray!25}3.1 &   \cellcolor{gray!25}5.9 & \textbf{68.6} & 82.1 &   \cellcolor{gray!25}2.5 &   \cellcolor{gray!25}4.1 & \textbf{53.4} & 71.3 & 51.4 & 70.0 &   \cellcolor{gray!25}0.2 &   \cellcolor{gray!25}0.5 &   \cellcolor{gray!25}0.7 &   \cellcolor{gray!25}3.1 &   \cellcolor{gray!25}6.3 &   \cellcolor{gray!25}16.8 \\

SNAP - KITTI & 68.1 & 84.5 &   \cellcolor{gray!25}13.8 &   \cellcolor{gray!25}33.6 &   \cellcolor{gray!25}24.6 &   \cellcolor{gray!25}52.9 &   \cellcolor{gray!25}16 &   \cellcolor{gray!25}33.3 &   \cellcolor{gray!25}11.6 &   \cellcolor{gray!25}29.1 &  \cellcolor{gray!25}6.7 &  \cellcolor{gray!25}29.7 & 48.2 & 66.3 &   \cellcolor{gray!25}43.1 &   \cellcolor{gray!25}71.3 \\

SNAP - Indoor &   \cellcolor{gray!25}3.1 &   \cellcolor{gray!25}15.3 & 66.0 & 81.3 &   \cellcolor{gray!25}5.9 &   \cellcolor{gray!25}18.6 & 49.9 & 74.2 & 51.9 & 76.9 &  \cellcolor{gray!25}0.2 &   \cellcolor{gray!25}2.1 &   \cellcolor{gray!25}1.1 &   \cellcolor{gray!25}6.5 &   \cellcolor{gray!25}3.4 &   \cellcolor{gray!25}33.6\\

SNAP - Outdoor & 71.3 & 85.7 &   \cellcolor{gray!25}14.9 &   \cellcolor{gray!25}43.5 &   \cellcolor{gray!25}36.4 &   \cellcolor{gray!25}57.9 &   \cellcolor{gray!25}15.1 &   \cellcolor{gray!25}39.2 &   \cellcolor{gray!25}14.9 &   \cellcolor{gray!25}43.6 & 18.3 & 44.6 & 68.5 & 86.0 &   \cellcolor{gray!25}45.9 &   \cellcolor{gray!25}78 \\

SNAP - Aerial &   \cellcolor{gray!25}27.1 &   \cellcolor{gray!25}57.9 &   \cellcolor{gray!25}6.3 &   \cellcolor{gray!25}19.1 & 65.8 & 79.1 &   \cellcolor{gray!25}9.4 &   \cellcolor{gray!25}21.2 &   \cellcolor{gray!25}7.3 &   \cellcolor{gray!25}20.6 &   \cellcolor{gray!25}5.6 &   \cellcolor{gray!25}19.4 &   \cellcolor{gray!25}25.1 &   \cellcolor{gray!25}60.4 & \textbf{74.2} & 86.9 \\

SNAP - C & \textbf{71.5} & \textbf{86.0} & 67.7 & \textbf{82.3} & \textbf{67.8} & \textbf{80.4} & 52.6 & \textbf{75.2} & \textbf{53.6} & \textbf{77.6} & \textbf{23.1} & \textbf{48.1} & \textbf{69.8} & \textbf{86.6} & 71.6 & \textbf{90.2} \\
\bottomrule
\end{tabular}
\vspace{-10pt}
\end{table*}

\begin{table}[tb]
\centering
\small
\caption{\textbf{Comparison of Automatic Prompt Points Generation Strategies.} We evaluate our iterative prompting approach against four baseline methods on ScanNet200. 
Best results are highlighted in \textbf{bold}, and second best results are \uline{underlined}.
}
\vspace{-0.8em}
\label{table:auto_spatial_prompting_ablation}
\setlength{\tabcolsep}{2.8pt}
\begin{tabular}{l|c c c c c c|c}
\toprule
Approach & AP & $\text{AP}_{50}$ & $\text{AP}_{25}$ & \makecell{Head\\mAP} & \makecell{Common\\mAP} & \makecell{Tail\\mAP} & \makecell{Time\\(s)} \\
\midrule
Uniform grid & \textbf{39.1} & \textbf{58.5} & 68.8 & \textbf{39.8} & \textbf{40.6} & 36.3 & 1421 \\
FPS & 36.2 & 54.5 & 65.0 & \uline{38.6} & 38.4 & 30.4 & 702 \\
HDBSCAN & 4.7 & 6.5 & 7.2 & 6.5 & 5.4 & 1.7 & 714 \\
Ours & \uline{38.7} & \uline{58.4} & \textbf{69.2} & 38.2 & \uline{39.7} & \textbf{38.1} & \textbf{461} \\
\bottomrule
\end{tabular}
\vspace{-1em}
\end{table}

\noindent\textbf{Effect of Input Properties.}
We conducted an ablation study to evaluate the contribution of different input properties to our model's performance, with results summarized in \textbf{Tab.~\ref{table:ablation_input_modalities}}. The experiments reveal that \shortname~relies predominantly on geometric features. On the ScanNet20 dataset, the inclusion of surface normals (N) consistently improves performance across all metrics, boosting IoU@1 from 67.4 to 68.3. In contrast, color information (C) appears to be largely redundant; its inclusion provides no discernible benefit when normals are present and only a marginal gain otherwise. This trend is consistent across other datasets, where Lidar intensity (S) on SemanticKITTI and color on STPLS3D offer only minor improvements to IoU@1. 
While surface normals improve performance, computing accurate normals is challenging for real-world point clouds. Therefore, SNAP operates solely on XYZ coordinates to ensure applicability across diverse data sources.

\vspace{0.2em}
\noindent\textbf{Effect of Loss functions on Mask Accuracy.}
\shortname~ uses a combination of loss functions to guide the network. To check the effectiveness of each module, we run ablations by adding each component sequentially. The results are summarized in \textbf{Tab.~\ref{table:ablation_loss_functions}}. Results indicate that auxiliary loss and weighted loss improve performance at a higher number of clicks. Including text classification loss leads to the highest improvements and indicates that semantic understanding is important for segmentation tasks.

\vspace{0.2em}
\noindent\textbf{Effect of Normalization Strategies.}
We evaluate 3 normalization strategies to validate our domain-based design choice: (1) a single model with standard batch normalization across all data, (2) dataset-specific normalization with individual layers for each training dataset, and (3) our proposed domain normalization with separate layers for each domain.
As shown in \textbf{Tab.~\ref{table:ablation_domain_norm}}, batch normalization shows reasonable in-distribution performance (64.3 IoU@1 on SemanticKITTI) but tends to underperform on zero-shot datasets.
Dataset normalization can achieve slightly higher scores on in-distribution sets, 
but this comes at the cost of limited generalization. Domain-specific normalization performs the best in zero-shot generalization, suggesting potential benefits to domain-level grouping over dataset-specific parameters. More importantly, this simple domain-norm translates into a simple domain-type checkbox selection, making it highly user-friendly.

\vspace{0.2em}
\noindent\textbf{Effect of Progressively Adding Datasets.}
To verify the effectiveness of scaling up training data, 
we evaluate various \shortname\ variants on in-distribution and unseen datasets. As shown in \textbf{Tab.~\ref{table:effect_adding_datasets}}, scaling up the training set leads to consistent improvements in performance over both in-distribution as well as unseen datasets. \shortname-C consistently matches or surpasses the performances of single dataset or single domain models, suggesting that the proposed domain normalization is largely able to reduce the effects of any negative transfer from cross-domain datasets.

\vspace{0.2em}
\noindent\textbf{Design Choices for Automatic Prompt Points Generation.}
To generate prompt points automatically, we initially considered uniform grid sampling following SAM~\cite{SAM}. However, this approach faces a fundamental trade-off: large voxel sizes miss small objects while small voxels yield a computationally prohibitive numbers of points. Alternative approaches like Farthest Point Sampling and HDBSCAN~\cite{mcinnes2017hdbscan} require scene-specific tuning and generally undersample dense regions while oversampling sparse regions. 
Our iterative approach addresses these limitations 
and maintains uniform sampling independent of local point density. As shown in \textbf{Tab.~\ref{table:auto_spatial_prompting_ablation}}, this strategy achieves comparable coverage to naive grid sampling (39.1 AP vs 38.7 AP) with significantly fewer points, reducing computation time by ~\textbf{68\%}. Notably, our approach excels on tail classes (38.1 mAP), outperforming all baselines.  


\section{Conclusion}

\label{sec:conclusion}
In this paper, we introduced \shortname, a unified model for flexible promptable point cloud segmentation that is compatible with both spatial and textual prompts. By training on heterogeneous datasets with cross-domain normalization, \shortname\ demonstrates state-of-the-art performance across a wide range of datasets. 
We believe that this unified model will come across as a handy tool for users, moving away from the need for dataset or domain specific models. 
For future work, following the promising trend found in this paper, investigating the effect of scaling up further via self-supervised or weakly-supervised learning on unlabeled data is an appealing direction.

\section{Acknowledgements}

This work was partially supported by the National Science Foundation under Award IIS-2310254.
\clearpage
{
    \small
    \bibliographystyle{IEEEtranN}
    \bibliography{main}
}
\clearpage
\setcounter{page}{1}
\maketitlesupplementary

\textit{This supplementary document is structured as follows:}
\begin{itemize}
    \item \darkblueref{sec:model_details}{Model Details}
    \begin{itemize}
        \item \blueref{sec:detailed_arch}{Detailed Model Architecture}
        \item \blueref{sec:loss_details}{Details on Loss functions}
        \item \blueref{sec:auto_prompt_generation}{Auto Prompt Generation}
    \end{itemize}
    \item \darkblueref{sec:implementation_details}{Implementation Details}
    \begin{itemize} 
        \item \blueref{sec:eval_metrics}{Details on Evaluation Metrics}
        \item \blueref{sec:dataset_details}{Dataset Details}
        \item \blueref{sec:hdbscan_details}{HDBSCAN Details}
        \item \blueref{sec:training_details}{Training Details}
    \end{itemize}
    
    \item \darkblueref{sec:additional_ablations}{Additional Ablations}
    \begin{itemize}
        \item \blueref{sec:backbone_ablation}{Backbone Ablation}
        \item \blueref{sec:click_strategy_ablation}{Click-Strategy Ablation}
        \item \blueref{sec:performance_with_cross_domain_inputs}{Cross-Domain Input Ablation}
    \end{itemize}
    \item \darkblueref{sec:quantitative_results}{Additional Quantitative Results}
    \begin{itemize}
        \item \blueref{sec:timing_memory}{Timing and Memory Comparison}
        \item \blueref{sec:comparison_against_fully_supervised}{Class Agnostic Interactive Segmentation against Non-Interactive Fully Supervised Methods}
        \item \blueref{sec:full_interactive_results}{Interactive segmentation results with all model variants}
    \end{itemize}
    
    \item \darkblueref{sec:qualitative_results}{Additional Qualitative results}
\end{itemize}

\section{Model Details}
\label{sec:model_details}

\subsection{Model Architecture}
\label{sec:detailed_arch}
\begin{figure*}[tb]
    \centering
    \setlength{\tabcolsep}{2pt}
    \resizebox{\textwidth}{!}{
    \begin{tabular}{cc}
        \includegraphics[width=0.5\textwidth]{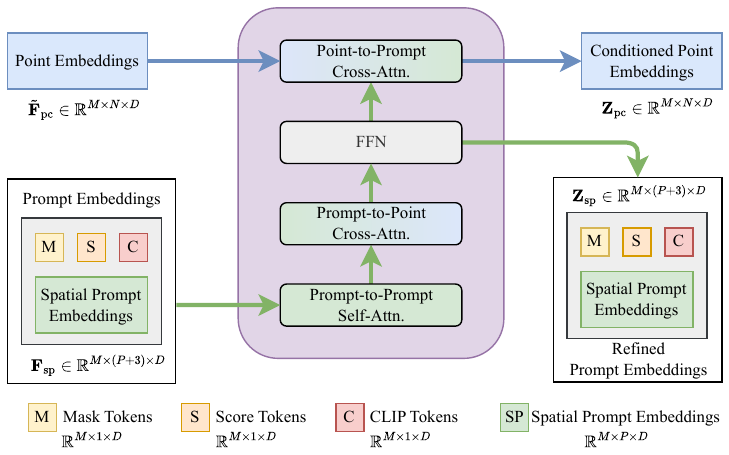} & 
        \includegraphics[width=0.5\textwidth]{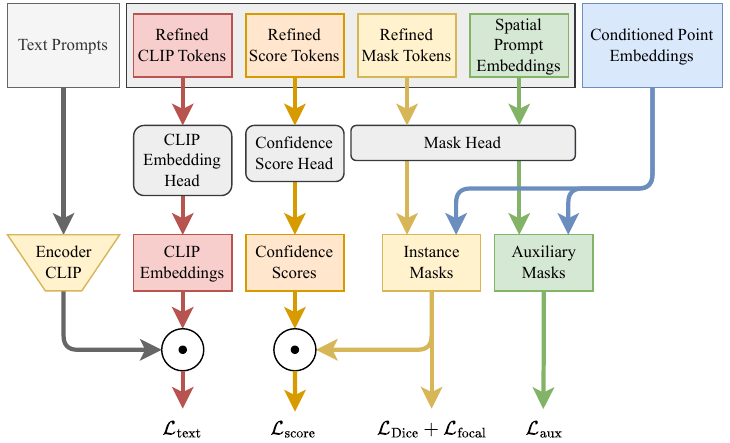} \\
        \footnotesize{(a) Prompt-Point Attention Module.} & \footnotesize{(b) Text Encoder \& Prediction Heads.}
    \end{tabular}
    }
    
    \caption{
    \textbf{Detailed architecture of \shortname}. (a) The Prompt-Point Attention module refines both point and prompt embeddings within the Mask Decoder. (b) The refined embeddings are then fed into several lightweight prediction heads for mask, confidence score, and CLIP embedding predictions. For completeness, we also indicate the external CLIP Text Encoder for processing text prompts and the supervision signals associated with each branch.
    }
    \label{fig:detailed_model_arch}
    \vspace{-1em}
    
\end{figure*}

\noindent To complement the simplified pipeline presented in 
the main paper, we provide a more detailed illustration of \shortname~in \textbf{Fig.~\ref{fig:detailed_model_arch}}. Specifically, the Mask Decoder is decomposed into a Prompt-Point Attention module and three dedicated MLP heads that process the mask, score, and CLIP tokens, together with the spatial prompt embeddings. For completeness, we also indicate the corresponding supervision signals, showing how each type of prediction contributes to the overall training objective through its associated loss functions.

\subsection{Details on Loss Functions}
\label{sec:loss_details}
\noindent\textbf{Focal and Dice Loss.} 
Following Interactive4D~\cite{interactive4D}, we apply a distance-based, click-localized weight to the point-wise Focal loss and Dice loss terms ($\mathcal{L}_\text{focal}$ and $\mathcal{L}_\text{dice}$). Concretely, for each point $\mathbf{p}_i\in \mathbf{P}$, we compute its normalized distance to its nearest spatial prompt point $\mathbf{p}_\text{sp}^*$: $d_i=\text{Dist}(\mathbf{p}_i, \mathbf{p}_\text{sp}^*)$. If $d_i$ is below a threshold $\tau_d$, the weight is defined to decay linearly from $w_\text{max}$ to $w_\text{min}$ as the distance increases; otherwise, the weight is set to $w_\text{min}$. Formally, the weight of each point is defined as:
\begin{equation}
w(\mathbf{p}_i) =
\begin{cases}
w_{\max} - (w_{\max}-w_{\min}) \, d_i, & d_i < \tau_d, \\[4pt]
w_{\min}, & \text{otherwise}.
\end{cases}
\end{equation}
In our implementation, we set $w_\text{max}=2$, $w_\text{min}=1$, and $\tau=0.5$. This weighting strategy increases the contribution of points closer to spatial prompts (clicks), encouraging the model to focus its supervision around click regions while preserving global mask consistency.

\noindent\textbf{Auxiliary Loss.} In addition to the final mask prediction loss described above, we strengthen supervision by leveraging the spatial prompt embeddings corresponding to individual clicks. The intuition is that each click should independently guide a plausible mask prediction, rather than only contributing through the aggregated mask token. To achieve this, we treat the $P$ prompt embeddings extracted from $\mathbf{Z}\text{sp}\in\mathbb{R}^{M\times(P+3)\times D}$ as \textit{auxiliary mask tokens} and feed them through the same mask head described in 
the method section.
This yields $M\times P$ auxiliary mask predictions, which are supervised using standard point-wise focal and Dice loss terms. The resulting auxiliary loss $\mathcal{L}_\text{aux}$ encourages individual clicks to directly align with the segmentation masks, thereby providing more fine-grained supervision.

\noindent\textbf{Confidence Score Loss.} To improve the reliability of mask confidence score estimation, we supervise this score prediction process with $\mathcal{L}_\text{score}$. Intuitively, this score should reflect the quality of the predicted mask, which we approximate by its intersection-over-union (IoU) with its corresponding ground-truth mask. Concretely, given the predicted mask $\mathcal{M}_i$, we first obtain a binarized mask by thresholding it: $\mathcal{M}_i>\tau$. The IoU between this mask and its ground-truth counterpart $\mathcal{M}_i^*$ is then computed and used as the regression target: $\mathcal{S}_i^*=\text{IoU}(\mathcal{M}_i>\tau, \mathcal{M}_i^*)$. Finally, we formulate the score loss as a mean squared error (MSE) between the predicted score and the IoU target:
\begin{equation}
    \mathcal{L}_\text{score}=\frac{1}{M}\sum_{i=1}^M\big(\mathcal{S}_i-\mathcal{S}_i^*\big)^2,
\end{equation}
where $\mathcal{S}_i$ denotes the predicted score for the $i$-th mask. 

\noindent\textbf{Text Loss.}
To supervise the predicted CLIP tokens $\mathbf{L}_\text{CLIP}$, we follow a prototype-based classification scheme against the CLIP text vocabulary embeddings $\mathbf{T}\in\mathbb{R}^{C\times D_\text{CLIP}}$. After L2 normalization, cosine similarities between $\mathbf{L}_\text{CLIP}$ and $\mathbf{T}$ yield logits $\mathbf{Z}=\mathbf{L}_\text{CLIP}\cdot\mathbf{T}^\top\in\mathbb{R}^{M\times C}$. For the $i$-th sample with ground-truth label $y$, let $p_i=\text{softmax}(\mathbf{Z}_i)_y$ denote the probability of the correct class. We then apply a focal loss with focusing parameter $\gamma=2.0$ and no additional class re-weighting, which gives the text loss:
\begin{equation}
\mathcal{L}_\text{text}
= \frac{1}{M}\sum_{i=1}^{M} \big(1-p_i\big)^{\gamma}\big(-\log p_i\big).
\end{equation}

\subsection{Auto Prompt Generation}
\label{sec:auto_prompt_generation}
\begin{figure*}[ht]
    \centering
    \setlength{\tabcolsep}{1pt}
    \setlength{\fboxsep}{0pt}  %
    \setlength{\fboxrule}{1pt} %
    \begin{small}
    \begin{tabular}{cc}
        \fbox{\includegraphics[width=0.35\linewidth]{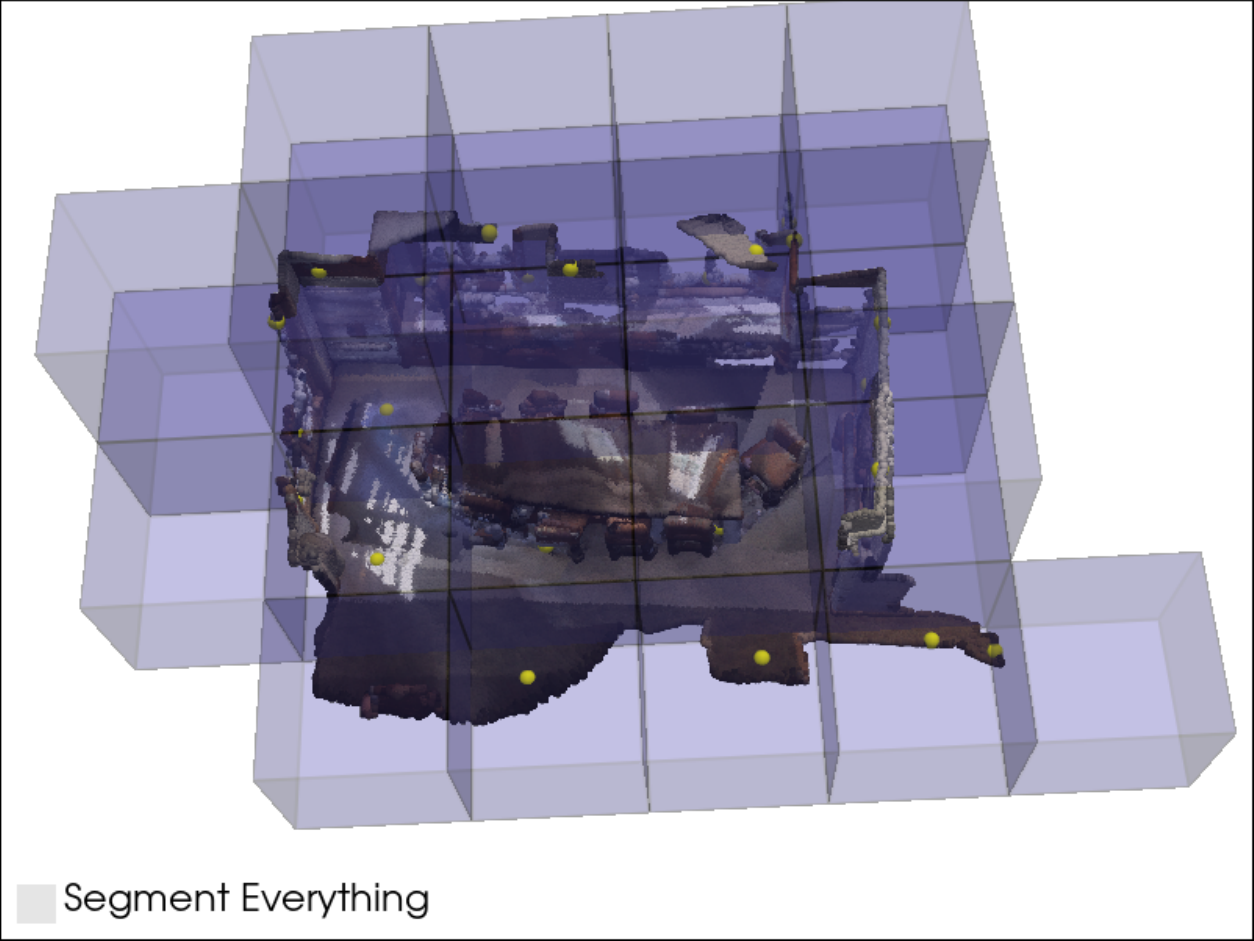}} & 
        \fbox{\includegraphics[width=0.35\linewidth]{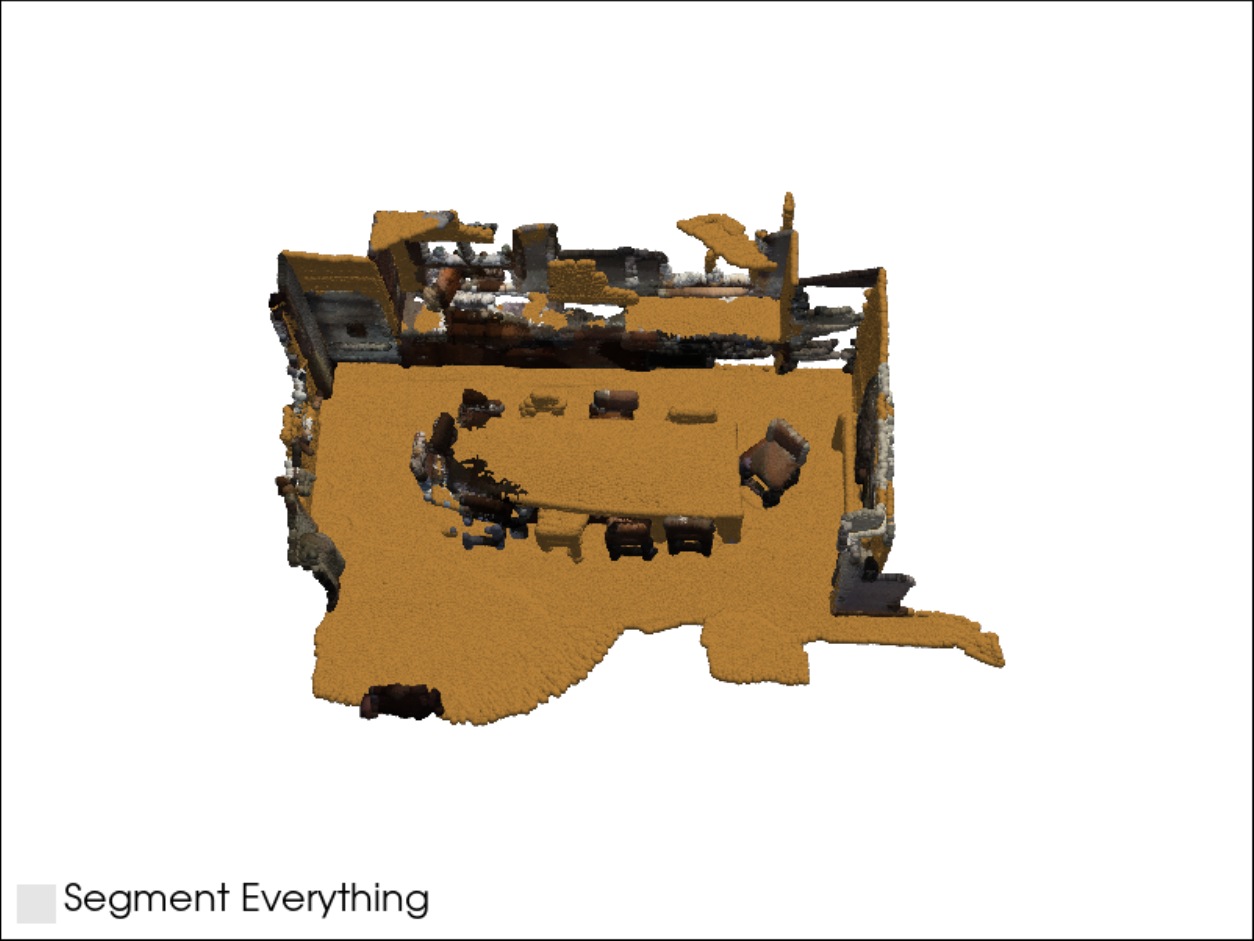}} \\
        (a.1) & (a.2) \\
        
        \fbox{\includegraphics[width=0.35\linewidth]{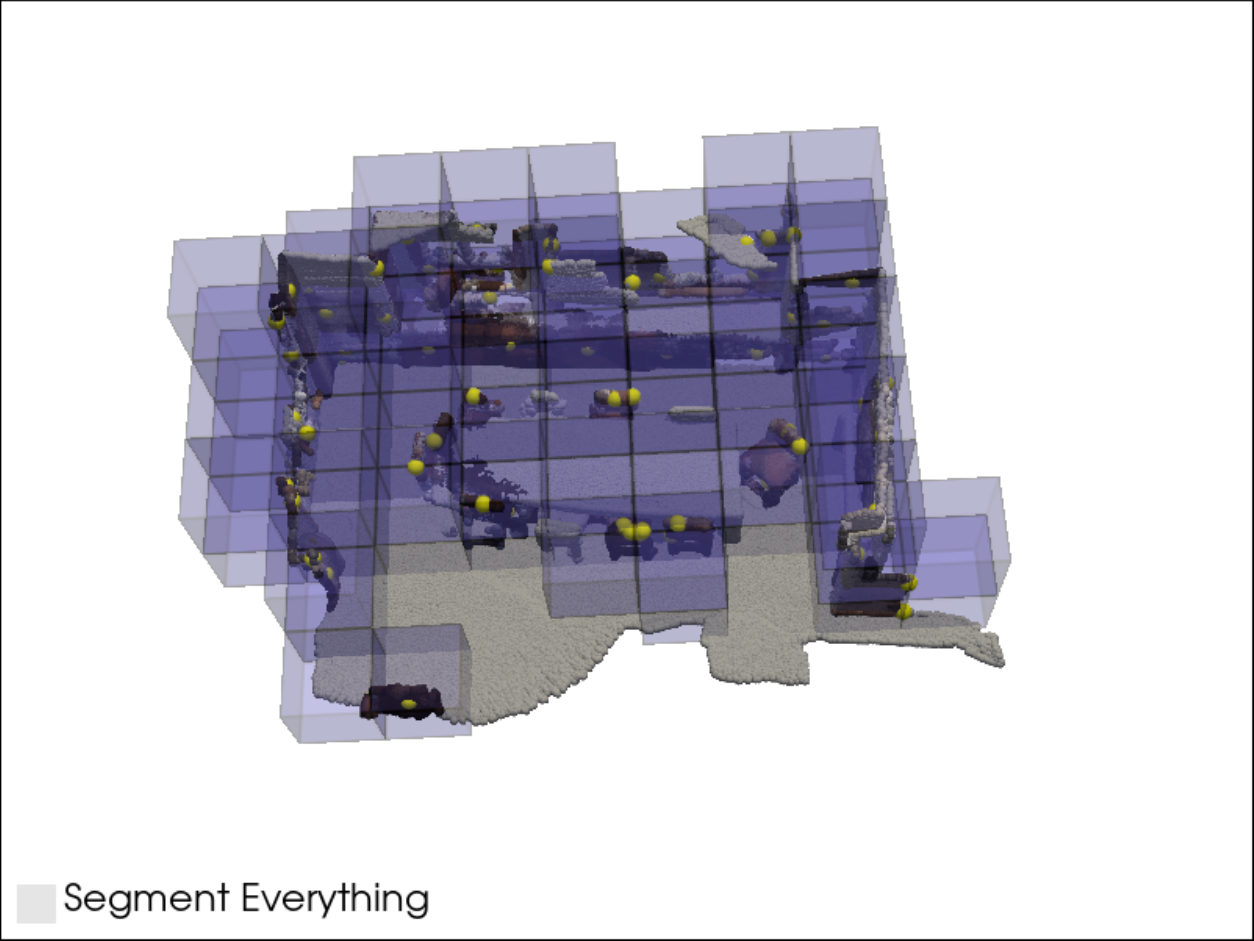}} & 
        \fbox{\includegraphics[width=0.35\linewidth]{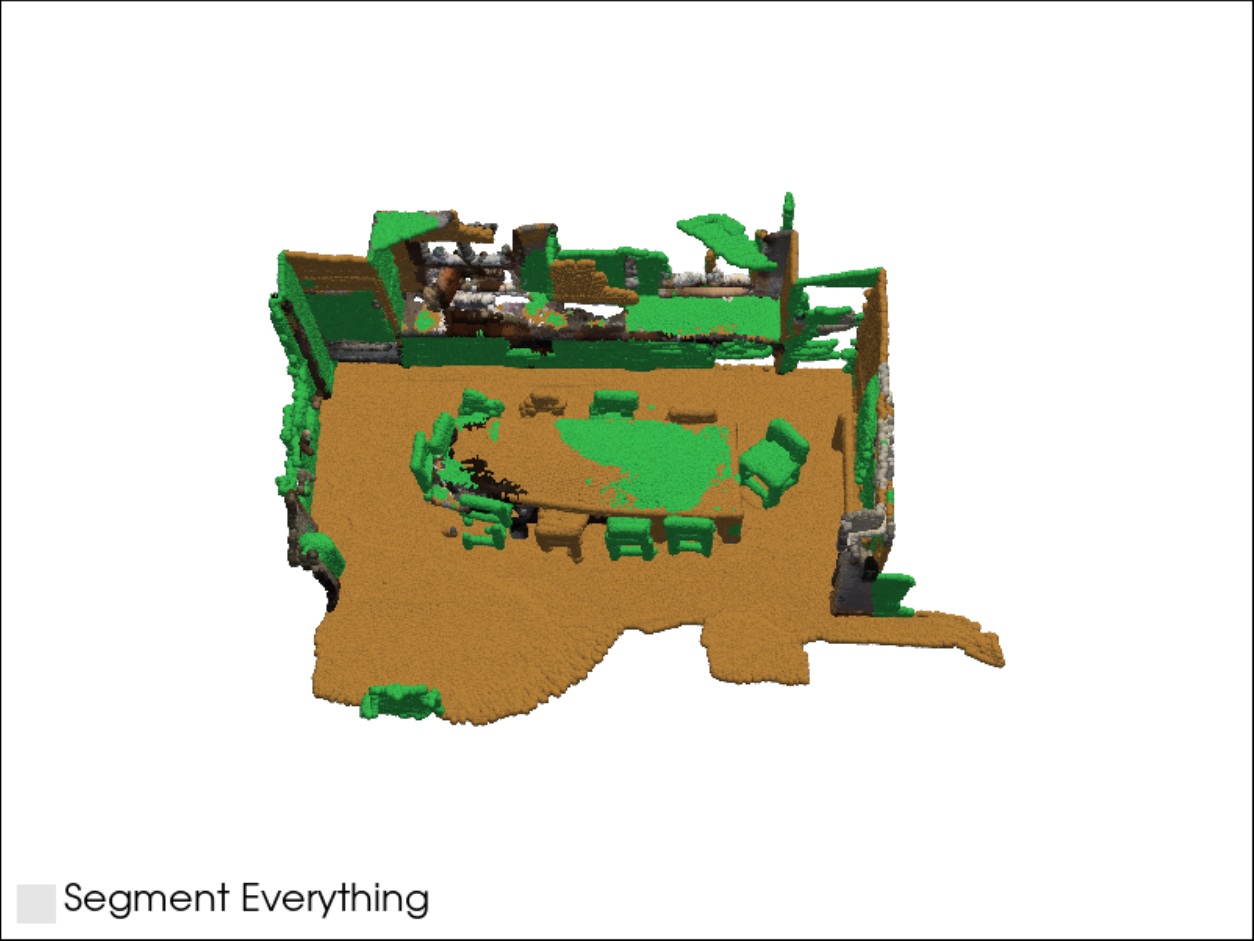}} \\
        (b.1) & (b.2) \\
        
        \fbox{\includegraphics[width=0.35\linewidth]{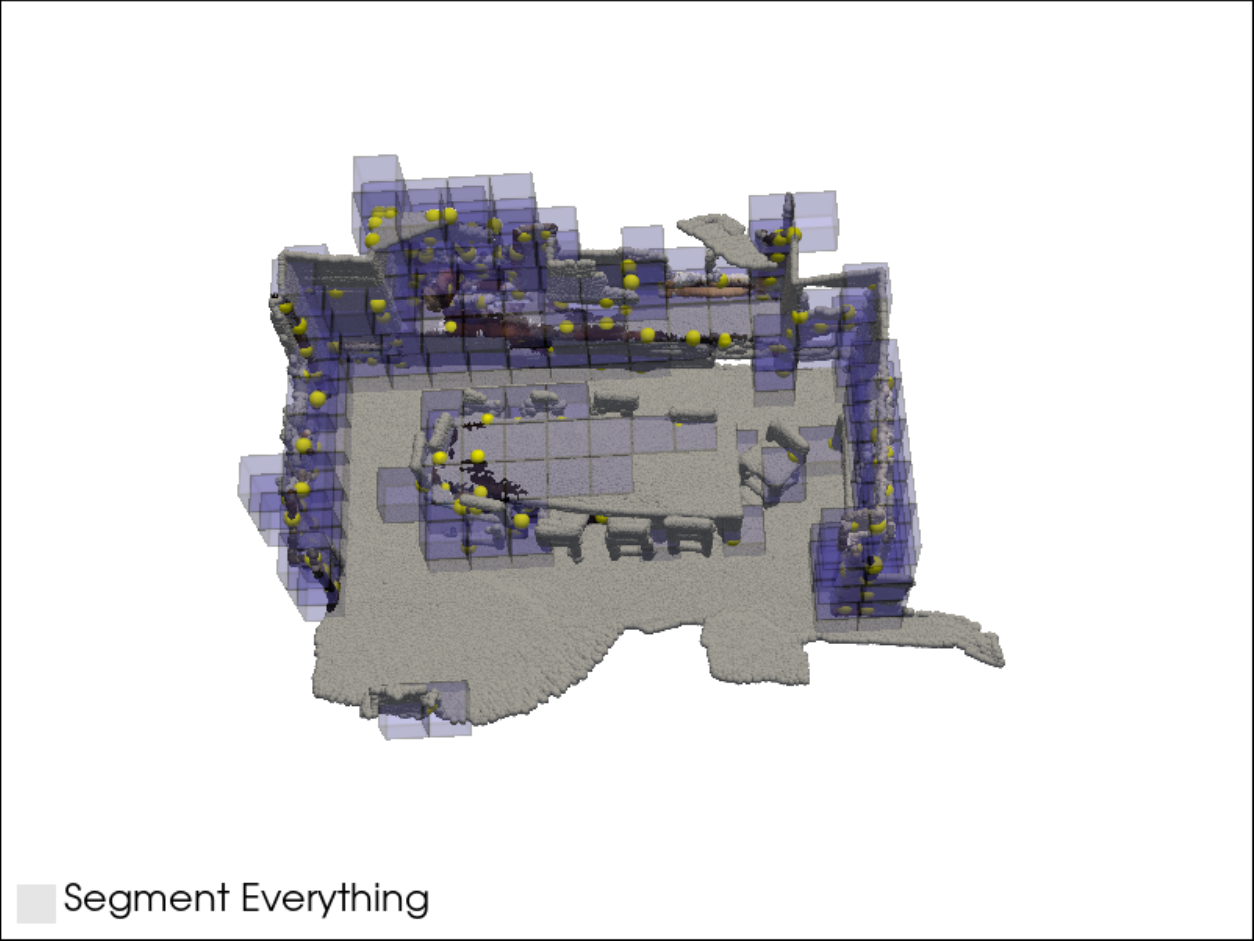}} & 
        \fbox{\includegraphics[width=0.35\linewidth]{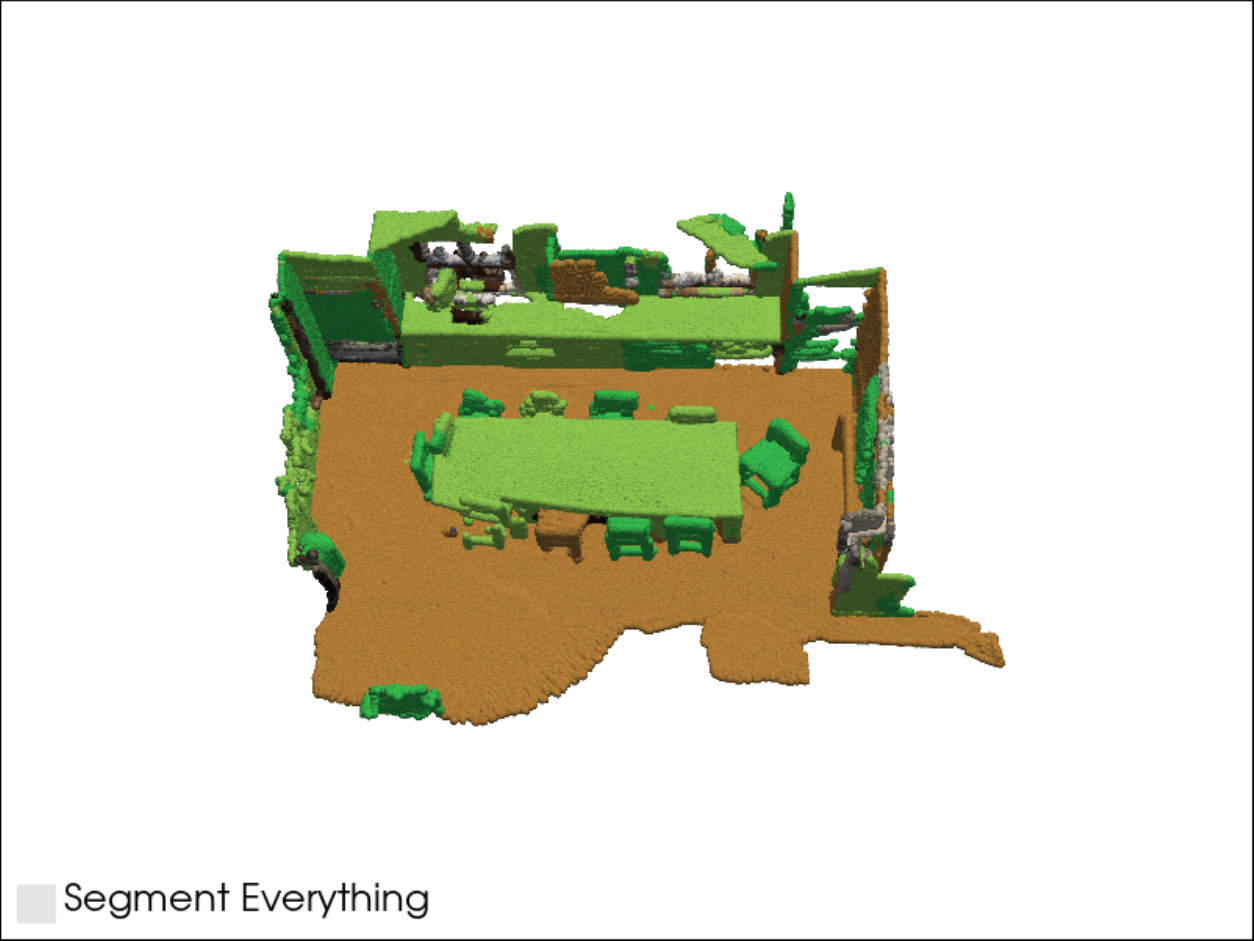}} \\
        (c.1) & (c.2) \\
        
        \multicolumn{2}{c}{\fbox{\includegraphics[width=0.35\linewidth]{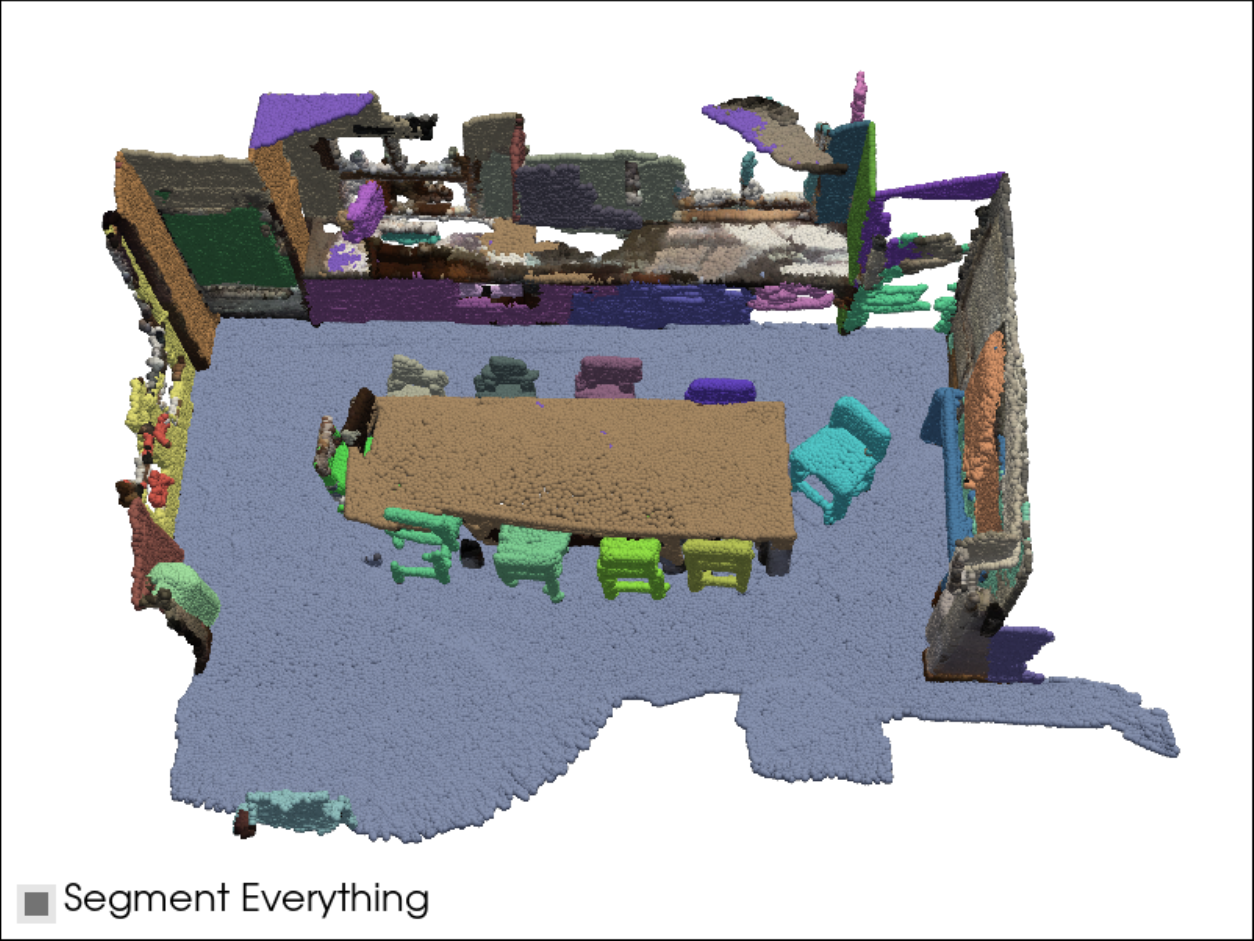}}} \\
        \multicolumn{2}{c}{(d)}
        
    \end{tabular}
    \end{small}
    \vspace{-0.8em}
    \caption{\textbf{Visualization of the Iterative Prompting algorithm.} \textbf{(a.1):} Generate prompt points with a large voxel size to segment out large objects in the scene, \textbf{(a.2):} All the points segmented after first iteration (Yellow color). \textbf{(b.1):} Reduce the voxel size and generate prompt points on the unsegmented points. \textbf{(b.2):} All points segmented after Iteration 1 (yellow) and 2(dark green). \textbf{(c.1):} Reduce voxel size again and repeat. \textbf{(c.2):} All points segmented after Iteration 1 (yellow), 2(dark green) and 3 (light green). \textbf{(d):} Final instance masks after Non Maximum Suppression.}
    \label{fig:iterative_prompting}
\end{figure*}

Given an input point cloud $\mathcal{P}$ with $N$ points, let $\mathcal{F}$ denote the segmentation model and $d$ the scene domain (outdoor / indoor / aerial). We define $v_0$ as the initial domain-specific voxel size, $K_\text{max}$ as the maximum number of iterations, $\tau_\text{s}$ as the predicted confidence score threshold, and $\tau_\text{nms}$ as the NMS IoU threshold. 
The algorithm iteratively generates prompt points, segments objects of decreasing scales, and refines the results across iterations, as illustrated in \textbf{Fig.~\ref{fig:iterative_prompting}}. Finally, it outputs a set of masks $\mathcal{M}$, their corresponding text embeddings $\mathcal{T}$, and confidence scores $\mathcal{S}$. 
The complete procedure is summarized in \textbf{Alg.~\ref{alg:auto_prompt}}. 

\begin{algorithm}
\caption{Mask generation with iterative prompting algorithm. 
}
\label{alg:auto_prompt}
\begin{algorithmic}[1]
\Require Point cloud $\mathcal{P} = \{p_i\}_{i=1}^N$, segmentation model $\mathcal{F}$, scene domain $d$, initial domain-specific voxel size $v_0$, maximum number of iterations $K_\text{max}$, confidence score threshold $\tau_\text{s}$, NMS IoU threshold $\tau_\text{nms}$
\Ensure Set of masks $\mathcal{M}$, text embeddings $\mathcal{T}$, confidence scores $\mathcal{S}$
\State $v_0 \gets$ domain-specific voxel size
\State $\mathcal{C} \gets \{0\}^N$ \Comment{Coverage mask}
\State $\mathcal{M}, \mathcal{T}, \mathcal{S} \gets \emptyset$
\For{$k = 0$ to $K_\text{max} - 1$}
    \State $\mathcal{U} \gets \{p_i : \mathcal{C}_i = 0\}$ \Comment{Get uncovered points}
    \State $v_k \gets v_0 / 2^k$ \Comment{Halve voxel size}
    \State $\mathcal{Q} \gets \text{VoxelDownsample}(\mathcal{U}, v_k)$ \Comment{Generate prompts}
    \State $\mathcal{M}^k, \mathcal{T}^k, \mathcal{S}^k \gets \mathcal{F}(\mathcal{P}, \mathcal{Q})$ \Comment{Run model}
    \For{$j = 1$ to $|\mathcal{M}^k|$}
        \If{$\mathcal{S}_j^k \geq \tau_{s}$}
            \State $\mathcal{M} \gets \mathcal{M} \cup \{\mathcal{M}_j^k\}$
            \State $\mathcal{T} \gets \mathcal{T} \cup \{\mathcal{T}_j^k\}$
            \State $\mathcal{S} \gets \mathcal{S} \cup \{\mathcal{S}_j^k\}$
            \State $\mathcal{C} \gets \mathcal{C} \lor \mathcal{M}_j^k$ \Comment{Update coverage}
        \EndIf
    \EndFor
\EndFor
\State $\mathcal{M}, \mathcal{T}, \mathcal{S} \gets \text{NMS}(\mathcal{M}, \mathcal{T}, \mathcal{S}, \tau_\text{nms})$
\State \textbf{return} $\mathcal{M}, \mathcal{T}, \mathcal{S}$
\end{algorithmic}
\end{algorithm}
\vspace{-5pt}

\section{Implementation Details}
\label{sec:implementation_details}

\subsection{Details on Evaluation Metrics}
\label{sec:eval_metrics}
\textbf{IoU@k.} Following conventions from \cite{interactive4D, AGILE3D, InterObject3D}, we evaluate using IoU@$k$, the average intersection over union (IoU) achieved with $k$ clicks per object, \uline{averaged across all objects}.  

\noindent\textbf{Average Precision.} In our comparisons against the baselines for open-vocabulary segmentation, we use the Average Precision metric defined as follows - 
\begin{equation}
\text{mAP} = \frac{1}{C \times 10} \sum_{c=1}^{C} \sum_{\tau=0.5}^{0.95} \text{AP}_{\tau}^{c} \quad \text{(step size = 0.05)}
\end{equation}
where $C$ is the number of classes. 

\noindent\textbf{Panoptic Segmentation metrics.} To assess panoptic quality, we utilize the Panoptic Segmentation metrics as defined in~\cite{kirillov2019panopticsegmentation} and as used by SAL~\cite{SAL}. Specifically, PQ is Panoptic Quality, SQ is Segmentation Quality and RQ is Recognition Quality. TP, FP, FN represent True Positives, False Positives and False Negatives respectively. For class-aware segmentation, A prediction is counted as a TP if it has a $\text{IoU} > 0.5$ and the correct label. For class-agnostic segmentation, we assume class predictions are correct and TP is counted if $\text{IoU} > 0.5$.

\begin{equation}
\text{PQ} = \underbrace{\frac{|TP|}{|TP| + \frac{1}{2}|FP| + \frac{1}{2}|FN|}}_{\text{Recognition Quality (RQ)}} \times \underbrace{\frac{\sum_{(p,g) \in TP} \text{IoU}(p,g)}{|TP|}}_{\text{Segmentation Quality (SQ)}}
\end{equation}

\subsection{Dataset Details}
\label{sec:dataset_details}

We provide a summary of the dataset statistics in \textbf{Tab.~\ref{table:supp_dataset_statistics}}. Samples from each dataset illustrating the various domains are visualized in \textbf{Fig.~\ref{fig:train_data_examples}},\textbf{ \ref{fig:val_data_examples_1} }and\textbf{ \ref{fig:val_data_examples_2}}.
\begin{table}[tb]
\centering
\caption{\textbf{Summary of Dataset Statistics.}}
\vspace{-0.8em}
\label{table:supp_dataset_statistics}
\small
\setlength{\tabcolsep}{3pt}
\begin{tabular}{@{}lccll@{}}
\toprule
\multicolumn{5}{c}{Training Datasets} \\
\midrule
Dataset & Train & Val & Domain & Sensor Type \\
\midrule
SemanticKITTI & 19,130 & 4,071 & Outdoor & HDL-64 LiDAR \\
nuScenes & 28,130 & 6,019 & Outdoor & 32-beam LiDAR \\
PandaSet & 2,000 & 400 & Outdoor & Pandar64 \\
ScanNet & 1,201 & 312 & Indoor & RGBD Camera \\
HM3D & 1805 & 481 & Indoor & RGBD camera \\
STPLS3D & 3,395 & 500 & Aerial & Photogrammetry \\
DALES & 2,900 & 1,100 & Aerial & Aerial LiDAR \\
\softmidrule
Total & 58,561 & 12,883 & & \\
\midrule
\multicolumn{5}{c}{Zero-Shot Validation Datasets} \\
\midrule
Dataset & \multicolumn{2}{c}{Val} & Domain & Sensor Type \\
\midrule
Waymo & \multicolumn{2}{c}{5,976} & Outdoor & Proprietary LiDAR \\
KITTI-360 SS & \multicolumn{2}{c}{13,440} & Outdoor & HDL-64 LiDAR \\
KITTI-360 Full & \multicolumn{2}{c}{61} & Outdoor & HDL-64 LiDAR \\
KITTI-360 Crops & \multicolumn{2}{c}{3,421} & Outdoor & HDL-64 LiDAR \\
Matterport3D & \multicolumn{2}{c}{233} & Indoor & RGBD Camera \\
ScanNet++ & \multicolumn{2}{c}{178} & Indoor & RGBD Camera \\
S3DIS Crops & \multicolumn{2}{c}{2,330} & Indoor & RGBD Camera \\
S3DIS Full & \multicolumn{2}{c}{68} & Indoor & RGBD Camera \\
UrbanBIS & \multicolumn{2}{c}{46} & Aerial & Photogrammetry \\
\softmidrule
Total & \multicolumn{2}{c}{25,753} & & \\
\bottomrule
\end{tabular}
\vspace{-1em}
\end{table}

\subsubsection{Indoor Datasets}
\textbf{ScanNetV2} \cite{scannet} is a richly annotated dataset of 3D indoor scenes, covering a wide variety of scenes including offices, rooms, hotels etc. It provides semantic segmentation masks for 200 fine-grained classes, known as ScanNet200, and 20 coarser classes known as ScanNet20. We evaluate on both the benchmarks. 

\noindent\textbf{Habitat Matterport 3D}~\cite{ramakrishnan2021hm3d} is a large scale annotated dataset for 3D indoor scenes which covers 216 3D spaces and 3100 rooms within these spaces. It provides instance annotation for 40 categories. After processing, it provides us with 1805 samples for training and 481 samples for validation. 

\noindent\textbf{ScanNet++} \cite{scannetpp} comes with high-fidelity 3D mask annotations including smaller objects which are not well labeled in the ScanNet datasets. It includes high-resolution 3D scans captured at sub-millimeter precision and annotated comprehensively, covering objects of varying sizes. 

\noindent\textbf{Matterport3D}~\cite{Matterport3D} is a collection of 90 high-quality 3D reconstructions of indoor environments with instance annotations for 21 object categories. After processing, it provides us with 233 samples for validation.

\noindent\textbf{S3DIS Full}~\cite{s3dis} is a collection of 6 large scale scenes covering 271 rooms. it provides annotations for 13 semantic classes. Following prior works ~\cite{Roh_2024_CVPR_EASE, AGILE3D, POINT-SAM} in instance segmentation, we use \textit{Area\_5} for evaluation which contains 68 samples for validation.

\noindent\textbf{S3DIS Crops} is proposed by AGILE3D~\cite{AGILE3D} in their evaluation setting, cropping the validation samples from the original S3DIS dataset around each instance into 3m $\times$ 3m blocks. They provide the processed data on their github \href{https://drive.google.com/file/d/1cqWgVlwYHRPeWJB-YJdz-mS5njbH4SnG/view?usp=sharing}{here}. We call this dataset variant S3DIS Crops.

\subsubsection{Outdoor Datasets} 
Outdoor datasets include two types of classes: \textit{things}, which have instance labels, and \textit{stuff}, which do not. This distinction can hinder the effectiveness of a promptable segmentation model. 
To address this, we use an off-the-shelf clustering algorithm, HDBSCAN~\cite{mcinnes2017hdbscan}, to provide us with pseudo instance labels for the \textit{stuff} classes, enabling instance-wise promptable training on these categories.
Details about HDBSCAN are provided in \S~\ref{sec:hdbscan_details}.

\noindent\textbf{SemanticKITTI}~\cite{semantickitti} is derived from the KITTI Odometry \cite{kitti} datasets. Each point in the dataset is densely labeled with one of $C=19$ classes divided into \textit{things} (with instance labels) and \textit{stuff} (without instance labels) classes. We run HDBSCAN~\cite{mcinnes2017hdbscan} to generate instance labels for the \textit{stuff} classes.

\noindent\textbf{nuScenes} \cite{nuscenes} is a comprehensive dataset that includes over 1000 diverse driving records, each lasting around 20 seconds. The LiDAR data from nuScenes is densely annotated with $C = 32$ classes, again divided into \textit{things} and \textit{stuff} classes. We run HDBSCAN~\cite{mcinnes2017hdbscan} to generate instance labels for the \textit{stuff} classes.

\noindent\textbf{PandaSet}~\cite{pandaset} is an autonomous driving dataset featuring 103 driving sequences lasting about 8 seconds each. We only use a subset of 39 out of 103 scenes as the original dataset download links have expired. The dataset used for our training can be found at \href{https://www.kaggle.com/datasets/usharengaraju/pandaset-dataset}{Kaggle}.
It provides semantic annotations for 37 classes divided into 28 \textit{things} and 9 \textit{stuff} classes. We run HDBSCAN~\cite{mcinnes2017hdbscan} to generate instance labels for the \textit{stuff} classes.

\noindent\textbf{KITTI-360 Full}~\cite{kitti360} is a large-scale outdoor driving dataset which provides 360-degree annotations on point clouds, including bounding boxes, semantic, and instance annotations. The original dataset only provides labels for down-sampled superimposed point clouds. We call this original version KITTI-360 Full. This dataset provides annotations on 37 semantic classes in 19 object categories. 

\noindent\textbf{KITTI-360 Single Scan} is derived from the KITTI-360~\cite{kitti360} Full dataset by following Interactive4D~\cite{interactive4D}, where we applied a nearest-neighbor algorithm to propagate labels to individual points in individual scans. We use publicly available scripts for this purpose \href{https://github.com/JulesSanchez/recoverKITTI360label}{(Sanchez, 2021)}. We call this derived version KITTI-360 Single Scan. This also contains annotations for 37 semantic classes in 19 object categories. Since \shortname\ is trained using HDBSCAN-based instance labels for \textit{stuff} classes, it would constitute a different evaluation setting if we evaluate on all classes. To keep evaluations fair, we only evaluate on the 11 \textit{things} classes and compare against baselines.

\noindent\textbf{KITTI-360 Crops} is also derived from KITTI-360~\cite{kitti360} Full dataset. Specifically, to keep consistent with prior works like AGILE3D~\cite{AGILE3D}, which evaluated their indoor models on this variant, we also use the cropped version from their provided list of preprocessed scenes. This preprocessing includes dividing the original superimposed point clouds into smaller 3m $\times$ 3m chunks centered around the object instance.

\noindent\textbf{Waymo}~\cite{Sun_2020_CVPR} is a large-scale outdoor driving dataset which provides semantic labels across 23 classes but does not provide any instance annotations. However, Waymo does provide bounding boxes for 4 classes, including \textit{vehicle, cyclist, sign}, and \textit{pedestrian}. We use the combination of bounding boxes and semantic labels to generate instance labels for these 4 classes. After preprocessing on the entire validation set of Waymo, we get 5,976 samples for validation. Our preprocessed dataset will be released for reproducibility. 

\subsubsection{Aerial Datasets} 

\noindent\textbf{STPLS3D}~\cite{stpls3d} is a large-scale photogrammetry point cloud dataset covering approximately 16$\text{km}^2$ of urban and rural areas in Malaysia. Released in 2020, it contains over 2 billion labeled points across 25 scenes with annotations for 14 semantic classes. To keep computational demands tractable, we crop the point clouds to 50m $\times$ 50m blocks. The dataset is generated from aerial imagery using photogrammetric techniques, providing dense colored point clouds.

\noindent\textbf{DALES}~\cite{dales_instance}  is a large-scale aerial LiDAR dataset covering 10$\text{km}^2$ of diverse landscapes, including urban, suburban, rural, and forested areas. It contains over 505 million points manually annotated with 8 semantic classes. Since DALES does not provide instance annotations, we again employ HDBSCAN~\cite{mcinnes2017hdbscan} to generate instance labels for training and validation. We also crop the point clouds to 50m $\times$ 50m blocks. The dataset provides high-density aerial LiDAR data (50 points per $\text{m}^2$) captured at varying altitudes, making it particularly challenging due to large variations in point density and object scales. 

\noindent\textbf{UrbanBIS}~\cite{UrbanBIS} is a dataset for large-scale 3D urban understanding, supporting practical urban-level semantic and building-level instance segmentation. UrbanBIS comprises six real urban scenes, with 2.5 billion points, covering a vast area of 10.78$\text{km}^2$ and 3,370 buildings, captured by 113,346 views of aerial photogrammetry. It provides annotations on 6 scenes, out of which we evaluate on the \textit{Yingrenshi} test scenes. After cropping to 50m $\times$ 50m blocks, this provides us with 46 validation samples.

\subsection{HDBSCAN Details}
\label{sec:hdbscan_details}
For the outdoor datasets used to train SNAP-C, the datasets include two types of classes: \textit{things}, which have instance labels and \textit{stuff} which do not have instance labels. From a promptable segmentation perspective, instance labels from \textit{things} classes fit in directly. However, \textit{stuff} includes classes such as vegetation, roads, buildings, \etc, and is assigned a single label for all of them. The objects from these classes can be far away from each other and thus using one label directly is counterproductive in training a promptable segmentation model. To solve this issue, we propose to preprocess the datasets with HDBSCAN \cite{mcinnes2017hdbscan}. Specifically, we first take all the points belonging to a \textit{stuff} class, and apply clustering on it. This helps in making multiple clusters from single class labels, which can then be used for promptable segmentation training. 

\subsection{Training and Inference Details}
\label{sec:training_details}
All \shortname\ variants are trained for 100 epochs using 8$\times$NVIDIA A6000 GPUs. We train with a batch size of 1, where each batch corresponds to a single point cloud, from which 32 objects are randomly sampled for supervision. During training, we set the maximum click budget to 10. In each iteration, the number of clicks is randomly sampled between 1 and 10, ensuring that the model is consistently exposed to varying levels of user interaction. We use mixed-precision training to speed up both the training and evaluation process. We employ a round-robin style multi-dataset dataloader that repeats smaller datasets multiple times to keep the sample count similar to large datasets. During training, this dataloader provides a point cloud from one of the datasets at each training iteration, with each batch containing samples from a single dataset. To ensure proper routing through the correct normalization layer, we follow \cite{pdnorm} and attach a \textit{domain} variable to each point cloud. During training, this approach allows the network to route the data to the correct normalization layer. For interactive inference, this functionality translates into a simple domain-type checkbox selection, making it highly user-friendly. 

\begin{figure*}[ht]
    \centering
    \setlength{\tabcolsep}{1pt}
    \setlength{\fboxsep}{0pt}  %
    \setlength{\fboxrule}{1pt} %
    \begin{small}
    \begin{tabular}{cc}
        \fbox{\includegraphics[width=0.48\linewidth]{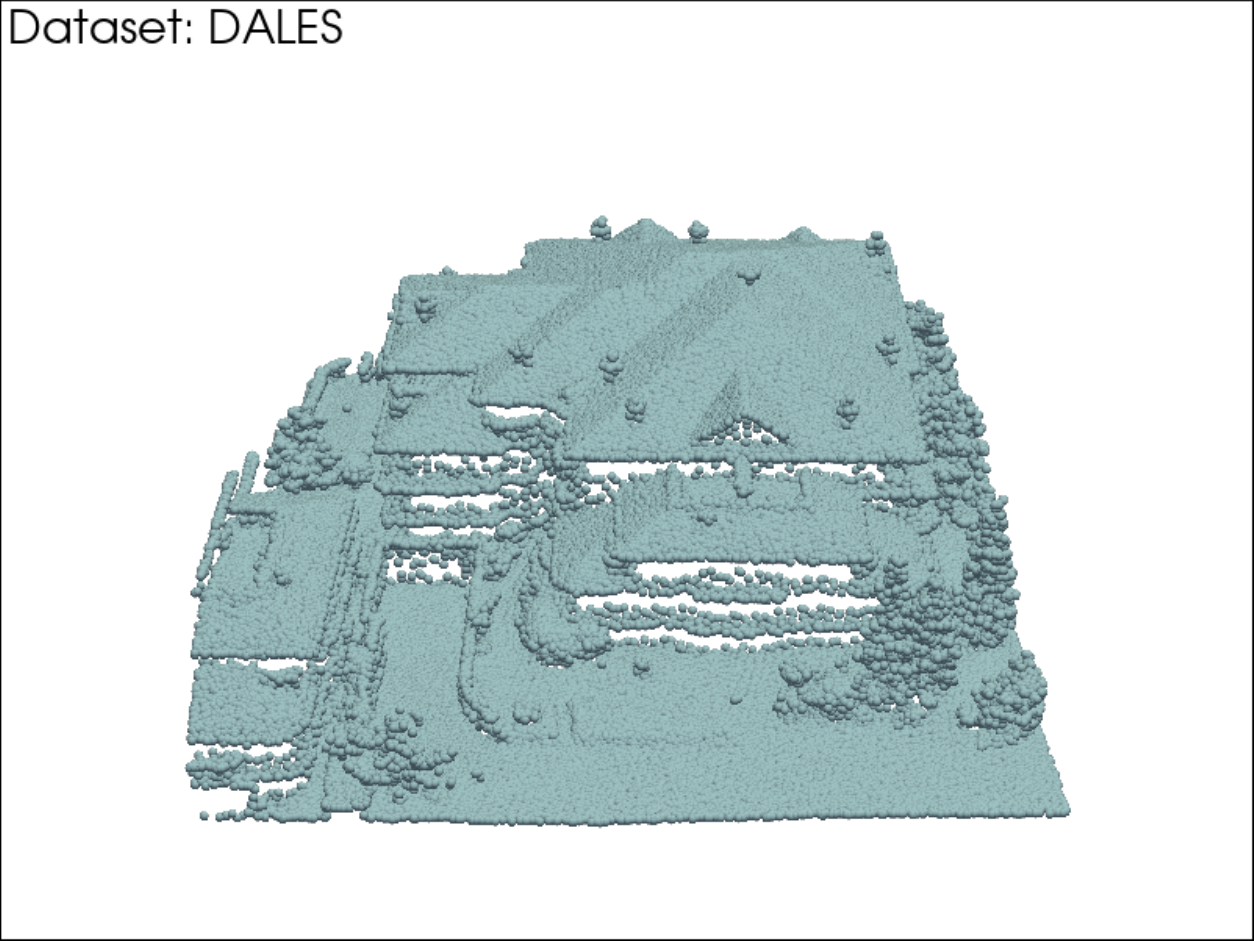}} & 
        \fbox{\includegraphics[width=0.48\linewidth]{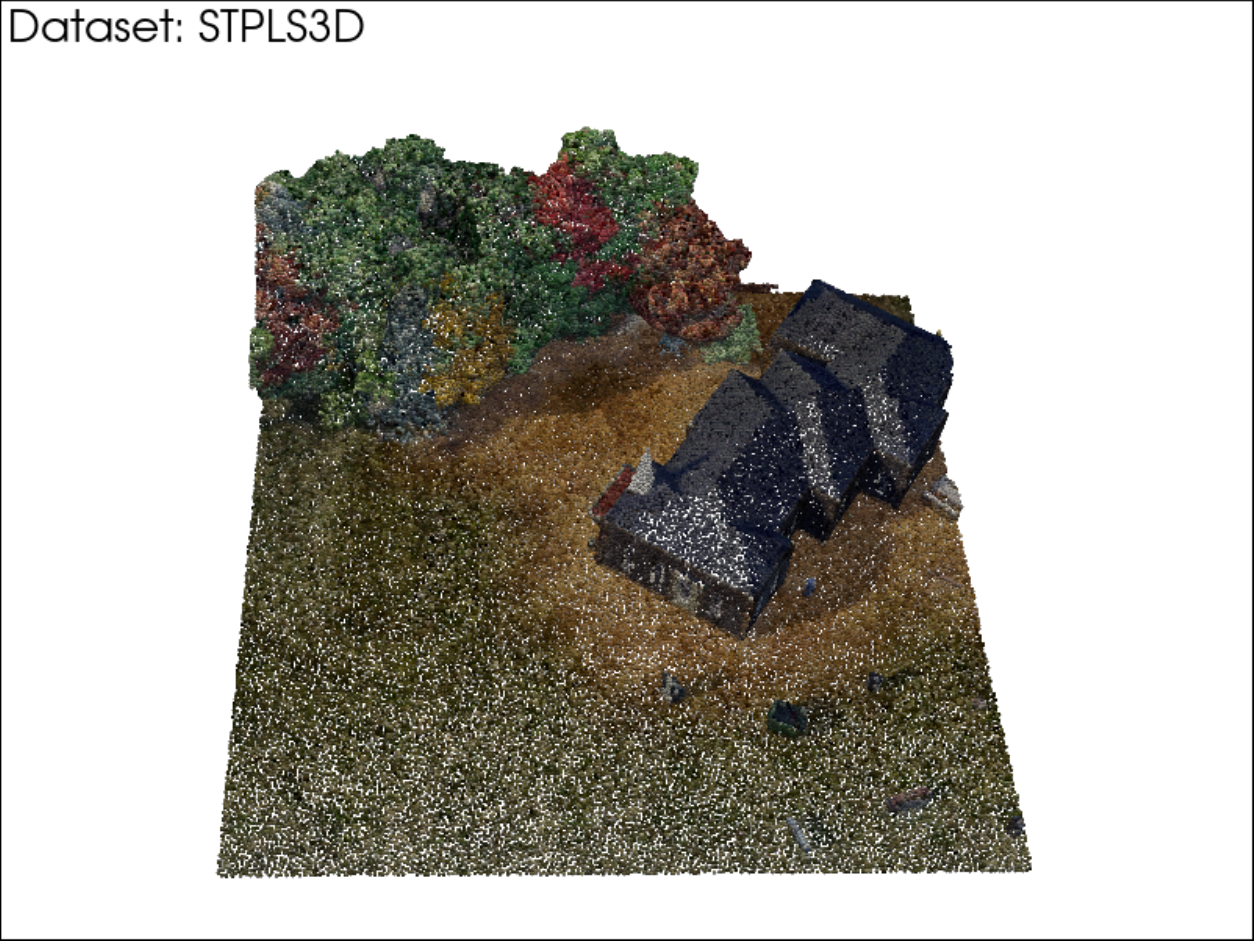}} \\
        
        \fbox{\includegraphics[width=0.48\linewidth]{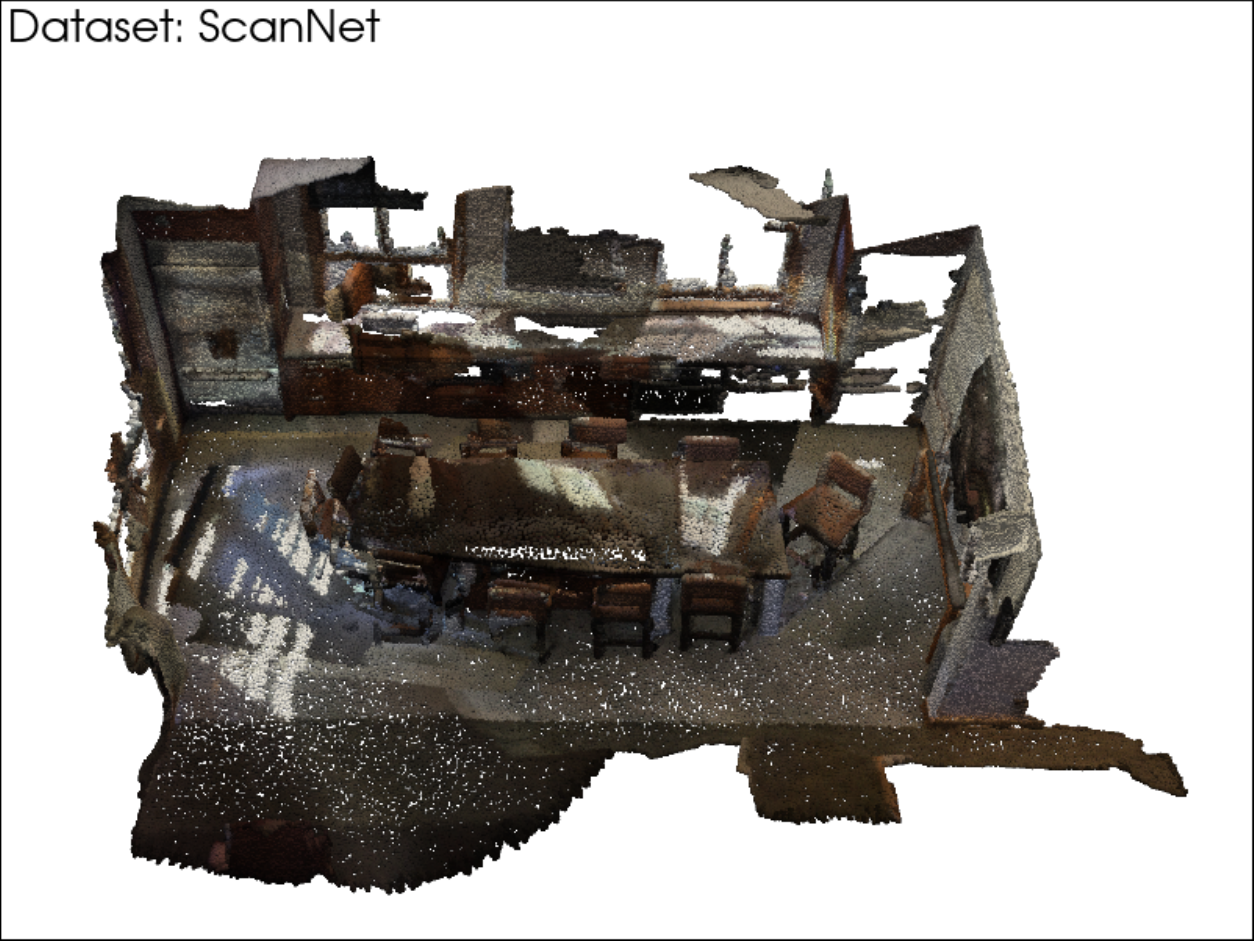}} & 
        \fbox{\includegraphics[width=0.48\linewidth]{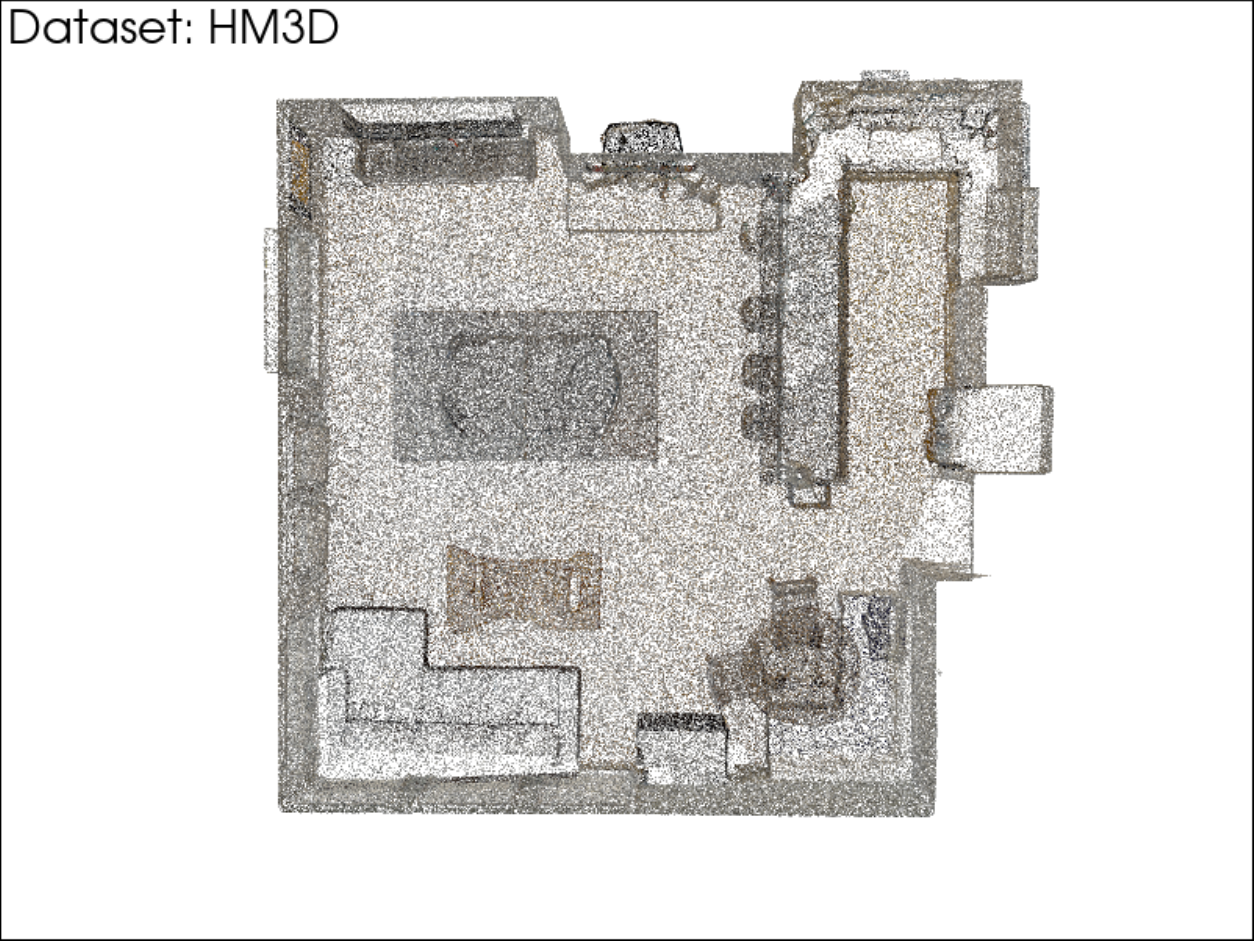}} \\
        
        \fbox{\includegraphics[width=0.48\linewidth]{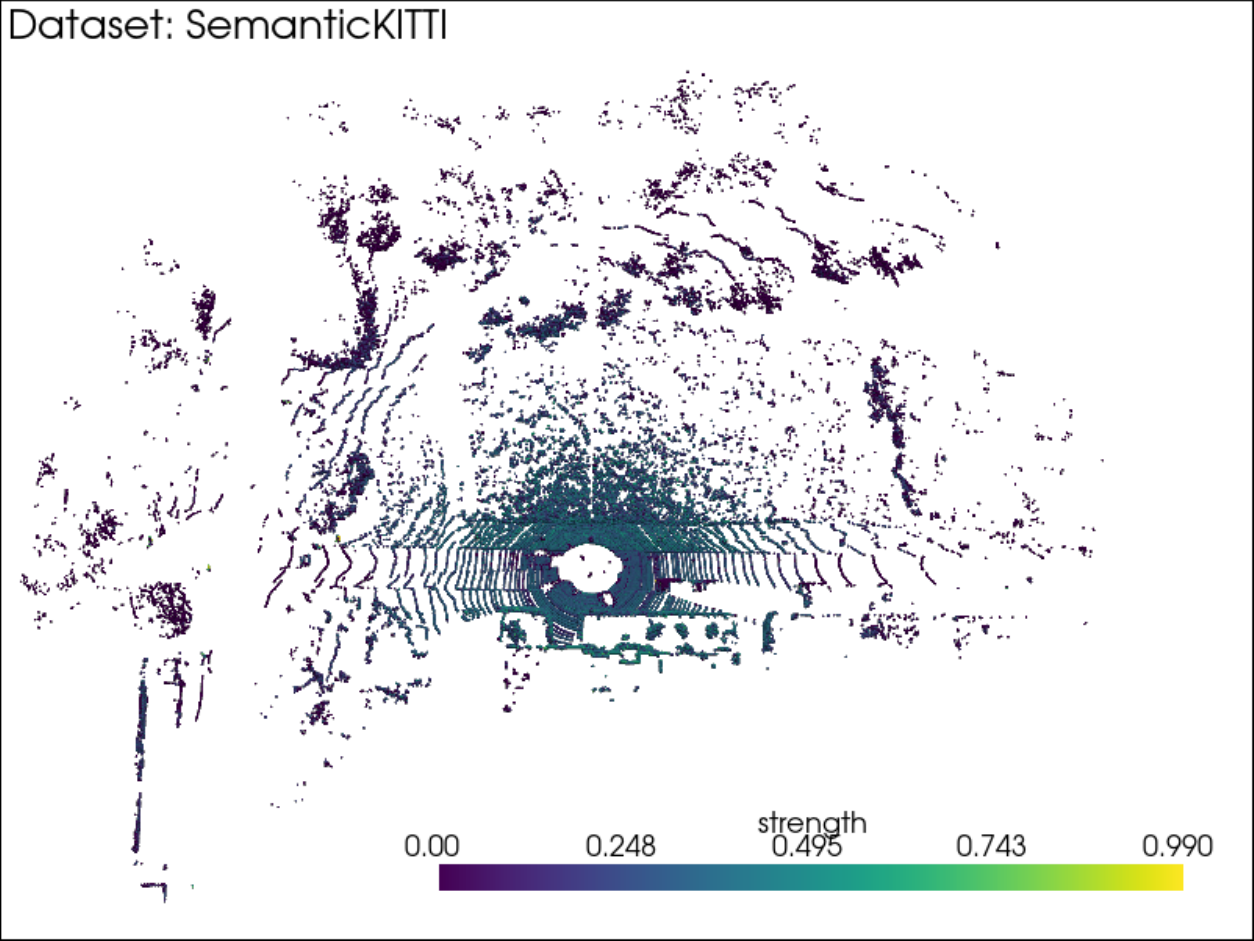}} & 
        \fbox{\includegraphics[width=0.48\linewidth]{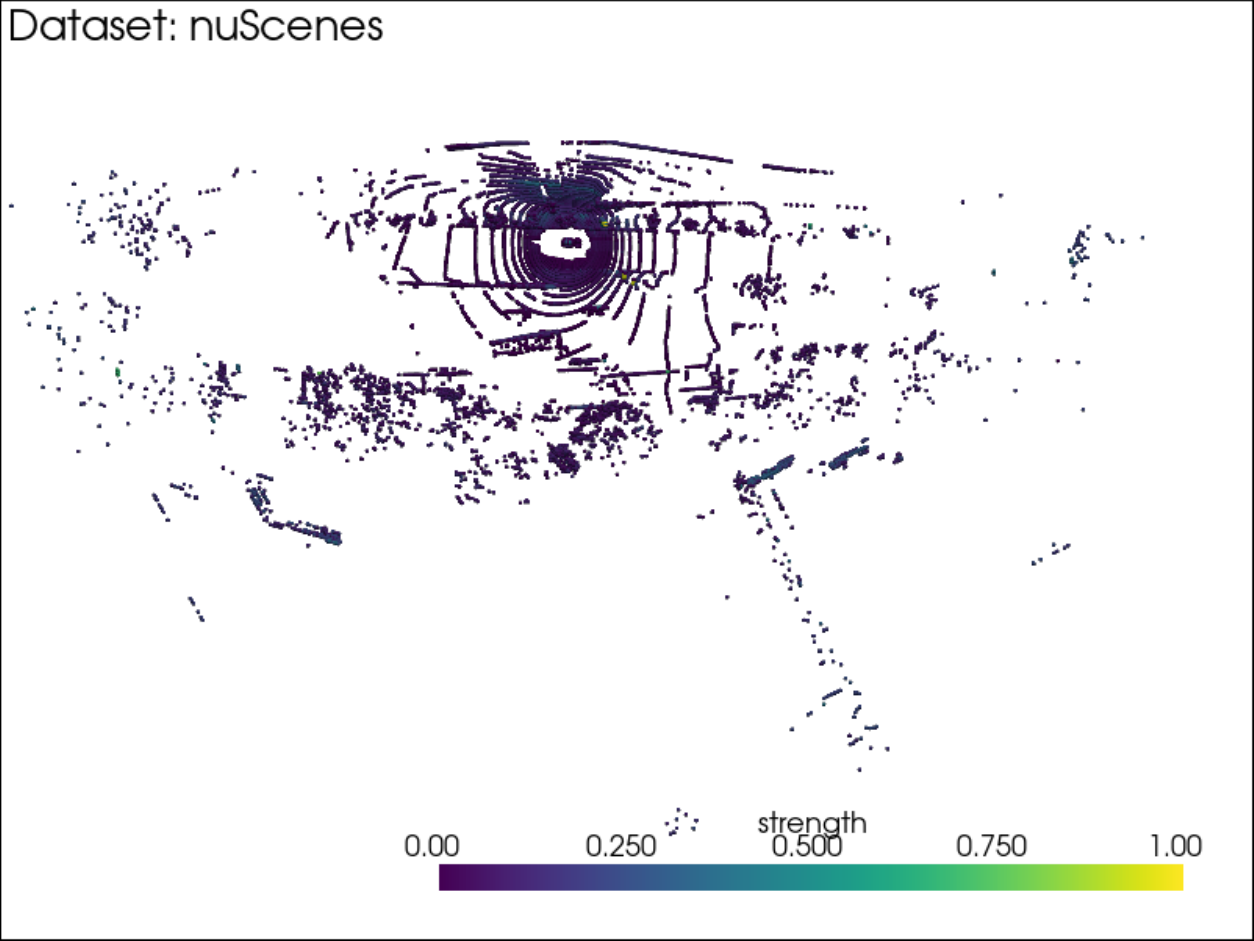}} \\
    \end{tabular}
    \end{small}
    \caption{\textbf{Samples from Training Datasets.} Here we present samples taken from the training datasets to showcase the difference in scale, point density and scene types. Note that the dataset name is displayed in each figure. From the top - DALES and STPLS3D are aerial datasets, ScanNet and HM3D are indoor scene datasets and SemanticKITTI and nuScenes are outdoor scene datasets. HM3D provides full room scenes, point size has been reduced for better understanding of the scene.}
    \label{fig:train_data_examples}
\end{figure*}

\begin{figure*}[ht]
    \centering
    \setlength{\tabcolsep}{1pt}
    \setlength{\fboxsep}{0pt}  %
    \setlength{\fboxrule}{1pt} %
    \begin{small}
    \begin{tabular}{cc}
        \fbox{\includegraphics[width=0.48\linewidth]{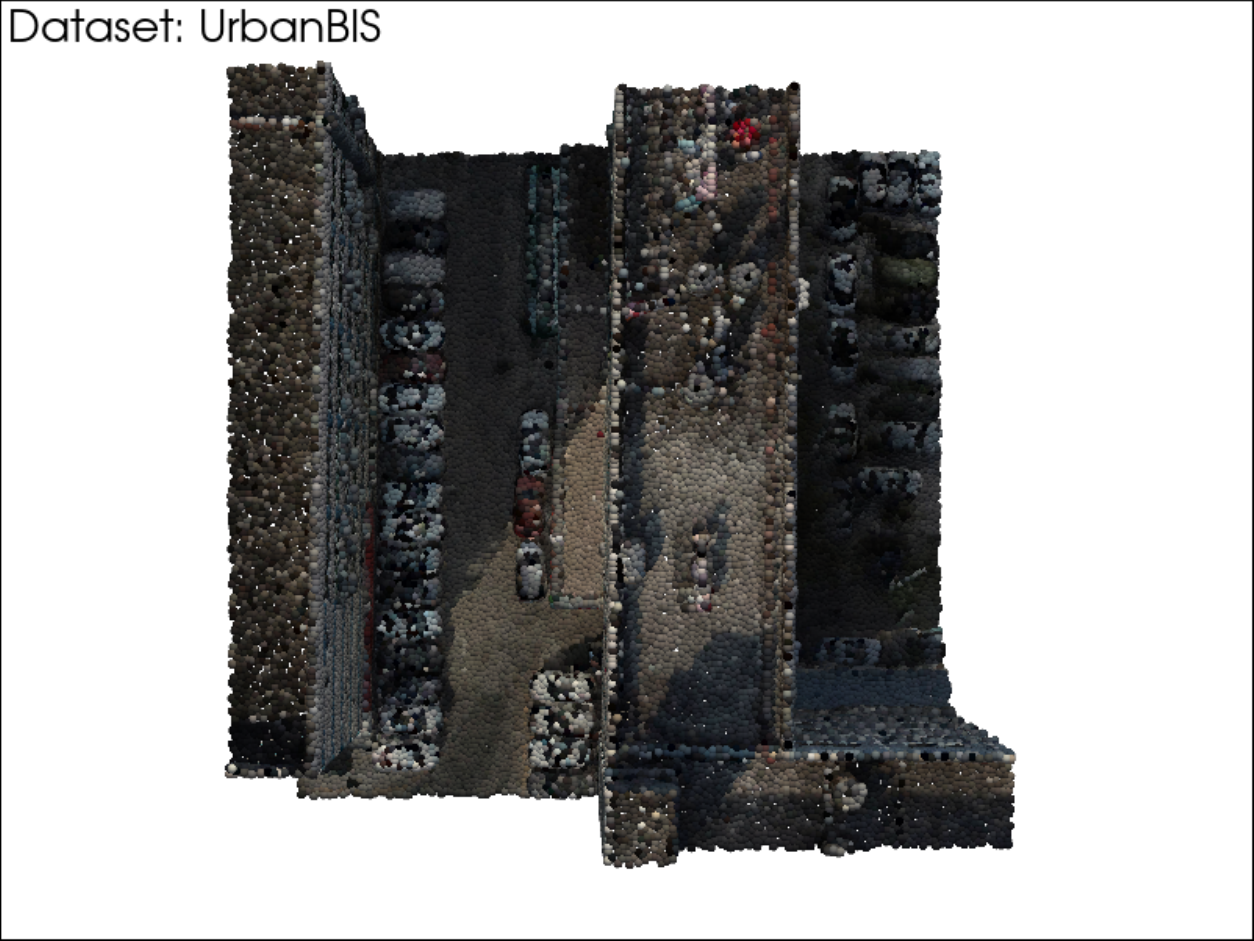}} &
        \fbox{\includegraphics[width=0.48\linewidth]{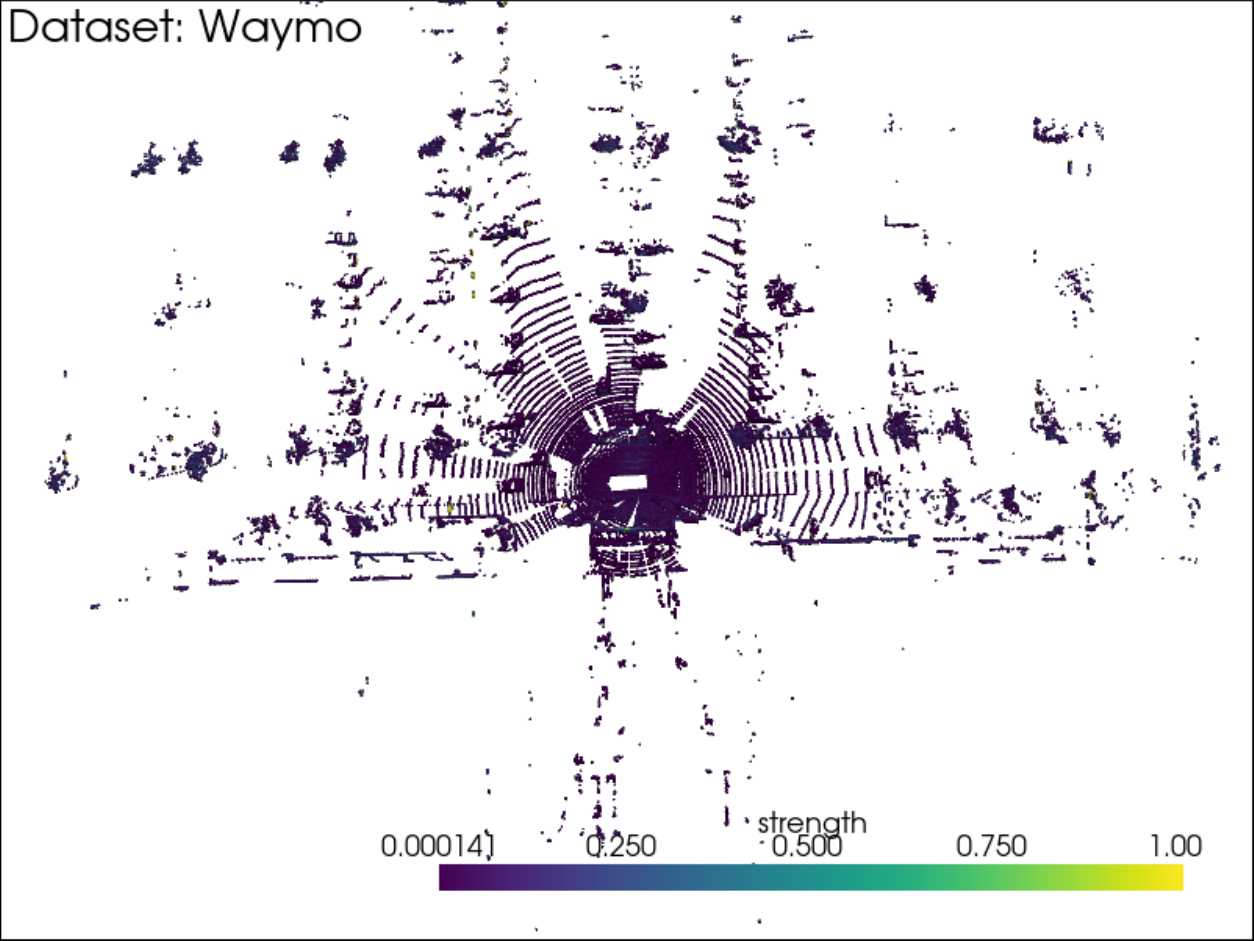}} \\
        
        \fbox{\includegraphics[width=0.48\linewidth]{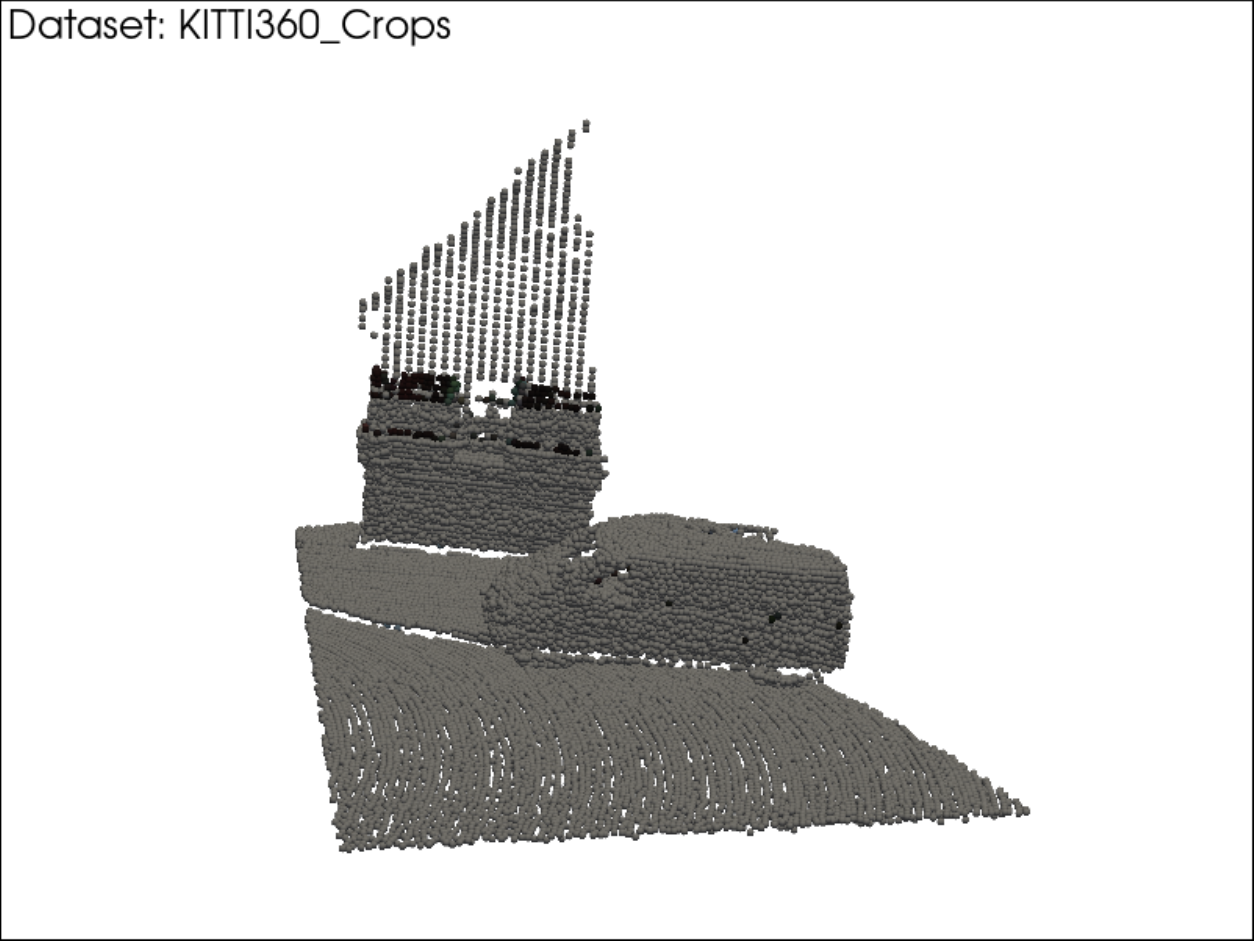}} & 
        \fbox{\includegraphics[width=0.48\linewidth]{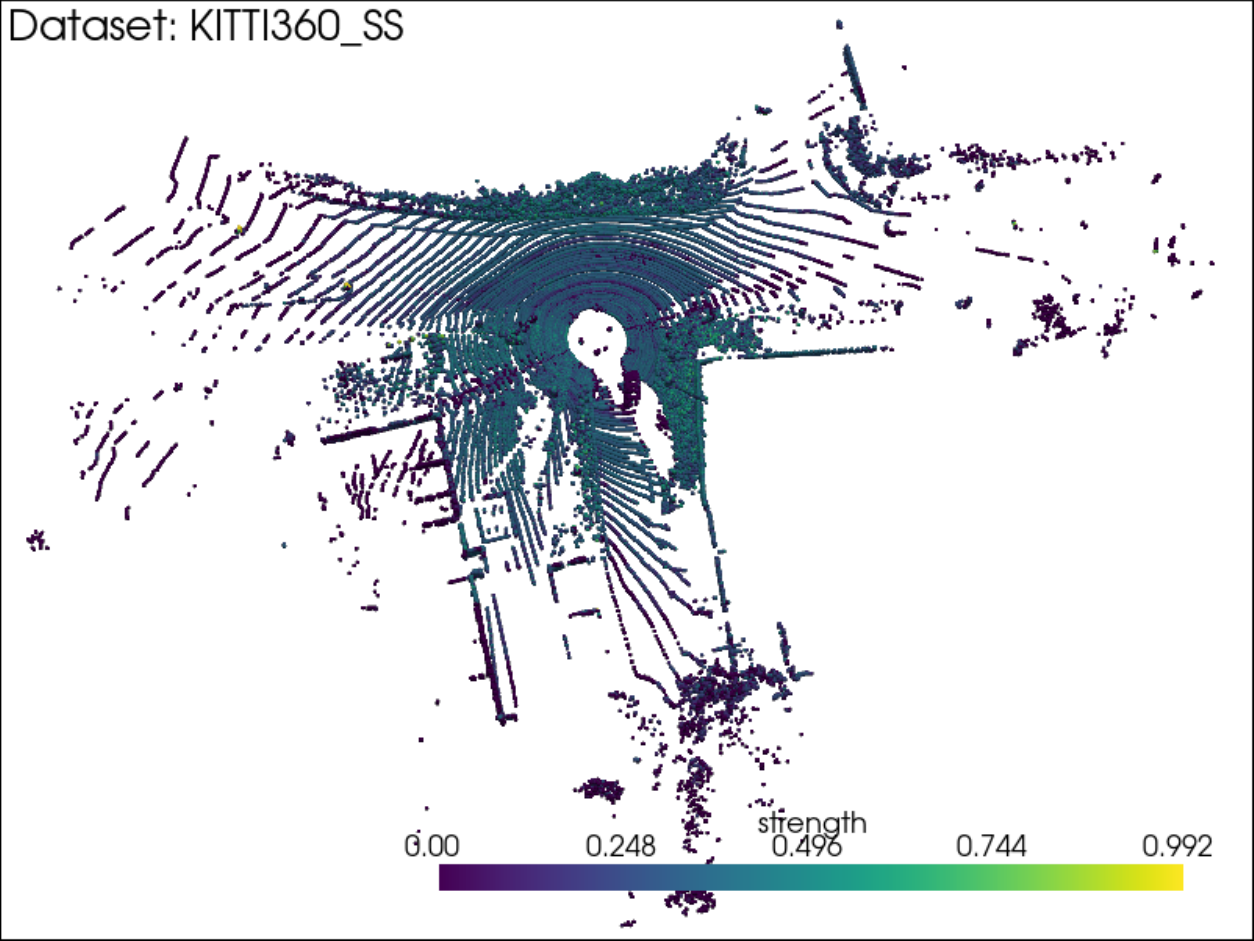}} \\

        \multicolumn{2}{c}{{\setlength{\fboxsep}{1pt}\fbox{\includegraphics[width=0.95\linewidth]{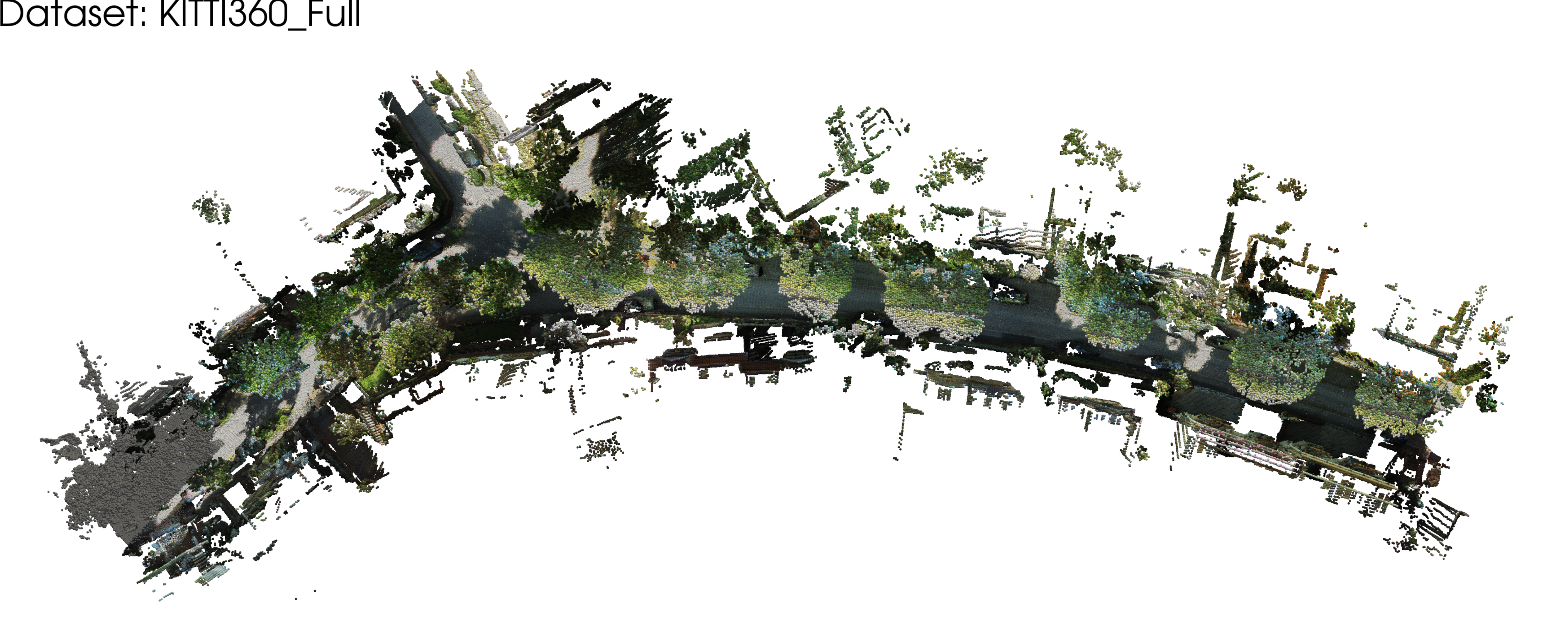}}}} \\

    \end{tabular}
    \end{small}
    \caption{\textbf{Samples from Validation Datasets.} We evaluate \shortname\ on a variety of datasets. From the top - UrbanBIS is an Aerial scene; Waymo, KITTI-360 Crops, KITTI-360 Single Scan, and KITTI-360 Full are outdoor scene datasets. Note the difference in the variants of KITTI-360. KITTI-360 Crops particularly represents small-scale dense scenes generally found in indoor point clouds, while KITTI-360 Single Scan shows a traditional point cloud collected with a LiDAR sensor, and KITTI-360 Full shows an aggregated point cloud map built by combining multiple LiDAR scans.}
    \label{fig:val_data_examples_1}
\end{figure*}

\begin{figure*}[ht]
    \centering
    \setlength{\tabcolsep}{1pt}
    \setlength{\fboxsep}{0pt}  %
    \setlength{\fboxrule}{1pt} %
    \begin{small}
    \begin{tabular}{cc}
        \fbox{\includegraphics[width=0.48\linewidth]{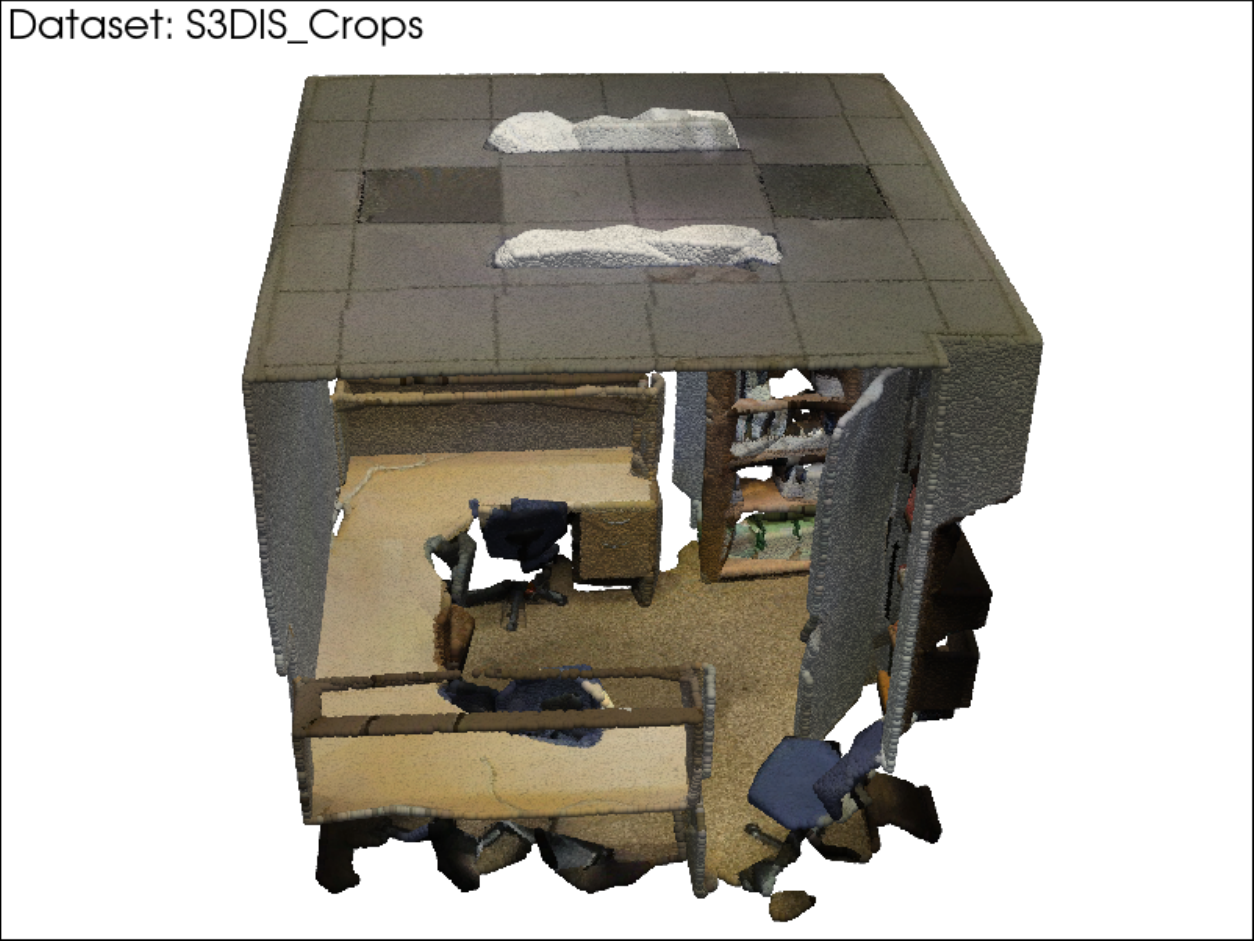}} & 
        \fbox{\includegraphics[width=0.48\linewidth]{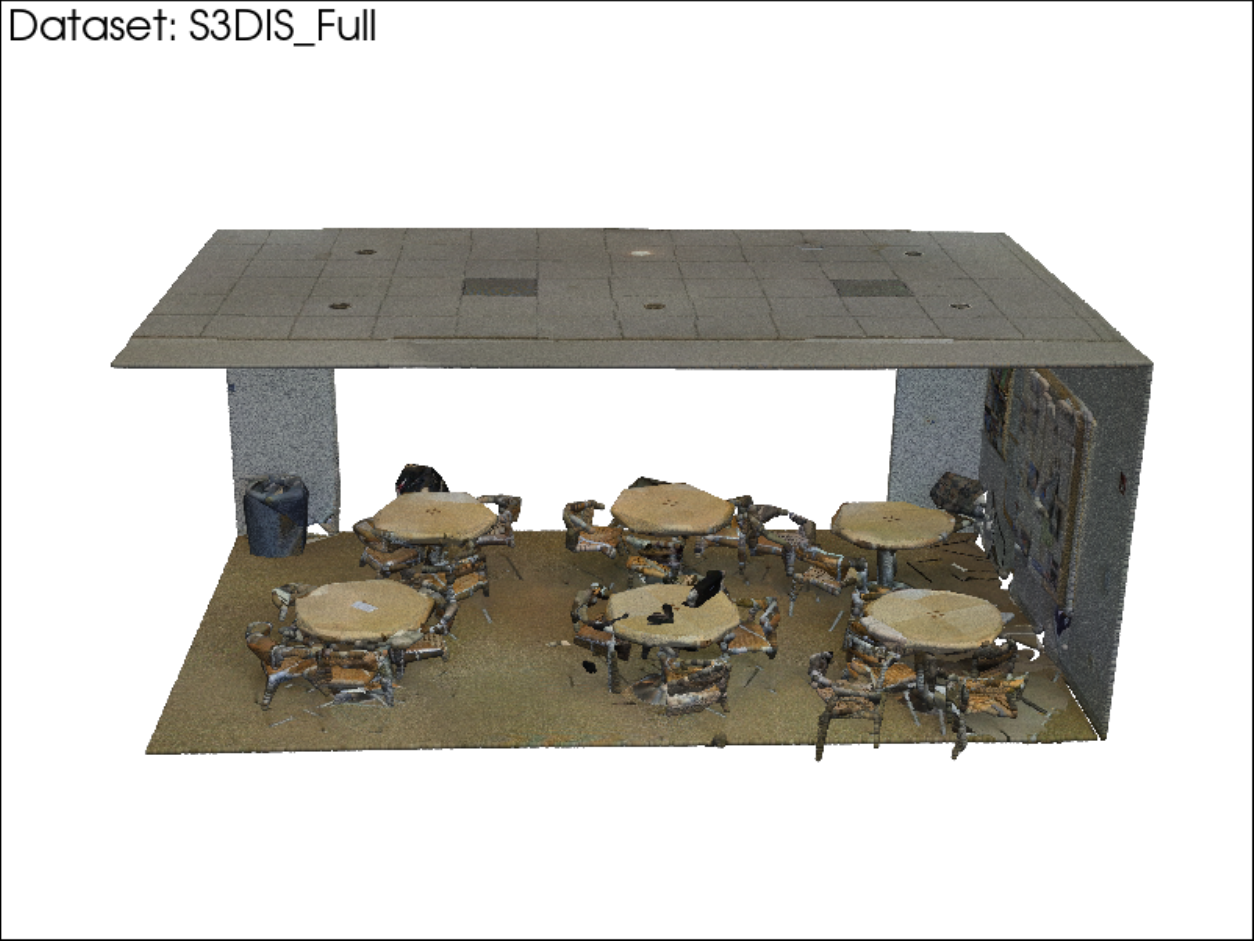}} \\
        
        \fbox{\includegraphics[width=0.48\linewidth]{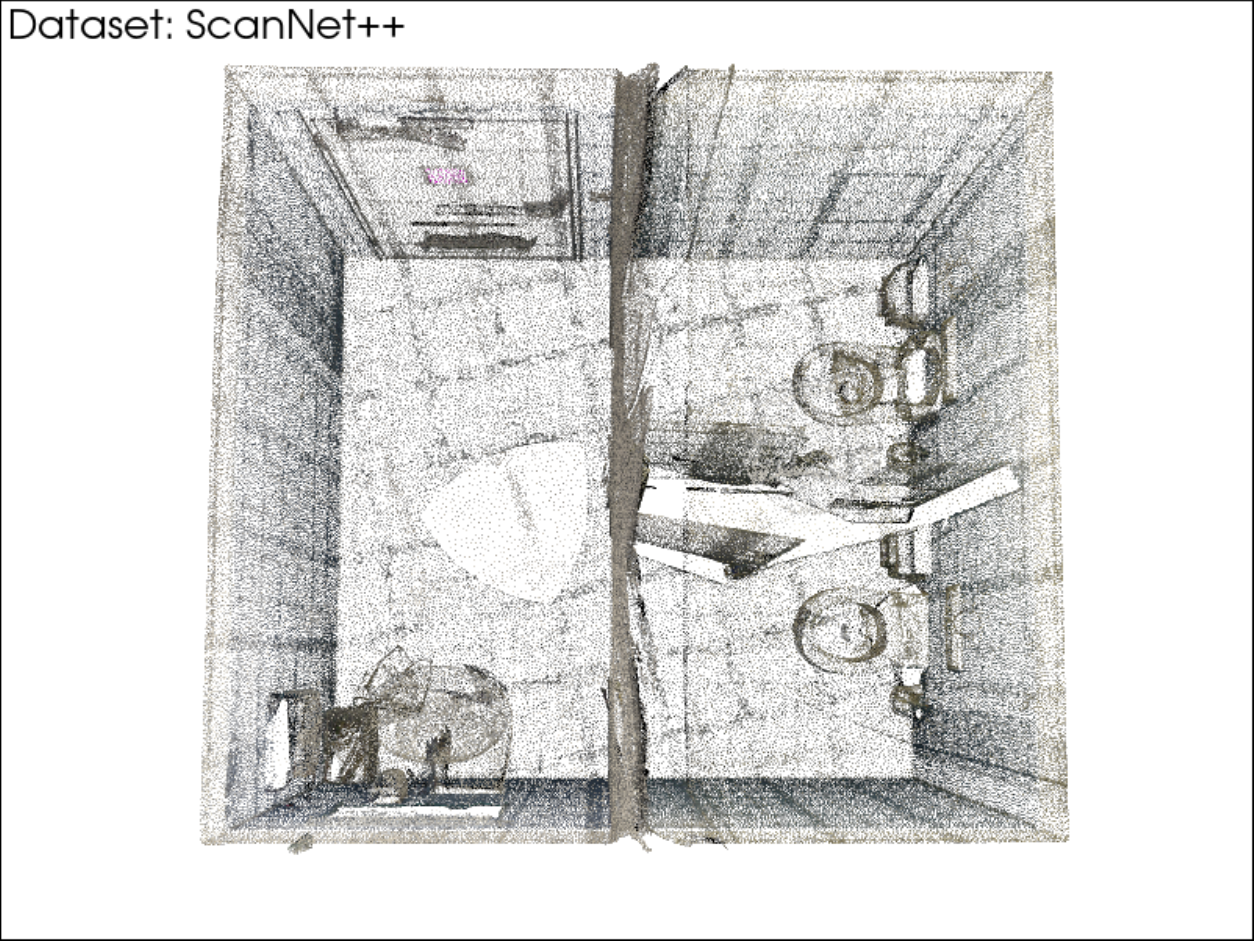}} & 
        \fbox{\includegraphics[width=0.48\linewidth]{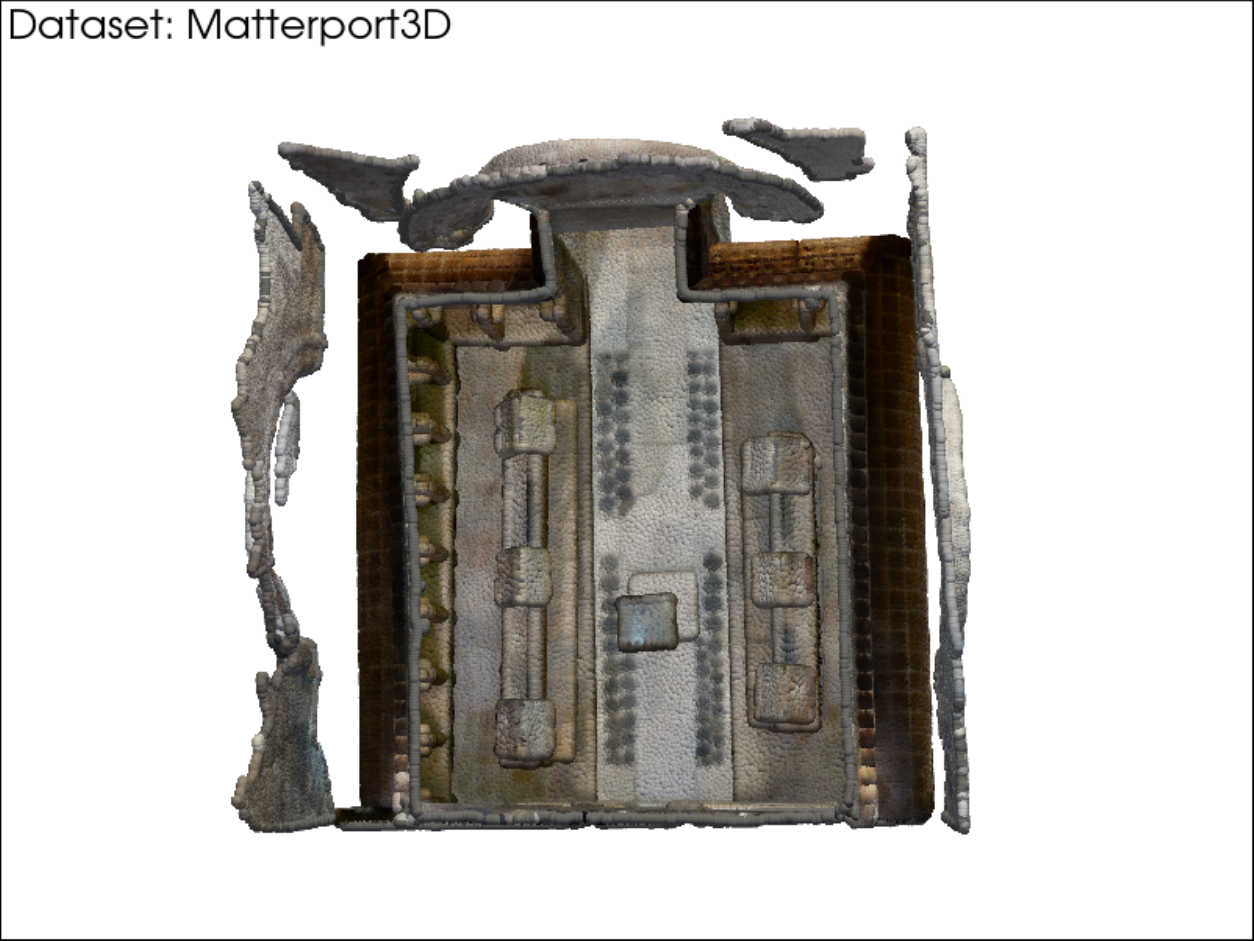}} \\
    \end{tabular}
    \end{small}
    \caption{\textbf{Samples from Validation Datasets.} Here we show the examples from Indoor datasets used for validation. Note that S3DIS Crops is a cropped version of the full S3DIS point clouds, therefore some objects appear truncated. ScanNet++ provides full room scenes, point size has been reduced for better understanding of the scene.}
    \label{fig:val_data_examples_2}
\end{figure*}

\section{Additional Ablations}
\label{sec:additional_ablations}

\subsection{Effect of Backbone Architecture}
\label{sec:backbone_ablation}
\begin{table}[tb]
\centering
\caption{\textbf{Backbone Ablation on the ScanNet dataset.} Note that memory and time statistics are reported for 1-Click experiments. 
}
\vspace{-0.8em}
\small
\label{table:ablation_backbone}
\setlength{\tabcolsep}{3.5pt}
\begin{tabular}{l|c c c|c|c}
\toprule
\multirow{2}{*}{Backbone} & \multicolumn{3}{c}{${\text{IoU}}@k \uparrow$}  & \multirow{2}{*}{\makecell{Memory\\@1 click}} & \multirow{2}{*}{\makecell{Time\\@1 click}} \\
\cmidrule{2-4}
 & @1 & @5 & @10 & & \\
\midrule
AGILE3D~\cite{AGILE3D} & 63.3 & 79.9 & 83.7 & 1.2 GB  & 203 ms \\
Minkowski~\cite{minkowski} & 68.4 & 82.2 & 83.4 & 1.8 GB & 213 ms \\
PTv3~\cite{ptv3}      & 68.6 & 82.1 & 84.6 & 1.3 GB & 197 ms \\
\bottomrule
\end{tabular}
\end{table}

\noindent We use the PTv3~\cite{ptv3} backbone for feature extraction, but a natural question to ask is, \textquotedblleft How is the model performance affected if we use a different backbone?\textquotedblright~To answer this, we compare PTv3~\cite{ptv3} with the Minkowski Res16UNet34C\cite{minkowski} backbone, which has been employed by~\cite{InterObject3D, AGILE3D, interactive4D}. The comparison, conducted on the ScanNet dataset, is summarized in \textbf{Tab.~\ref{table:ablation_backbone}}. 
We observe consistent improvement in both PTv3 and Minkowski Engine backbones against AGILE3D~\cite{AGILE3D},
showing that our approach is equally applicable across both recent transformer-based as well as the common sparse-convolution-based backbones. To compute the memory and timing requirements, we use a random uniform point cloud with 100,000 points on all methods.

\subsection{Effect of Click Strategy}
\label{sec:click_strategy_ablation}
\begin{table}[tb]
\centering
\small
\caption{\textbf{Inference Click Strategy Ablations.} We evaluate different click strategies on the ScanNet20 dataset. Random Sampling represents all click points sampled randomly on the target object. Iterative sampling represents additional click points sampled in the unsegmented region from previous click mask.}
\vspace{-0.8em}
\label{table:ablation_click_strategy}
\begin{tabular}{l|c c|c}
\toprule
Strategy & \multicolumn{2}{c}{${\text{IoU}}@k \uparrow$} & Time (ms) \\
\midrule
\multirow{3}{*}{\makecell{Random\\Sampling}}
 & @1  & 68.6 & 170 \\
 & @5  & 79.4 & 178 \\
 & @10 & 80.5 & 185 \\
\midrule
\multirow{3}{*}{\makecell{Iterative\\Sampling}}
 & @1  & 68.6 & 170 \\
 & @5  & 82.3 & 190 \\
 & @10 & 85.5 & 211 \\
\bottomrule
\end{tabular}
\end{table}

\noindent We evaluate two click strategies during inference: (1) Random-sampling and (2) Iterative Refinement. The results are shown in \textbf{Tab.~\ref{table:ablation_click_strategy}}, with timing measurements obtained by running the evaluation on a single NVIDIA RTX 3090 GPU.
While Iterative Refinement performs much better than random sampling, it also runs slower in comparison. The random sampling strategy is especially helpful for users when trying to segment objects in the scene, because users can give multiple clicks at the beginning (which is equivalent to random sampling) to get a high-quality mask and later use refinement clicks to further improve the mask quality. In \shortname, we provide the flexibility to use both approaches during inference. To compute inference time, we used a point cloud from the ScanNet dataset with about 80,000 points.

\subsection{Cross-Domain Input Ablation}
\label{sec:performance_with_cross_domain_inputs}
\begin{table*}[tb]
\centering
\small
\vspace{3em}
\caption{\textbf{Effect of Cross-Domain Selection in Domain Normalization.} Evaluation results of \shortname-C when applying different domain types (Indoor, Outdoor, Aerial) in Domain Norm. The domain used for normalization is indicated in \textcolor{teal}{\textit{teal}} beneath each result.
}
\label{table:supp_effect_evaluating_cross_domain}
\setlength{\tabcolsep}{4.5pt}
\begin{tabular}{l|c c c c c c|c c c c c c c c c c}
\toprule
\multirow{5}{*}{Model} & \multicolumn{16}{c}{IoU @ k} \\
\cmidrule{2-17}
\addlinespace[-2.5pt]
& \multicolumn{6}{c}{\cellcolor{mediumblue!20}In-Distribution} & \multicolumn{10}{c}{\cellcolor{green!20}Zero-Shot} \\
\addlinespace[-2pt]
\cmidrule{2-17} 
& \multicolumn{2}{c}{SemanticKITTI} & \multicolumn{2}{c}{ScanNet20} & \multicolumn{2}{c}{STPLS3D} & \multicolumn{2}{c}{Matterport3D} & \multicolumn{2}{c}{S3DIS Full} & \multicolumn{2}{c}{KITTI-360 Full} & \multicolumn{2}{c}{Waymo} & \multicolumn{2}{c}{UrbanBIS} \\
\cmidrule(lr){2-3} \cmidrule(lr){4-5} \cmidrule(lr){6-7} \cmidrule(lr){8-9} \cmidrule(lr){10-11} \cmidrule(lr){12-13} \cmidrule(lr){14-15} \cmidrule(lr){16-17}
& @1 & @5 & @1 & @5 & @1 & @5 & @1 & @5 & @1 & @5 & @1 & @5 & @1 & @5 & @1 & @5 \\
\midrule

SNAP - C & \textbf{71.5} & \textbf{86.0} & 19.2 & 52.3 & 41.4 & 66.9 & 17.4 & 45.6 & 15.7 & 51.3 & \textbf{23.1} & \textbf{48.1} & \textbf{69.8} & \textbf{86.6} & 55.2 & 80.1 \\
\textit{\textcolor{gray}{Norm used}} & \multicolumn{2}{c}{\textit{\textcolor{teal}{Outdoor}}} & \multicolumn{2}{c}{\textit{\textcolor{teal}{Outdoor}}} & \multicolumn{2}{c}{\textit{\textcolor{teal}{Outdoor}}} & \multicolumn{2}{c}{\textit{\textcolor{teal}{Outdoor}}} & \multicolumn{2}{c}{\textit{\textcolor{teal}{Outdoor}}} & \multicolumn{2}{c}{\textit{\textcolor{teal}{Outdoor}}} & \multicolumn{2}{c}{\textit{\textcolor{teal}{Outdoor}}} & \multicolumn{2}{c}{\textit{\textcolor{teal}{Outdoor}}}\\ 

SNAP - C & 7.2 & 19.1 & \textbf{67.7} & \textbf{82.3} & 4.9 & 6.6 & \textbf{52.6} & \textbf{75.2} & \textbf{53.6} & \textbf{77.6} & 2.4 & 4.1 & 1.2 & 7.8 & 19.9 & 27.9 \\
\textit{\textcolor{gray}{Norm used}} & \multicolumn{2}{c}{\textit{\textcolor{teal}{Indoor}}} & \multicolumn{2}{c}{\textit{\textcolor{teal}{Indoor}}} & \multicolumn{2}{c}{\textit{\textcolor{teal}{Indoor}}} & \multicolumn{2}{c}{\textit{\textcolor{teal}{Indoor}}} & \multicolumn{2}{c}{\textit{\textcolor{teal}{Indoor}}} & \multicolumn{2}{c}{\textit{\textcolor{teal}{Indoor}}} & \multicolumn{2}{c}{\textit{\textcolor{teal}{Indoor}}} & \multicolumn{2}{c}{\textit{\textcolor{teal}{Indoor}}}\\ 

SNAP - C & 27.1	& 57.9 & 11 & 22.1 & \textbf{67.8} & \textbf{80.4} & 12.6 & 22.3 & 8.2 & 25.6 & 6.8 & 28.3 & 25.1 &	60.4 & \textbf{71.6} & \textbf{90.2} \\
\textit{\textcolor{gray}{Norm used}} & \multicolumn{2}{c}{\textit{\textcolor{teal}{Aerial}}} & \multicolumn{2}{c}{\textit{\textcolor{teal}{Aerial}}} & \multicolumn{2}{c}{\textit{\textcolor{teal}{Aerial}}} & \multicolumn{2}{c}{\textit{\textcolor{teal}{Aerial}}} & \multicolumn{2}{c}{\textit{\textcolor{teal}{Aerial}}} & \multicolumn{2}{c}{\textit{\textcolor{teal}{Aerial}}} & \multicolumn{2}{c}{\textit{\textcolor{teal}{Aerial}}} & \multicolumn{2}{c}{\textit{\textcolor{teal}{Aerial}}}\\

\bottomrule
\end{tabular}
\end{table*}

To determine the effect of passing the wrong \textit{domain input} when running zero-shot evaluations, we evaluate different domain settings of \shortname-C on 3 in-distribution and 6 zero-shot datasets. As demonstrated in \textbf{Tab.~\ref{table:supp_effect_evaluating_cross_domain}}, the correct domain input is crucial for getting good performance from the model on both in-distribution and zero-shot datasets. Moreover, while using the outdoor domain on indoor scenes completely disrupts performance, using the outdoor domain on aerial scenes still yields reasonable segmentation results, and vice versa. 
This is expected because aerial LiDAR captures are often collected over outdoor environments, which introduces partial similarities between aerial and outdoor domains while still retaining distinct characteristics.

\section{Additional Quantitative Results}
\label{sec:quantitative_results}

\begin{table}[h!]
\centering
\small
\caption{\textbf{Model Efficiency Comparison Results.} We compare the timing and memory consumption on an RTX 3090 GPU when performing single-object segmentation with 1 click.}
\label{table:timing_memory}
\begin{tabular}{l|c c c c}
\toprule
Method & \makecell{Model\\Size (M)} & \makecell{Memory\\(GB)} & \makecell{Inference\\Time (ms)} \\
\midrule
AGILE3D & 39.3 & 1.20 & 203 \\
Point-SAM & 311.0 & 3.70 & 287 \\
Interactive4D & 39.3 & 1.05 & 200 \\
SNAP & 49.6 & 1.27 & 197 \\
\bottomrule
\end{tabular}
\end{table}

\begin{table*}[tb]
\centering
\caption{\textbf{Class Agnostic Instance Segmentation Comparison against Non-Interactive Fully-Supervised Methods}. We compare \shortname\ against state-of-the-art baselines for in-distribution and zero-shot datasets on the instance segmentation task in a class-agnostic fashion.
}
\vspace{-0.8em}
\label{table:fully_supervised_comparison}
\small
\begin{tabular}{l|ccc|ccc|ccc}

\bottomrule
\rowcolor{mediumblue!20}\multicolumn{10}{c}{\textbf{In-distribution Evaluation}} \\
\toprule

\multicolumn{10}{c}{ScanNet200 Validation Set} \\
\midrule
\textbf{Method} & \multicolumn{3}{c|}{mAP} & \multicolumn{3}{c|}{mAP50} & \multicolumn{3}{c}{mAP25} \\
\midrule
EASE~\cite{Roh_2024_CVPR_EASE} & \multicolumn{3}{c|}{29.9} & \multicolumn{3}{c|}{38.8} & \multicolumn{3}{c}{44.7} \\
\midrule
& @1 & @5 & @10 & @1 & @5 & @10 & @1 & @5 & @10 \\
\cmidrule(lr){2-4} \cmidrule(lr){5-7} \cmidrule(lr){8-10}
SNAP - SN200 & 47.9 & 68.1 & 73 & 69.7 & 89.6 & 92.4 & 84.2 & 97.3 & 98.8 \\
SNAP - Indoor & 45.2 & 66.7 & 73.3 & 69.3 & 90.5 & 95.3 & 84.8 & 98.8 & 99.7 \\
SNAP - C & \textbf{49.2} & \textbf{69.8} & \textbf{77.5} & \textbf{73.2} & \textbf{91.6} & \textbf{95.9} & \textbf{87.7} & \textbf{99.1} & \textbf{99.8} \\

\midrule
\multicolumn{10}{c}{STPLS3D Validation Set} \\
\midrule
\textbf{Method} & \multicolumn{3}{c|}{mAP} & \multicolumn{3}{c|}{mAP50} & \multicolumn{3}{c}{mAP25} \\
\midrule
Mask3D~\cite{Schult23ICRA} & \multicolumn{3}{c|}{57.3} & \multicolumn{3}{c|}{74.3} & \multicolumn{3}{c}{81.6}\\
\midrule
& @1 & @5 & @10 & @1 & @5 & @10 & @1 & @5 & @10 \\
\cmidrule(lr){2-4} \cmidrule(lr){5-7} \cmidrule(lr){8-10}
SNAP - Aerial & 56.2 & 72.9 & 80.7 & 74.4 & 88.9 & 94 & 86.5 & 97.1 & 98.6 \\
SNAP - C & \textbf{58.3} & \textbf{75.7} & \textbf{84.4} & \textbf{76.7} & \textbf{91.1} & \textbf{95.3} & \textbf{88.8} & \textbf{98.0} & \textbf{99.1} \\

\midrule
\multicolumn{10}{c}{Semantic KITTI Validation Set} \\
\midrule
\textbf{Method} & \multicolumn{3}{c|}{PQ} & \multicolumn{3}{c|}{SQ} & \multicolumn{3}{c}{RQ} \\
\midrule
Mask4Former~\cite{mask4former} & \multicolumn{3}{c|}{61.7} & \multicolumn{3}{c|}{81} & \multicolumn{3}{c}{71.4} \\
\midrule
& @1 & @5 & @10 & @1 & @5 & @10 & @1 & @5 & @10 \\
\cmidrule(lr){2-4} \cmidrule(lr){5-7} \cmidrule(lr){8-10}
SNAP - KITTI & 68.6 & 83.4 & 87.6 & 80.9 & 84.9 & 88.9 & 82.7 & 91.8 & 97.5 \\
SNAP - Outdoor & 69.9 & 85.1 & 90.1 & 82.3 & 87.9 & 91.3 & 84.1 & 96.3 & 98.3 \\
SNAP - C & \textbf{71.1} & \textbf{86.5} & \textbf{90.7} & \textbf{82.7} & \textbf{88.7} & \textbf{91.7} & \textbf{84.8} & \textbf{97.4} & \textbf{98.4} \\
\bottomrule

\rowcolor{green!20}\multicolumn{10}{c}{\textbf{Zero-Shot Evaluation}} \\
\toprule
\multicolumn{10}{c}{ScanNet++ Validation Set} \\
\midrule
\textbf{Method} & \multicolumn{3}{c|}{mAP} & \multicolumn{3}{c|}{mAP50} & \multicolumn{3}{c}{mAP25} \\
\midrule
LaSSM~\cite{yao2025lassm} & \multicolumn{3}{c|}{29.1} & \multicolumn{3}{c|}{43.5} & \multicolumn{3}{c}{51.6} \\
\midrule
& @1 & @5 & @10 & @1 & @5 & @10 & @1 & @5 & @10 \\
\cmidrule(lr){2-4} \cmidrule(lr){5-7} \cmidrule(lr){8-10}
SNAP - SN200 & 32.9 & 52.1 & 58.1 & 49.1 & 71.9 & 77.3 & 62.9 & 86.1 & 89.6 \\
SNAP - Indoor & 35.9 & 59.8 & 66.5 & \textbf{55.5} & \textbf{84.9} & 89.4 & \textbf{73.3} & \textbf{97.3} & \textbf{98.6} \\
SNAP - C & \textbf{37.7} & \textbf{60.4} & \textbf{70.1} & 55.4 & 83.0 & \textbf{90.9} & 73.2 & 94.6 & 98.1 \\

\midrule
\multicolumn{10}{c}{Matterport3D Validation Set} \\
\midrule
\textbf{Method} & \multicolumn{3}{c|}{mAP} & \multicolumn{3}{c|}{mAP50} & \multicolumn{3}{c}{mAP25} \\
\midrule
ODIN~\cite{jain2024odin} & \multicolumn{3}{c|}{24.7} & \multicolumn{3}{c|}{--} & \multicolumn{3}{c}{63.8} \\
\midrule
& @1 & @5 & @10 & @1 & @5 & @10 & @1 & @5 & @10 \\
\cmidrule(lr){2-4} \cmidrule(lr){5-7} \cmidrule(lr){8-10}
SNAP - SN200 & \textbf{42.7} & 62.3 & 68.8 & \textbf{64.4} & 85.6 & 90.9 & \textbf{77.5} & 95.9 & 97.1 \\
SNAP - Indoor & 36.5 & 63 & 70.7 & 59.2 & \textbf{90.3} & 93.3 & 74.2 & \textbf{98.1} & \textbf{98.8} \\
SNAP - C & 39.2 & \textbf{64.6} & \textbf{74.3} & 59.4 & 88.7 & \textbf{94.5} & 77.3 & 96.9 & 98.7 \\

\midrule
\multicolumn{10}{c}{UrbanBIS Validation Set} \\
\midrule
\textbf{Method} & \multicolumn{3}{c|}{mAP} & \multicolumn{3}{c|}{mAP50} & \multicolumn{3}{c}{mAP25} \\
\midrule
B-Seg~\cite{UrbanBIS} & \multicolumn{3}{c|}{62.1} & \multicolumn{3}{c|}{70} & \multicolumn{3}{c}{73.9} \\
\midrule
& @1 & @5 & @10 & @1 & @5 & @10 & @1 & @5 & @10 \\
\cmidrule(lr){2-4} \cmidrule(lr){5-7} \cmidrule(lr){8-10}
SNAP - Aerial & \textbf{62.9} & 84.2 & 91.1 & \textbf{85.6} & 95.5 & 98.2 & \textbf{96.4} & 99.1 & 98.2 \\
SNAP - C & 62.2 & \textbf{89.1} & \textbf{94.9} & 84.1 & \textbf{100} & \textbf{100} & 94.6 & \textbf{100} & \textbf{100}\\
\bottomrule
\end{tabular}
\end{table*}

\begin{table*}[tb]
\centering
\small
\caption{\textbf{In-distribution Interactive Point Cloud Segmentation Results.} * indicates models not trained on the evaluation dataset and $\dagger$ denotes that the methods are evaluated by us.}
\label{table:in_dist_full_results}
\begin{tabular}{l|l|l|ccccc}
\toprule
\multirow{2}{*}{\textbf{Domain}} & \multirow{2}{*}{\textbf{Dataset}} & \multirow{2}{*}{\textbf{Method}} & \multicolumn{5}{c}{\textbf{IoU@k}} \\
\cmidrule(lr){4-8}
& & & \textbf{@1} & \textbf{@3} & \textbf{@5} & \textbf{@7} & \textbf{@10} \\
\midrule

\multirow{13}{*}{\textbf{Outdoor}} 
& \multirow{5}{*}{SemanticKITTI} 
& AGILE3D~\cite{AGILE3D} & 53.1 & 70 & 76.7 & - & 83  \\
& & Interactive4D~\cite{interactive4D} & 67.5 & 78.3 & 83.4 & - & 88.2  \\
& & SNAP-KITTI & 68.1 & 80.1 & 84.5 & 87.5 & 88.7 \\
& & SNAP-Outdoor & 71.3 & \textbf{81.9} & 85.7 & 87.7 & 89.3 \\
& & SNAP-C & \textbf{71.5} & \textbf{81.9} & \textbf{86} & \textbf{88.1} & \textbf{90} \\
\cmidrule{2-8}

& \multirow{5}{*}{nuScenes} 
&  AGILE3D*~\cite{AGILE3D} & 32.4 & 47.1 & 56.4 & - & 68.4  \\
& & Interactive4D*~\cite{interactive4D} & 45.5 & 57.2 & 64.6 & - & 74.3  \\
& & SNAP-KITTI* & 50.2 & 64.3 & 71.3 & 74.6 & 76.9 \\
& & SNAP-Outdoor & \textbf{72.4} & 83.1 & \textbf{88.1} & 90.2 & 91.2 \\
& & SNAP-C & 72.2 & \textbf{83.3} & \textbf{88.1} & \textbf{90.3} & \textbf{92.2} \\
\cmidrule{2-8}

& \multirow{3}{*}{Pandaset} 
& SNAP-KITTI & 17 & 29.7 & 34.2 & 35.8 & 36.7 \\
& & SNAP-Outdoor & \textbf{60.5} & \textbf{74.8} & 80.1 & 82.1 & 84.3 \\
& & SNAP-C & 56.3 & 74.6 & \textbf{80.2} & \textbf{82.6} & \textbf{84.4} \\
\midrule

\multirow{9}{*}{\textbf{Indoor}} 
& \multirow{6}{*}{ScanNet20} 
& InterObject3D~\cite{InterObject3D} & 40.8 & 63.9 & 72.4 & - & 79.9  \\
& & AGILE3D~\cite{AGILE3D} & 63.3 & 75.4 & 79.9 & - & 83.7  \\
& & Point-SAM†~\cite{POINT-SAM} & 52.7 & 75.9 & 80.6 & 82.9 & 83.3 \\
& & SNAP-SN & \textbf{68.6} & 78.4 & 82.1 & 83.4 & 84.6 \\
& & SNAP-Indoor & 66 & 77.6 & 81.3 & 83 & 84 \\
& & SNAP-C & 67.7 & \textbf{78.5} & \textbf{82.3} & \textbf{84.1} & \textbf{85.5} \\
\cmidrule{2-8}

& \multirow{3}{*}{HM3D} 
& SNAP-SN & 38.7 & 52.4 & 58.6 & 61.4 & 63.3 \\
& & SNAP-Indoor & 47.1 & 65.2 & 71.2 & 74.0 & 75.9 \\
& & SNAP-C & \textbf{50} & \textbf{66.7} & \textbf{72.9} & \textbf{76.1} & \textbf{78.7} \\
\midrule

\multirow{4}{*}{\textbf{Aerial}} 
& \multirow{2}{*}{STPLS3D} 
& SNAP-Aerial & 65.8 & 74.5 & 79.1 & 81.6 & 83.6 \\
& & SNAP-C & \textbf{67.8} & \textbf{75.5} & \textbf{80.4} & \textbf{83.3} & \textbf{85.8} \\
\cmidrule{2-8}

& \multirow{2}{*}{DALES} 
& SNAP-Aerial & 60.7 & 72.5 & 76.8 & 78.7 & 80 \\
& & SNAP-C & \textbf{61.6} & \textbf{74} & \textbf{78.2} & \textbf{80.4} & \textbf{82.3} \\
\bottomrule
\end{tabular}
\end{table*}

\begin{table*}[tb]
\centering
\small
\caption{\textbf{Zero-Shot Interactive Point Cloud Segmentation Results.} $\dagger$ denotes that the methods are evaluated by us.}
\label{table:zero_shot_full_results}
\begin{tabular}{l|l|l|ccccc}
\toprule
\multirow{2}{*}{\textbf{Domain}} & \multirow{2}{*}{\textbf{Dataset}} & \multirow{2}{*}{\textbf{Method}} & \multicolumn{5}{c}{\textbf{IoU@k}} \\
\cmidrule(lr){4-8}
& & & \textbf{@1} & \textbf{@3} & \textbf{@5} & \textbf{@7} & \textbf{@10} \\
\midrule
\multirow{21}{*}{\textbf{Outdoor}}
& \multirow{5}{*}{Waymo} 
& Interactive4D & 7.2 & 7.3 & 7.5 & 7.7 & 7.9 \\
& & Point-SAM & 12.8 & 43 & 53.1 & 57.4 & 60.2 \\
& & SNAP-KITTI & 48.2 & 61.8 & 66.3 & 68.4 & 70 \\
& & SNAP-Outdoor & 68.5 & 81.8 & 86 & 87.4 & 88.3 \\
& & SNAP-C & \textbf{69.8} & \textbf{82.3} & \textbf{86.6} & \textbf{88.2} & \textbf{89.3} \\
\cmidrule{2-8}

& \multirow{4}{*}{KITTI 360 Full} 
& Point-SAM & 6.8 & 22.7 & 28.1 & 30.7 & 32.5 \\
& & SNAP-KITTI & 6.7 & 25.9 & 29.7 & 31.2 & 32.4 \\
& & SNAP-Outdoor & 18.3 & 35.2 & 44.6 & 47.3 & 50.2 \\
& & SNAP-C & \textbf{23.1} & \textbf{40.1} & \textbf{48.1} & \textbf{51.7} & \textbf{54.2} \\
\cmidrule{2-8}

& \multirow{5}{*}{KITTI 360 Single Scan} 
& AGILE3D & 36.3 & 47.3 & 53.5 & - & 63.3  \\
& & Interactive4D & 47.7 & 59.4 & 64.1 & - & 70\\
& & SNAP-KITTI & 54.4 & 60.9 & 63.9 & 65.5 & 66.8 \\
& & SNAP-Outdoor & 59.8 & 63.3 & 65.9 & 67.5 & 69.1 \\
& & SNAP-C & \textbf{60.4} & \textbf{64.6} & \textbf{67.7} & \textbf{70.1} & \textbf{72.6} \\
\cmidrule{2-8}

& \multirow{7}{*}{KITTI 360 Crops} 
& AGILE3D & 34.8 & 42.7 & 44.4 & 45.8 & 49.6 \\
& & Point-SAM & 49.4 & 74.4 & \textbf{81.7} & \textbf{84.3} & \textbf{85.8} \\
& & SNAP-KITTI & 56.1 & 68.8 & 72.8 & 74.3 & 75.3 \\
& & SNAP-Outdoor & 56.9 & 70.6 & 75.6 & 78.1 & 80.3 \\
& & SNAP-SN & 54.5 & 68.6 & 74.1 & 76.8 & 78.6 \\
& & SNAP-Indoor & 54.9 & 70.2 & 76.6 & 79.3 & 80.4 \\
& & SNAP-C & \textbf{65.6} & \textbf{76.1} & 80 & 82.1 & 83.6 \\
\midrule

\multirow{17}{*}{\textbf{Indoor}}
& \multirow{4}{*}{ScanNet++} 
& Point-SAM† & 28.6 & 56.3 & 62.9 & 65.5 & 67.2 \\
& & SNAP-SN & 45.5 & 59.9 & 65.3 & 67.8 & 69.5 \\
& & SNAP-Indoor & 51.5 & \textbf{67.9} & \textbf{73.4} & 75.9 & 77.6 \\
& & SNAP-C & \textbf{52} & 67.3 & 73.2 & \textbf{76.3} & \textbf{78.6} \\
\cmidrule{2-8}

& \multirow{4}{*}{Matterport3D} 
& Point-SAM† & 41.1 & 67.2 & 73.7 & 76.2 & 77.9 \\
& & SNAP-SN & \textbf{53.4} & 66.6 & 71.3 & 73.7 & 75.3 \\
& & SNAP-Indoor & 49.9 & 68.4 & 74.2 & 76.4 & 78.3 \\
& & SNAP-C & 52.6 & \textbf{69.6} & \textbf{75.2} & \textbf{78.2} & \textbf{80.5} \\
\cmidrule{2-8}

& \multirow{5}{*}{S3DIS Crops} 
& AGILE3D & \textbf{58.7} & 77.4 & 83.6 & 86.4 & \textbf{88.5} \\
& & Point-SAM & 45.9 & \textbf{77.6} & \textbf{84.6} & \textbf{86.9} & 88.4 \\
& & SNAP-SN & 55.8 & 68.7 & 74.1 & 77.2 & 79.4 \\
& & SNAP-Indoor & 54.8 & 73.5 & 80.5 & 83.7 & 85.9 \\
& & SNAP-C & 56.6 & 73.8 & 80.9 & 84.4 & 87 \\
\cmidrule{2-8}

& \multirow{4}{*}{S3DIS Full} 
& Point-SAM† & 35.6 & 68 & 76.3 & 78.9 & 80.6 \\
& & SNAP-SN & 51.4 & 64.3 & 70 & 72.6 & 74.8 \\
& & SNAP-Indoor & 51.9 & 70.1 & 76.9 & 79.9 & 81.9 \\
& & SNAP-C & \textbf{53.6} & \textbf{71.1} & \textbf{77.6} & \textbf{80.8} & \textbf{83.2} \\
\midrule

\multirow{3}{*}{\textbf{Aerial}}
& \multirow{3}{*}{UrbanBIS} 
& Point-SAM & 39.3 & 79.1 & 89.4 & 92.7 & 94.3 \\
& & SNAP-Aerial & \textbf{74.2} & 83.2 & 86.9 & 89.8 & 90.6 \\
& & SNAP-C & 71.6 & \textbf{86.2} & \textbf{90.2} & \textbf{92.8} & \textbf{94.7} \\
\bottomrule
\end{tabular}
\end{table*}

\subsection{Timing and Memory Consumption Comparison}
\label{sec:timing_memory}
We compare the computational efficiency of \shortname~with other interactive segmentation methods in \textbf{Tab.~\ref{table:timing_memory}}, reporting inference time and memory consumption on an RTX 3090 GPU for single-object segmentation with 1 click. The results indicate that \shortname~maintains competitive efficiency across both time and memory. For this test, we use the same uniform random point cloud with 100,000 points for all methods.

\subsection{Class-Agnostic Interactive Segmentation against Non-interactive Fully-Supervised Methods}
\label{sec:comparison_against_fully_supervised}

We evaluate our interactive model variants against state-of-the-art \textit{non-interactive baselines} on both in-distribution and zero-shot datasets as a sanity check for their effectiveness. Since \shortname\ variants benefit from click supervision on all objects in the scene, they are expected to outperform non-interactive instance segmentation methods.
As shown in \textbf{Tab.~\ref{table:fully_supervised_comparison}}, all \shortname\ variants achieve substantial gains over the current SOTA method EASE\cite{Roh_2024_CVPR_EASE} on the ScanNet200~\cite{scannet} benchmark, surpassing it by a 19.3 point margin with a single click and further improving as the number of clicks increases (5 and 10). 
This large advantage comes from the fact that predicting 200 categories, especially the long-tailed proportion, is inherently difficult for non-interactive methods, whereas \shortname~benefits from click guidance that helps disambiguate object boundaries, leading to a much stronger performance.
On the aerial STPLS3D~\cite{stpls3d} dataset with 14 semantic classes, \shortname-C@1 Click shows slightly better performance than the current SOTA methods, and as expected, this continues to improve with additional clicks. 
On the outdoor SemanticKITTI~\cite{semantickitti} dataset, \shortname\ variants show very strong performance, significantly outperforming the SOTA Mask4Former~\cite{mask4former} with 1-Click. 

When evaluating zero-shot on unseen datasets like ScanNet++~\cite{scannetpp} and Matterport3D~\cite{Matterport3D}, \shortname\ outperforms these method by 8.6 points on ScanNet++ and 14.5 points on Matterport3D. Notably, both the baseline methods LaSSM~\cite{yao2025lassm} and ODIN~\cite{jain2024odin} were trained on ScanNet++ and Matterport3D datasets respectively. Further on the aerial UrbanBIS~\cite{UrbanBIS} dataset, \shortname\ again outperforms the B-Seg~\cite{UrbanBIS} baseline which was trained on the dataset. With additional clicks, this performance continues to improve.

\subsection{Interactive segmentation results with all model variants}
\label{sec:full_interactive_results}

\textbf{Tab.~\ref{table:in_dist_full_results}} presents results for all \shortname\ variants on in-distribution datasets. For datasets lacking established baselines, we compare against zero-shot results from single-dataset models and in-distribution results from single-domain models. \shortname-C achieves the best 1-click performance on 4/7 datasets, best 3-click results on 6/7 datasets, and optimal performance across all datasets for higher click counts, demonstrating effective performance with a unified model.

\textbf{Tab.~\ref{table:zero_shot_full_results}} evaluates all \shortname\ variants on unseen datasets. \shortname-C outperforms baselines on 6/9 datasets for 1-click performance and maintains strong performance across different click counts (7/9 for 3-click, 6/9 for 5-click, 7/9 for 7-click, and 7/9 for 10-click). This demonstrates robust generalization across diverse domains. Notably, while \shortname-C may not achieve state-of-the-art performance on every individual dataset, it is the only method that operates across all domains with a single set of weights, unlike approaches such as AGILE3D~\cite{AGILE3D} that require separate models for different scene types.

\section{Additional Qualitative Results}
\label{sec:qualitative_results}

\begin{figure*}[ht]
    \centering
    \includegraphics[width=0.8\textwidth]{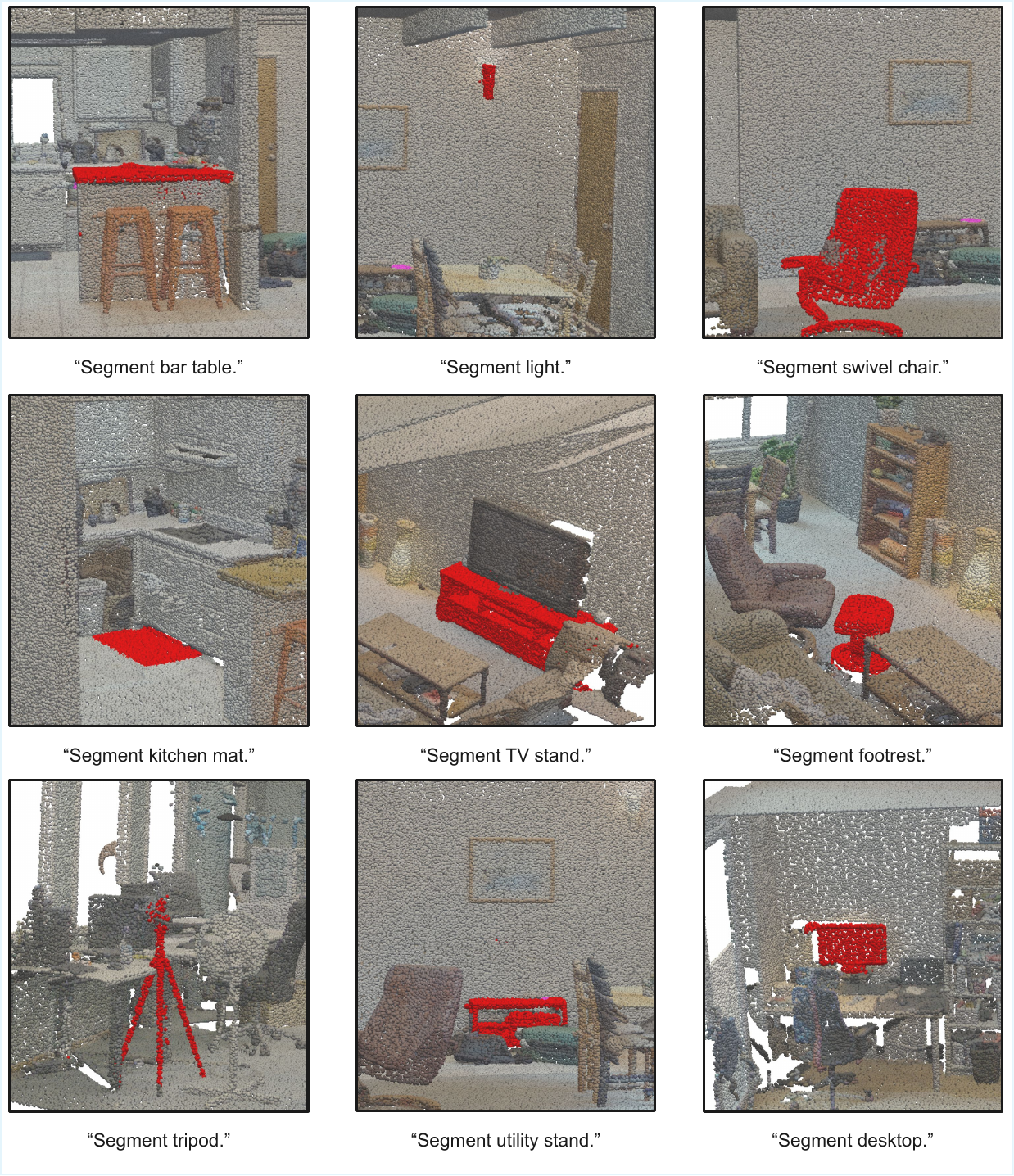}
    \caption{\textbf{Additional qualitative segmentation results of open-set scene understanding on the ScanNet++ Dataset.} Given a text prompt in the format of ``\textit{Segment \{open-set vocabulary\}}'', our \shortname~model finds the corresponding masks {\textcolor{red}{\raisebox{-0.1em}{\rule{0.8em}{0.8em}}}} in the scenes.}
    \label{fig:supp-open-set}
\end{figure*}

\begin{figure*}[ht]
    \centering
    \includegraphics[width=0.78\textwidth]{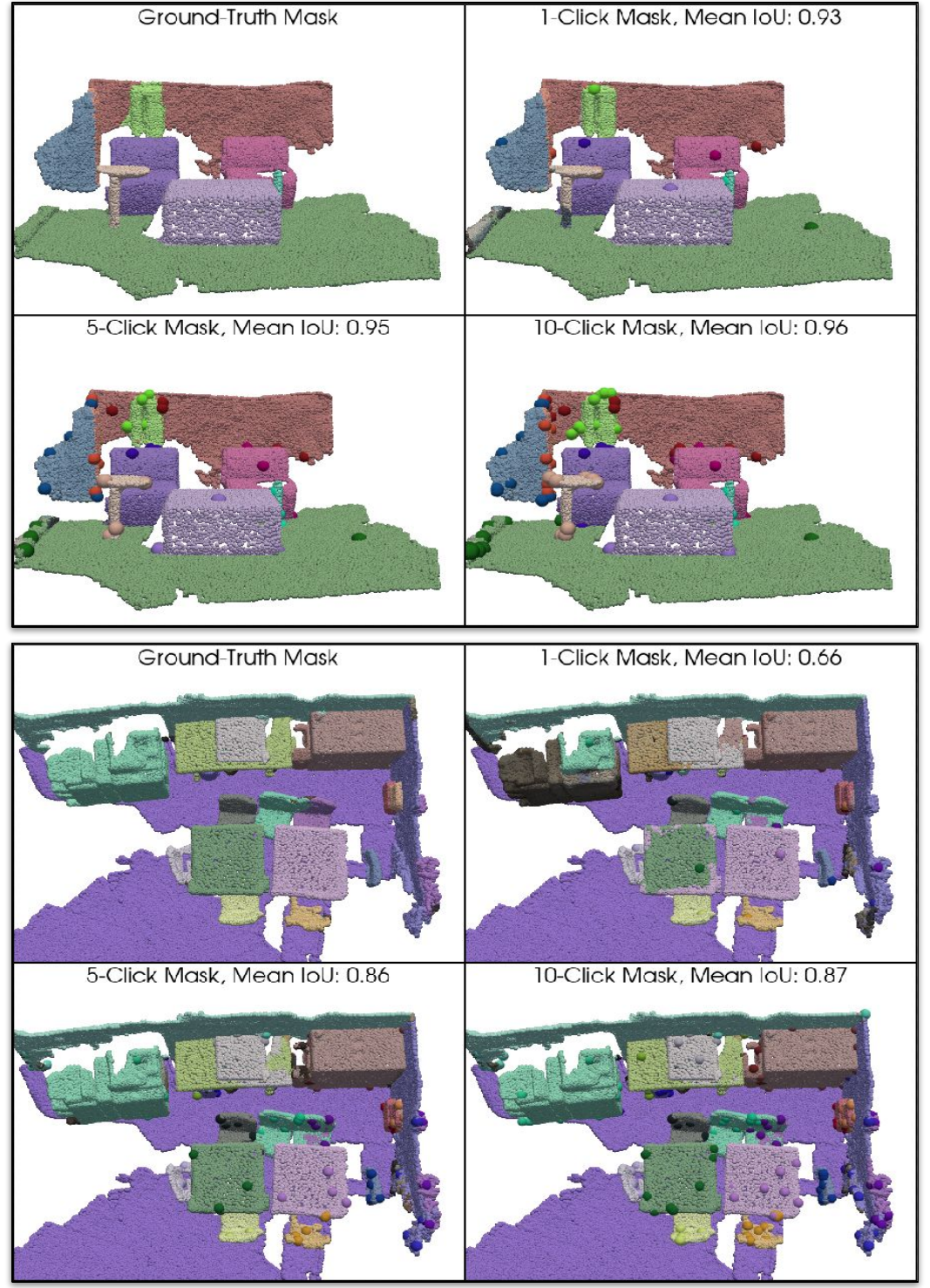}
    \caption{\textbf{Additional qualitative results for point-based segmentation on the ScanNet dataset.} For each block, we show the ground truth masks alongside our segmentation results for 1-click, 5-click, and 10-click interactions, including the corresponding mean IoU values. Points with the same color represent the same object, while clicks are highlighted using darker colors and larger spheres for better visibility.}
    \label{fig:supp-indoor-1}
\end{figure*}

\begin{figure*}[ht]
    \centering
    \includegraphics[width=0.78\textwidth]{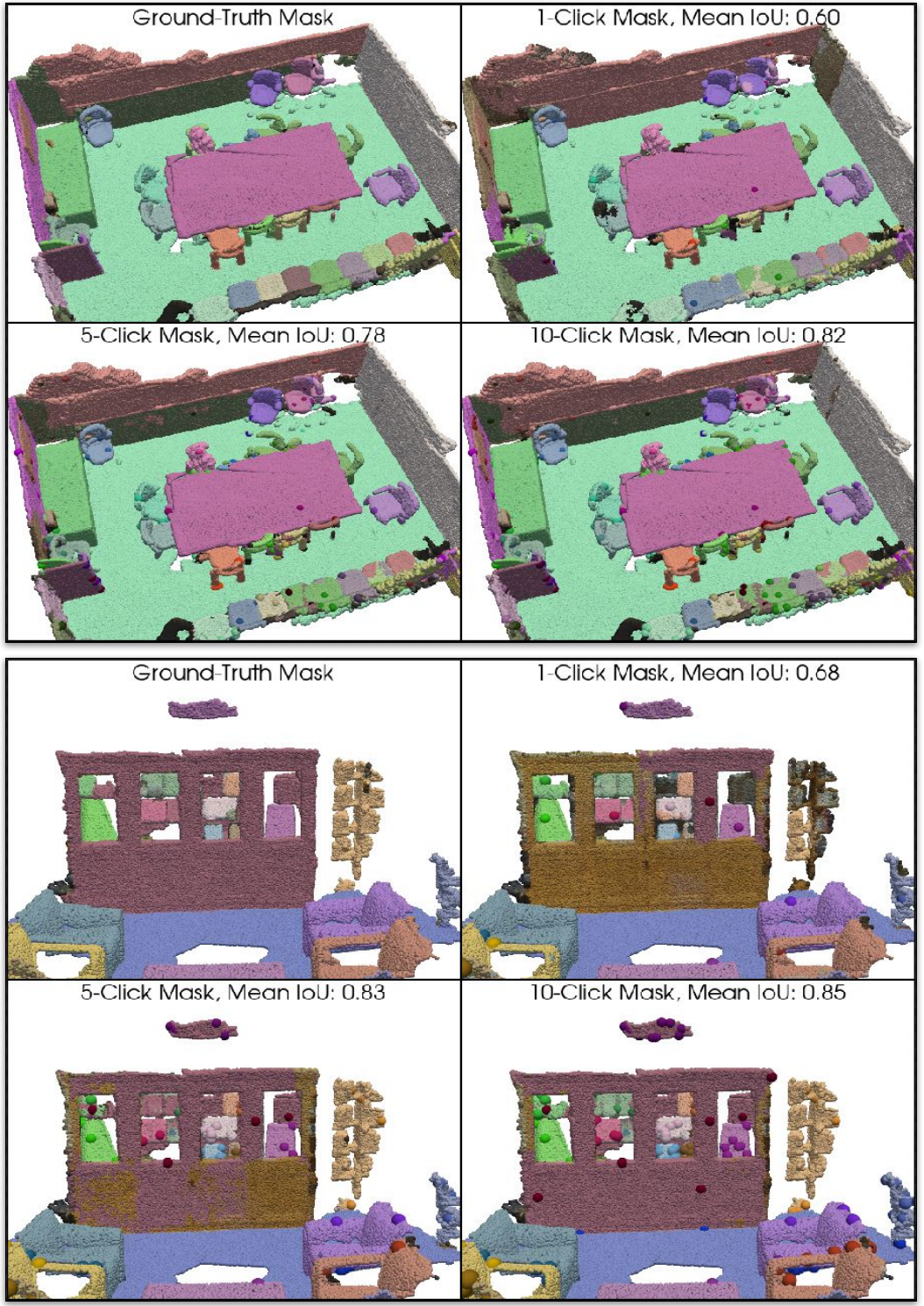}
    \caption{\textbf{Additional qualitative results for point-based segmentation on the ScanNet dataset.} For each block, we show the ground truth masks alongside our segmentation results for 1-click, 5-click, and 10-click interactions, including the corresponding mean IoU values. Points with the same color represent the same object, while clicks are highlighted using darker colors and larger spheres for better visibility.}
    \label{fig:supp-indoor-2}
\end{figure*}

\begin{figure*}[ht]
    \centering
    \includegraphics[width=0.78\textwidth]{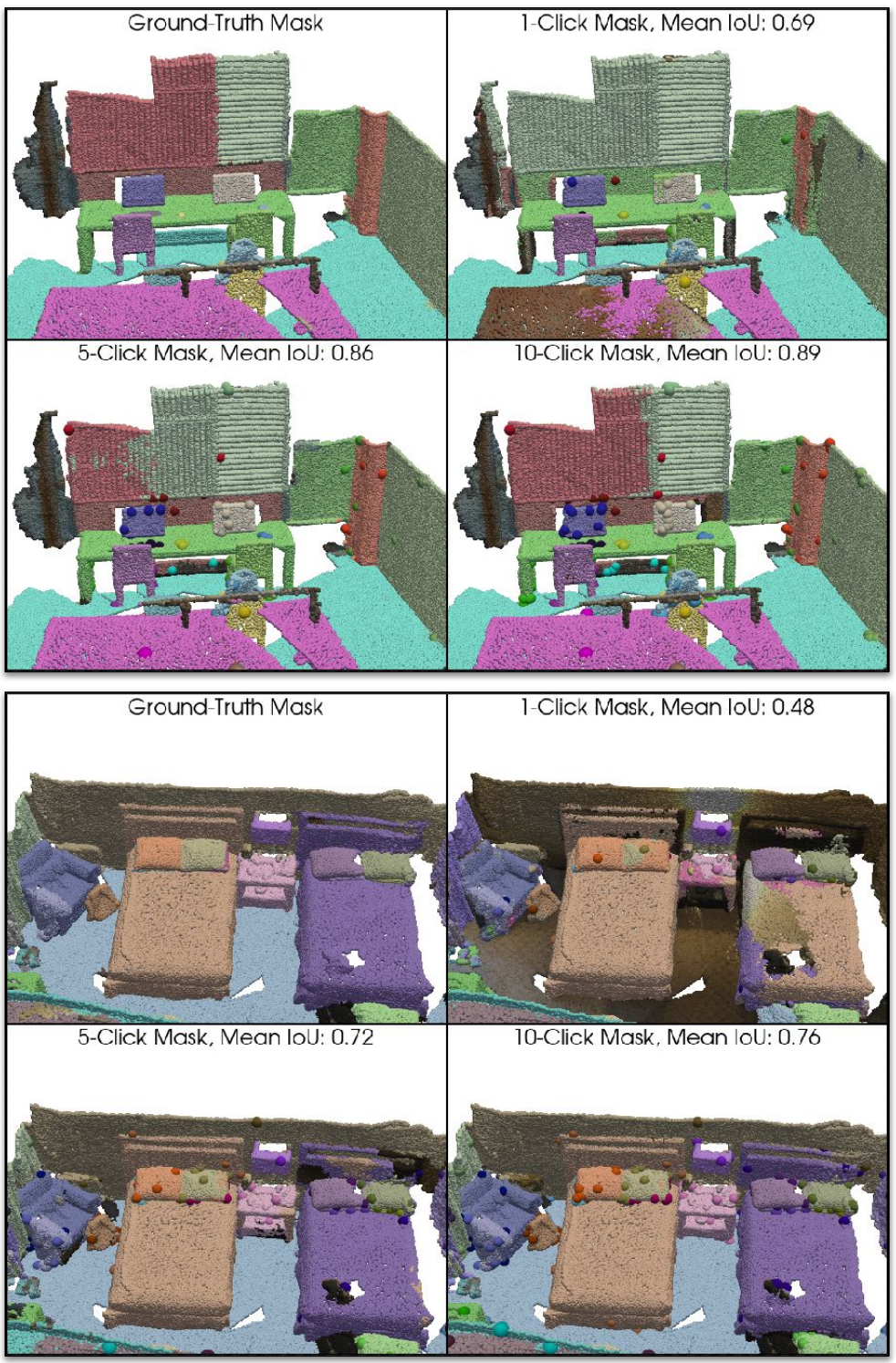}
    \caption{\textbf{Additional qualitative results for point-based segmentation on the ScanNet dataset.} For each block, we show the ground truth masks alongside our segmentation results for 1-click, 5-click, and 10-click interactions, including the corresponding mean IoU values. Points with the same color represent the same object, while clicks are highlighted using darker colors and larger spheres for better visibility.}
    \label{fig:supp-indoor-3}
\end{figure*}

\begin{figure*}[ht]
    \centering
    \includegraphics[width=0.78\textwidth]{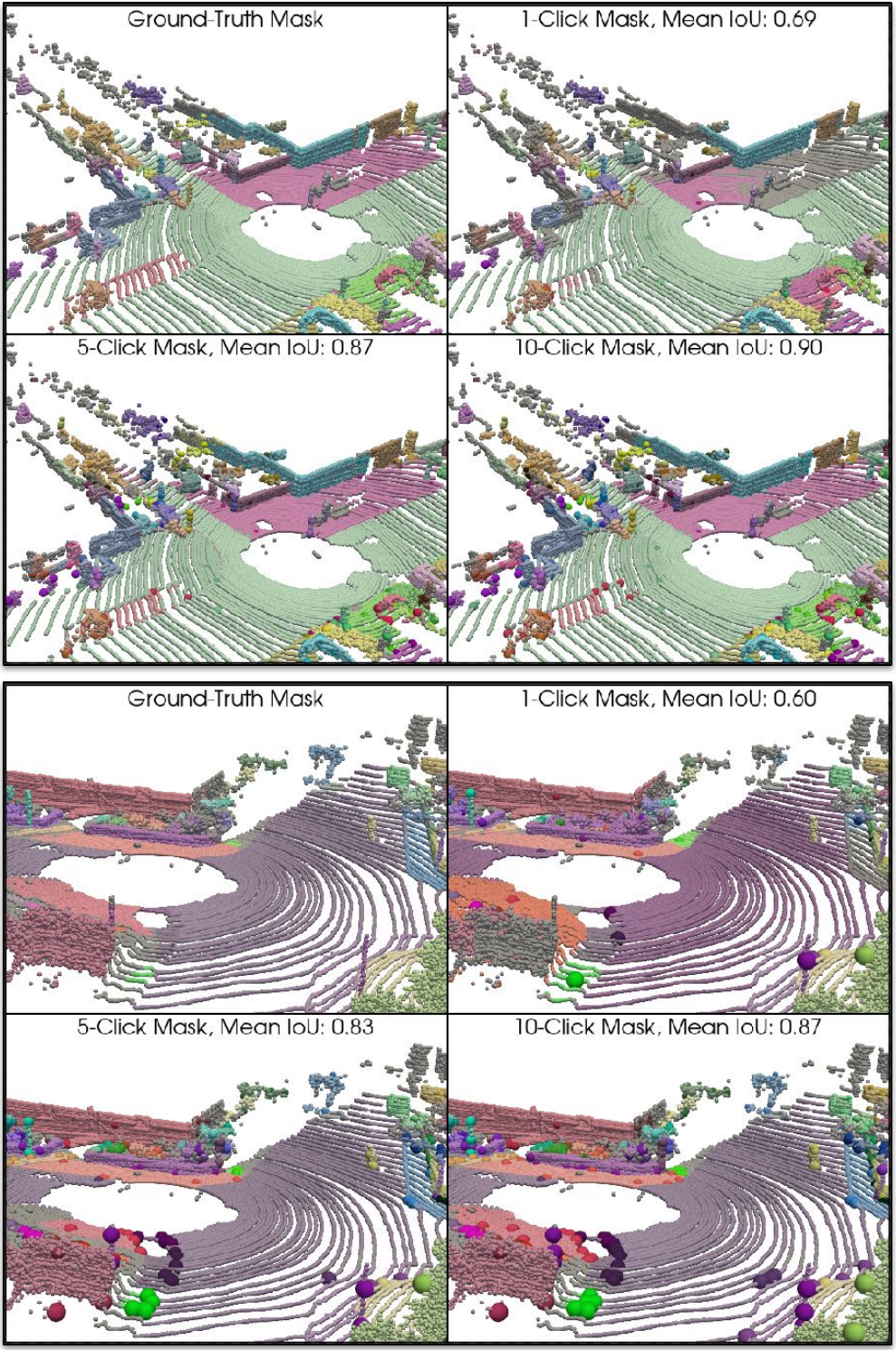}
    \caption{\textbf{Additional qualitative results for point-based segmentation on the SemanticKITTI dataset.} For each block, we show the ground truth masks alongside our segmentation results for 1-click, 5-click, and 10-click interactions, including the corresponding mean IoU values. Points with the same color represent the same object, while clicks are highlighted using darker colors and larger spheres for better visibility.}
    \label{fig:supp-outdoor-1}
\end{figure*}

\begin{figure*}[ht]
    \centering
    \includegraphics[width=0.78\textwidth]{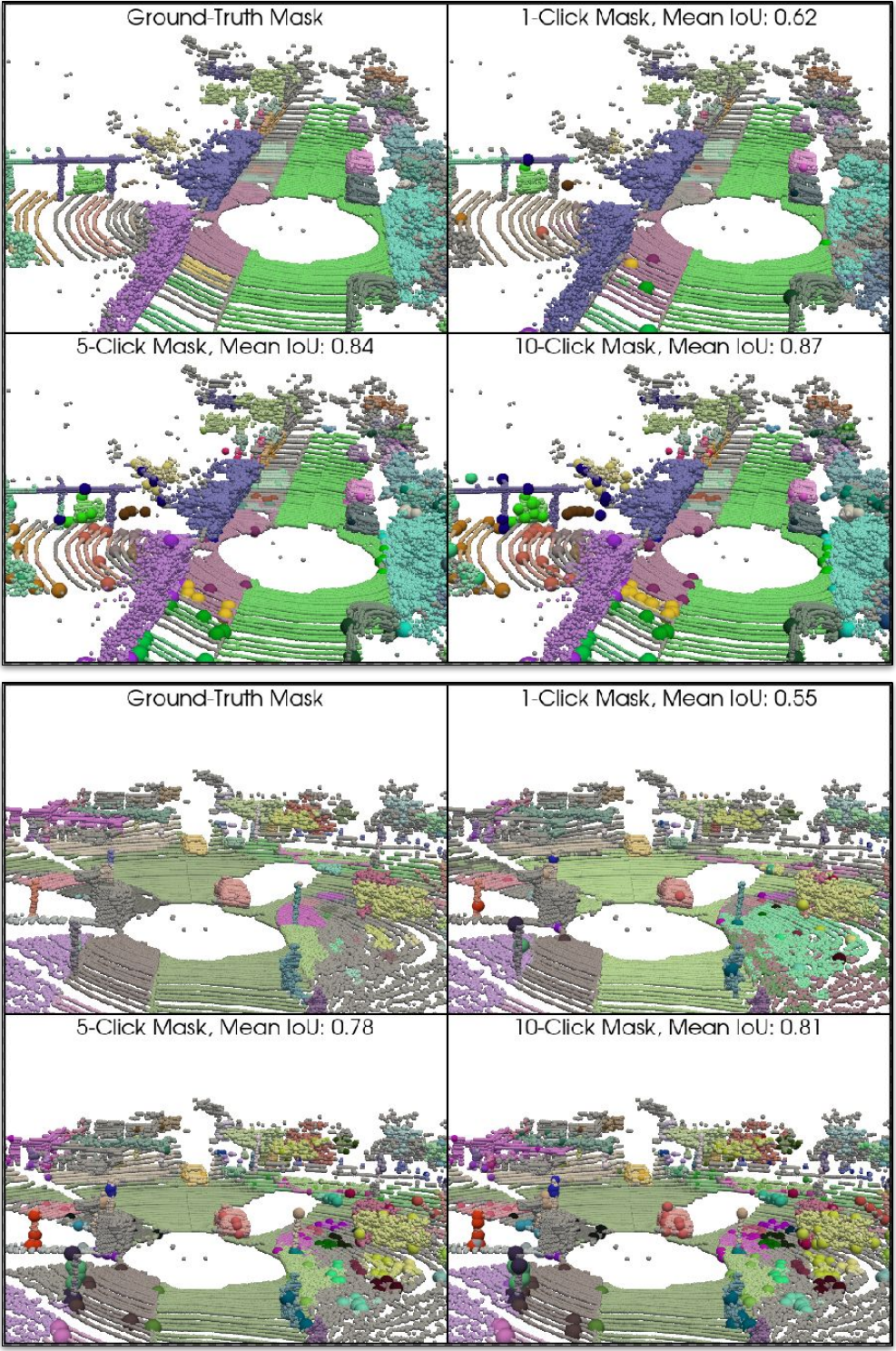}
    \caption{\textbf{Additional qualitative results for point-based segmentation on the SemanticKITTI dataset.} For each block, we show the ground truth masks alongside our segmentation results for 1-click, 5-click, and 10-click interactions, including the corresponding mean IoU values. Points with the same color represent the same object, while clicks are highlighted using darker colors and larger spheres for better visibility.}
    \label{fig:supp-outdoor-2}
\end{figure*}

\begin{figure*}[ht]
    \centering
    \includegraphics[width=0.78\textwidth]{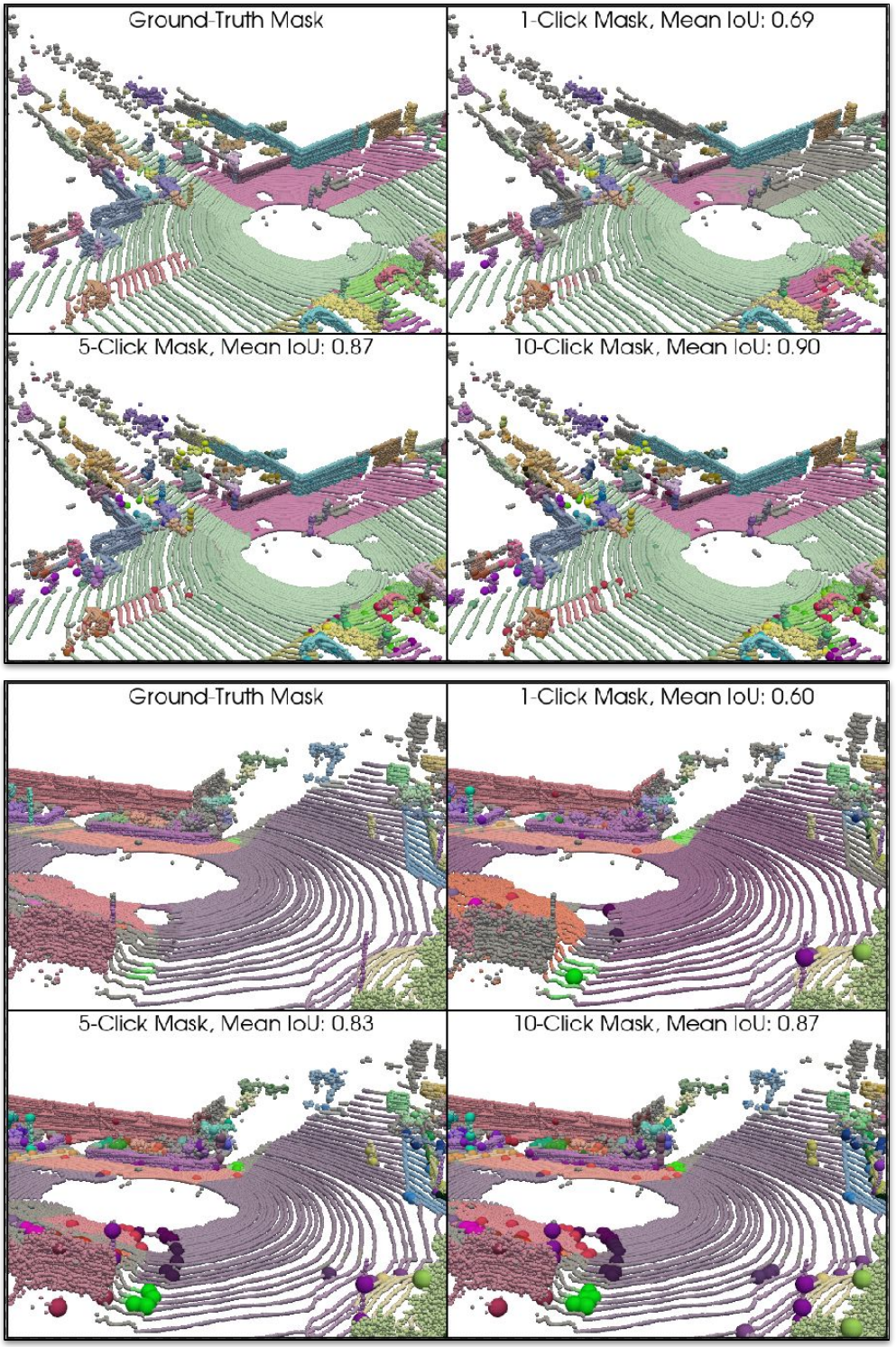}
    \caption{\textbf{Additional qualitative results for point-based segmentation on the SemanticKITTI dataset.} For each block, we show the ground truth masks alongside our segmentation results for 1-click, 5-click, and 10-click interactions, including the corresponding mean IoU values. Points with the same color represent the same object, while clicks are highlighted using darker colors and larger spheres for better visibility.}
    \label{fig:supp-outdoor-3}
\end{figure*}

\begin{figure*}[ht]
    \centering
    \includegraphics[width=0.78\textwidth]{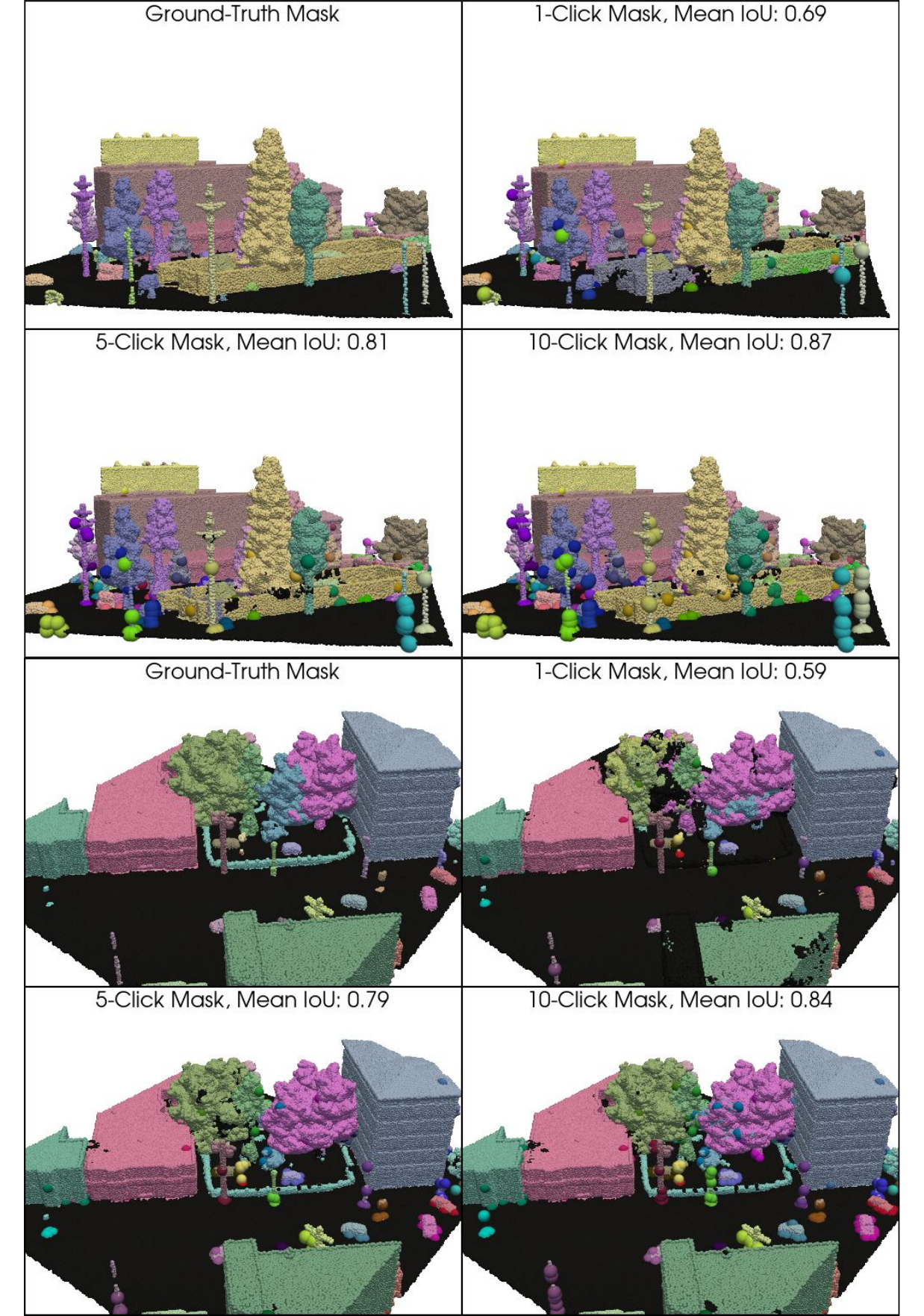}
    \caption{\textbf{Additional qualitative results for point-based segmentation on the STPLS3D dataset.} For each block, we show the ground truth masks alongside our segmentation results for 1-click, 5-click, and 10-click interactions, including the corresponding mean IoU values. Points with the same color represent the same object, while clicks are highlighted using darker colors and larger spheres for better visibility.}
    \label{fig:supp-aerial-1}
\end{figure*}

\begin{figure*}[ht]
    \centering
    \includegraphics[width=0.78\textwidth]{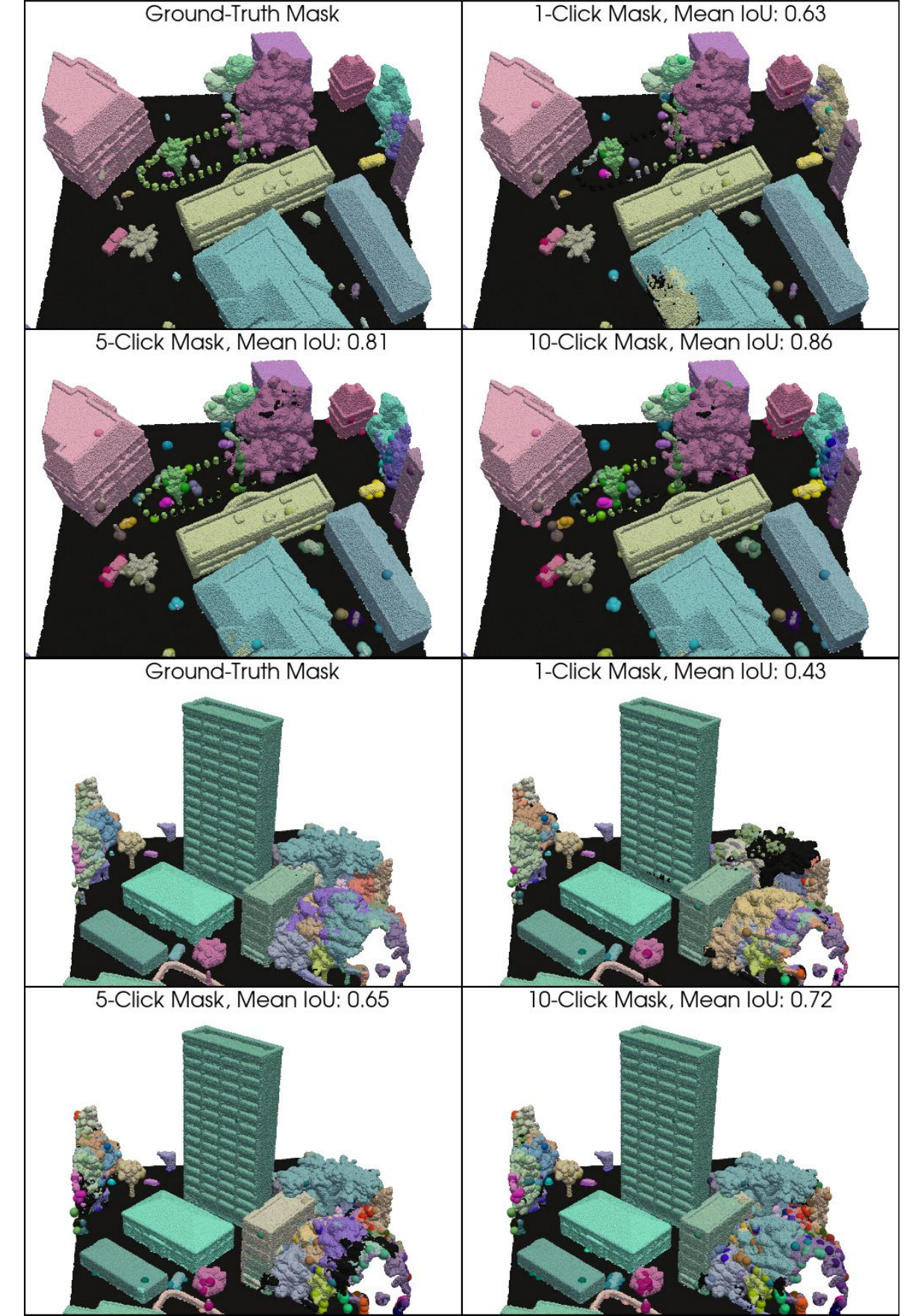}
    \caption{\textbf{Additional qualitative results for point-based segmentation on the STPLS3D dataset.} For each block, we show the ground truth masks alongside our segmentation results for 1-click, 5-click, and 10-click interactions, including the corresponding mean IoU values. Points with the same color represent the same object, while clicks are highlighted using darker colors and larger spheres for better visibility.}
    \label{fig:supp-aerial-2}
\end{figure*}

\begin{figure*}[ht]
    \centering
    \includegraphics[width=0.78\textwidth]{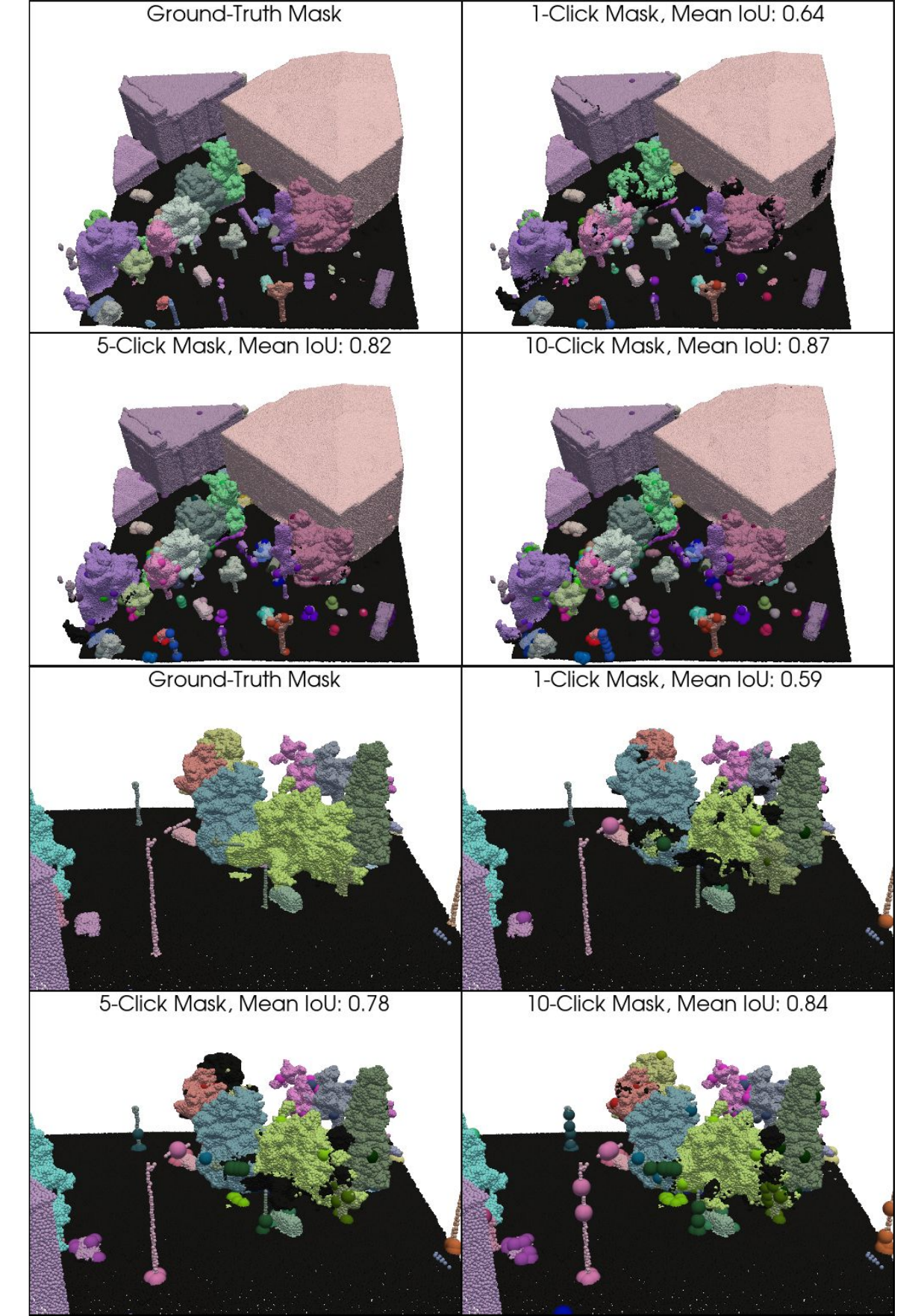}
    \caption{\textbf{Additional qualitative results for point-based segmentation on the STPLS3D dataset.} For each block, we show the ground truth masks alongside our segmentation results for 1-click, 5-click, and 10-click interactions, including the corresponding mean IoU values. Points with the same color represent the same object, while clicks are highlighted using darker colors and larger spheres for better visibility.}
    \label{fig:supp-aerial-3}
\end{figure*}

We provide additional qualitative results, showcasing the performance of our \shortname~model across different tasks. \textbf{Fig.~\ref{fig:supp-open-set}} demonstrates the model's capability in open-vocabulary scene understanding on the ScanNet++ dataset by using arbitrary queries involving object categories that are not present during training. \textbf{Fig.~\ref{fig:supp-indoor-1}-\ref{fig:supp-aerial-3}} present point-based segmentation results on the ScanNet, SemanticKITTI, and STPLS3D datasets, respectively. For each domain, we compare the ground truth masks with our model's outputs under 1-click, 5-click, and 10-click interaction settings, reporting the corresponding mean IoU values. These results demonstrate the effectiveness of our approach in diverse environments, emphasizing the flexibility and robustness of our method across diverse segmentation challenges.

\end{document}